\documentclass[10pt,journal,compsoc]{IEEEtran}
\usepackage{color, colortbl}
\definecolor{Gray}{gray}{0.9}
\usepackage{tikz} 
\usetikzlibrary{calc}
\usetikzlibrary{shapes,arrows} 
\usepackage{enumitem}
\usepackage{fontenc}
\usepackage[font=footnotesize,labelfont=sf,textfont=sf,up]{caption}
\usepackage{color}
\newcommand{\cred}[1]{{\color{black}#1}}
\newcommand{\cblue}[1]{{\color{black}#1}}
\newcommand{\crevision}[1]{{\color{black}#1}}
\newcommand{\cgram}[1]{{\color{black}#1}}

\ifCLASSOPTIONcompsoc
  \usepackage[nocompress]{cite}
\else
  \usepackage{cite}
\fi
\ifCLASSINFOpdf
  \usepackage{graphicx} 
  \usepackage{epstopdf}
  \graphicspath{
    {fig/}
    {fig_paper/}
  }
\else
  \usepackage[dvips]{graphicx}
\fi
\usepackage[cmex10]{amsmath}
\usepackage{amssymb}
\usepackage{algorithm}
\usepackage{algorithmic}

\usepackage{array}

\ifCLASSOPTIONcompsoc
 \usepackage[caption=false,font=footnotesize,labelfont=sf,textfont=sf]{subfig}
\else
 \usepackage[caption=false,font=footnotesize]{subfig}
\fi
\usepackage[pagebackref=true,breaklinks=true,letterpaper=true,colorlinks,bookmarks=false]{hyperref}

\newtheorem{theorem}{Theorem}[section]
\newtheorem{lemma}[theorem]{Lemma}

\newtheorem{proposition}[theorem]{Proposition}

\newtheorem{innercustomtheorem}{Theorem}
\newtheorem{innercustomproposition}{Proposition}
\newtheorem{innercustomlemma}{Lemma}

\newenvironment{customTheorem}[1]
  {\renewcommand\theinnercustomtheorem{#1}\innercustomtheorem}
  {\endinnercustomtheorem}
\newenvironment{customProposition}[1]
  {\renewcommand\theinnercustomproposition{#1}\innercustomproposition}
  {\endinnercustomproposition}
\newenvironment{customLemma}[1]
  {\renewcommand\theinnercustomlemma{#1}\innercustomlemma}
  {\endinnercustomlemma}
  
\newenvironment{proof}{\par\noindent{\bf Proof:\ }}{\hfill$\Box$\\[2mm]}

\def\RR{\mathbb{R}}

\def\FC{\mathcal{F}}

\def\TC{\mathcal{T}}
\def\GC{\mathcal{G}}

\def\bx{\mathbf{x}}
\def\by{\mathbf{y}}
\def\bz{\mathbf{z}}
\def\bt{\mathbf{t}}

\def\tbx{\tilde{\mathbf{x}}}
\def\tby{\tilde{\mathbf{y}}}
\def\tbz{\tilde{\mathbf{z}}}

\newcommand{\bu}{\mathbf{u}}
\def\bv{\mathbf{v}}
\def\be{\mathbf{e}}

\def\MF{\textit{F}}

\def\MG{\textit{G}}

\def\<{\langle}
\def\>{\rangle}

\def\set#1{\{ #1 \}}
\def\Set#1{\left\{ #1 \right\}}
\def\Bigbar#1{\mathrel{\left|\vphantom{#1}\right.}}
\def\Setbar#1#2{\Set{#1 \Bigbar{#1 #2} #2}}

\newcommand{\inner}[1]{\left\langle#1\right\rangle}

\def\R{\mathbb{R}}

\newcommand{\norm}[1]{\left\|#1\right\|}
\newcommand{\abs}[1]{\left|#1\right|}
\def\bydef{:=}
\def\Nc{\mathcal{N}}

\def\ones{\mathop{\rm e}\nolimits}

\def\argmax{\mathop{\rm arg\,max}\limits}

\def\min{\mathop{\rm min}\nolimits}
\def\max{\mathop{\rm max}\nolimits}
\def\ones{\mathbf{1}}

\def\eps{\varepsilon}

\def\eg{\emph{e.g.}~} 
\def\ie{\emph{i.e.}~}

\def\etal{\emph{et al.}~}

\begin{document}
%
\title{An Efficient Multilinear Optimization Framework for Hypergraph Matching}
%
%
%
%

\author{
Quynh~Nguyen,
~Francesco~Tudisco,
~Antoine~Gautier,
~and~Matthias~Hein
\IEEEcompsocitemizethanks{
\IEEEcompsocthanksitem Q. Nguyen, F. Tudisco, A. Gautier and M. Hein are with the Department
of Mathematics and Computer Science, Saarland University, Germany.\protect\\
Corresponding Author: Quynh Nguyen. \protect\\
E-mail: quynh@cs.uni-saarland.de
}
}

\IEEEtitleabstractindextext{
\begin{abstract}
Hypergraph matching has recently become a popular approach 
for solving correspondence problems in computer vision as it allows \cgram{the use of} higher-order geometric information. 
Hypergraph matching can be formulated as a \cblue{third-order} optimization problem subject 
to assignment constraints which turns out to be NP-hard. 
In recent work, we have proposed an algorithm for hypergraph matching
which first lifts the third-order problem to a fourth-order problem
and then solves the \cblue{fourth-order} problem via optimization of the corresponding multilinear form. This leads
to a tensor block coordinate ascent scheme which has the guarantee of providing monotonic ascent in the
original matching score function and leads to state-of-the-art performance both in terms of achieved matching
score and accuracy. In this paper we show that the lifting step to a \cblue{fourth-order} problem can be avoided yielding
a third-order scheme with the same guarantees and performance but being two times faster. Moreover, we introduce a homotopy
type method which further improves the performance. 
\end{abstract}


} 

\maketitle

\IEEEdisplaynontitleabstractindextext

%
\IEEEpeerreviewmaketitle

\IEEEraisesectionheading{\section{Introduction}\label{sec:introduction}}

\IEEEPARstart{G}{raph} matching is among the most challenging tasks of graph processing and
lies at the heart of various fields in pattern recognition, machine learning and computer vision.
In computer vision, it has been used for solving several \cred{types} of problems,
for instance, object recognition \cite{Low1999}, feature correspondence \cite{ChoLee2012,TorEtal2008}, 
texture regularity discovery \cite{HaysEtal2006}, shape matching \cite{DucJouPon2011,SharEtal2011,ZasEtl2009}, 
object tracking \cite{AroGlo2013} and surface registration \cite{Zeng2010}. 
Graph matching also finds its applications in \cred{document} processing tasks like optical character recognition
\cite{LeeLiu1999, FilatovEtal1995}, 
image analysis (2D and 3D) \cite{WangHancock2006, LuoHancock2000, CarHancork2003, SchSchnor2005},
or bioinformatics \cite{SinghEtal2007, WangEtal2004, Taylor2002}. 

In this paper, we focus on the application of graph matching to the feature correspondence problem.
Given sets of points/features extracted from two images, the task is to find for each point/feature in the first image
a corresponding point/feature in the second image while maximizing as much as possible the similarity between them.
The most simple \cblue{first-order} approach is to match each feature in the first set to its nearest neighbor in the second set
by comparing their local descriptors.
This naive approach fails in the presence of repeated patterns, textures or non-discriminative local appearances.
To overcome this problem, \cblue{second-order} methods 
\cite{LeoHeb2005, LeoHeb2009, LeoHebSuk2009, LeeChoLee2010, ChoEtal2013, ShiEtal2007, TorEtal2008, ZhoTor2012, ZhoTor2013}
try to enforce geometric consistency between pairs of correspondences, 
for instance using distances between points.
However, pairwise relations are often not enough to capture the entire geometrical structure of point set.
Therefore, higher-order approaches \cite{ZasSha2008, CheKel2010,  DucEtAl2011, LeeChoLee2011, Zeng2010} 
that take into account \cblue{higher-order} relations between features/points have been proposed.
This paper falls into this line of research. 
In particular, we cast the correspondence problem as a hypergraph matching problem using higher-order similarities 
instead of unary or pairwise ones used by previous methods.
Depending on the chosen similarities, this formulation allows \cgram{for both scaling and rotation invariance}.
However, instead of \cred{concentrating} on how to build these similarities, 
see \cite{DucEtAl2011, ChengEtal2012, HartleyZiss1994, MundyEtal1993},
the focus of this paper is how to solve the resulting optimization problem.

Compared to graph matching, algorithms for hypergraph matching are less studied in the computer vision literature.
The reason might be the difficult optimization problem which requires not only to deal with 
the combinatorial constraints but also to maximize a high degree polynomial function.
Most prior work has relaxed the constraints in order to use concepts 
from continuous optimization \cite{DucEtAl2011, LeeChoLee2011,ZasSha2008,CheKel2010}.
Our main idea is to reformulate the problem. 
Instead of maximizing a \cblue{third-degree} polynomial we maximize the corresponding multilinear form. 
This allows \cgram{the direct integration of} assignment constraints and yields a simple yet efficient block-coordinate ascent framework.
The idea of this paper is based on our previous work \cite{Quynh2015}, 
where the third-order problem has been lifted to a \cblue{fourth-order} problem in order to show
the equivalence of maximizing the score function and its associated multilinear form. 
In this paper we show by a careful analysis that the lifting to a \cblue{fourth-order} problem can be avoided. 
The resulting \cblue{third-order} algorithms based on maximization of the \cblue{third-order} multilinear form 
have the same guarantees and properties as in \cite{Quynh2015} and achieve similar state-of-the-art matching score and 
\cgram{matching accuracy} while being two times faster. 
Moreover, we provide in this paper a quite generic \cgram{framework}, which
allows \cgram{the adaptation of} our whole approach to matching problems with different assignment constraints,
as well as to other general third-order optimization problems with combinatorial constraints. 
 
Extensive experiments on both synthetic and realistic datasets show that 
all our algorithms, including the third-order and fourth-order ones, significantly outperform the current state-of-the-art
in terms of both matching score and matching accuracy,
in particular for very challenging settings where different kinds of noise are present in the data.
In terms of running time, our algorithms are also on average significantly faster than previous approaches.
All proofs and additional experimental results can be found in the appendix.

\subsection{Related Work}
The graph resp. hypergraph matching problem is known to be NP-hard 
except for special cases where polynomial-time algorithms exist \eg for planar graphs \cite{HopWong1974}.
In order to make the problem computationally tractable, a myriad of approximate algorithms \cgram{have} been proposed
over the last three \cred{decades} aiming at an acceptable trade-off between the complexity of the algorithm and matching accuracy. 
They can be categorized from different perspectives and 
\cgram{we refer to \cite{ConteEtal2004, FoggiaEtal2014} for an extensive review.}
In this section, we review only those approximate algorithms that use Lawler's formulations for both graph and hypergraph matching
as they are closely related to our work.
In particular, the graph matching problem can be formulated as a quadratic assignment problem (QAP) 
$$
    \max\limits_{\bx\in M} \bx^T A \bx
$$
while the hypergraph matching problem is formulated as a higher-order assignment problem (HAP)
$$
    \max\limits_{\bx\in M} \sum_{i,j,k=1}^n \FC_{ijk} \bx_i \bx_j \bx_k
$$
In these formulations, $A$ and $\FC$ refer to the affinity matrix and affinity tensor respectively,
and $M$ is some matching constraint set depending on specific applications.
Also, depending on whether the problem is graph matching or hypergraph matching, 
the corresponding algorithms will be called \cblue{second-order} methods or higher-order methods.

Among recent \cblue{second-order} methods, 
Leordeanu and Hebert \cite{LeoHeb2005} \cgram{introduce} the Spectral Matching (SM) algorithm, 
and Cour \etal \cite{ShiEtal2007} \cgram{introduce} Spectral Matching with Affine Constraint (SMAC). 
Both algorithms are based on the best rank-1 approximation of the affinity matrix. 
Torresani \etal \cite{TorEtal2008} design a complex objective function which can be efficiently optimized by dual decomposition.
Leordeanu \etal \cite{LeoHebSuk2009} propose the Integer Projected Fixed Point (IPFP) algorithm, which optimizes the 
quadratic objective function \cgram{via} a gradient-type algorithm interleaved with projection onto the assignment constraints using
the Hungarian method.
Lee \etal \cite{LeeChoLee2010} tackle the graph matching problem using stochastic sampling, whereas
Cho \etal \cite{ChoLeeLee2010} introduce a reweighted random walk (RRWM) where the reweighting is done in order to
enforce the matching constraints.
Zaslavskiy \etal \cite{ZasEtl2009} propose a path-following algorithm based on convex-concave relaxations.
Zhou and Torre \cite{ZhoTor2012} factorize the affinity matrix and 
then compute a convex-concave relaxation of the objective function which is finally solved by a path-following method.
Along this line there is also work of Liu \etal \cite{LiuEtal2012} and Zhou and Torre \cite{ZhoTor2013} 
extending previous methods \cite{ZhoTor2012, ZasEtl2009} to deal with directed graph matching.
Recently, Cho \etal \cite{ChoEtAl2014} propose a novel max pooling matching algorithm, 
in which they tweak the power method to better cope with noise in the affinities.
Although the algorithm comes without theoretical guarantees, it turns out to perform very well in practice, in particular, when
one has a large number of outliers.

The hypergraph matching problem (\ie HAP) is much less studied in the literature.
Duchenne \etal \cite{DucEtAl2011} formulate the hypergraph matching problem 
as a tensor eigenvalue problem and propose a \cblue{higher-order} power method for solving it.
Zass and Shashua \cite{ZasSha2008} introduce a probabilistic view on the problem with their Hypergraph Matching Method (HGM). 
Their idea is to marginalize the tensor to a vector and solve then the lower dimensional problem.
Chertok and Keller \cite{CheKel2010} extend this idea and marginalize
the tensor to a matrix, leading to a quadratic assignment problem which is then tackled using spectral methods.
Since both methods are based on tensor marginalization, some part of the higher-order information is lost.
Moreover, they cannot handle the one-to-one matching constraint during the iterations 
\cgram{since} it is only considered at the final discretization step.
Lee \etal \cite{LeeChoLee2011} extend the reweighted random walk approach of \cite{LeeChoLee2010} to hypergraph matching.
Their algorithm aims at enforcing the matching constraint via
a bi-stochastic normalization scheme done at each iteration.
In \cite{Zeng2010}, Zeng \etal propose to use pseudo-boolean optimization \cite{Boros2002} for 3D surface matching,
where \cblue{higher-order} terms are decomposed into \cblue{second-order} terms and 
the quadratic pseudo-boolean optimization (QPBO) algorithm \cite{Kolmogorov2007} is employed to solve the problem.
We have proposed in \cite{Quynh2015} a tensor block-coordinate ascent framework, see also \cite{Yan2015},
for hypergraph matching based on a multilinear reformulation of the HAP where the \cblue{third-order} problem has been lifted to
a \cblue{fourth-order} problem and \crevision{one directly optimizes over the set of assignment matrices}.

\subsection{Notation}
Vectors are denoted by bold lowercase letters (\eg $\bx$).
The elements of vectors are denoted by subscripts while superscripts denote iteration numbers.
Tensors will be denoted by calligraphic uppercase letters (\eg $\FC$), while their associated multilinear forms
will be denoted by the corresponding standard uppercase letter (\eg $\MF$).
The symbol $\otimes$ is used to denote the tensor product.

\section{Hypergraph Matching Formulation}\label{sec:form}
\begin{figure}
\centering
\begin{tikzpicture}[vertex/.style={circle,fill=black,minimum size=7pt,inner sep=0}]
	  \path (0,2) .. controls (1,0) and (2,1) .. (1,3)
	      \foreach \p/\name/\r in {0//,0.1/$i_1$/i1,0.45/$j_1$/j1,0.75//,0.90/$k_1$/k1} 
	      {node[vertex,pos=\p,name=\r,label=above right:\name]{}};
	  \path (4,3) .. controls (3,1) and (5,0) .. (6,3)
	      \foreach \p/\name/\r in {0//,0.15/$i_2$/i2,0.52/$j_2$/j2,0.88/$k_2$/k2,0.98//} 
	      {node[vertex,pos=\p,name=\r,label=above right:\name]{}};
	  \draw[-,red,thick] (i1) -- (j1);
	  \draw[-,red,thick] (i1) -- (k1);
	  \draw[-,red,thick] (j1) -- (k1);
	  \draw[-,red,thick] (i2) -- (j2);
	  \draw[-,red,thick] (i2) -- (k2);
	  \draw[-,red,thick] (j2) -- (k2);
	  \draw[-,dashed,draw=blue] (i1) to[bend left=15] node[above,near end]{} (i2);
	  \draw[-,dashed,draw=blue] (j1) to[bend left=15] node[above,midway]{} (j2);
	  \draw[-,dashed,draw=blue] (k1) to[bend left=23] node[above,midway]{} (k2);
	  \draw (i1) +(-25:2.5mm) arc (-25:45:2.5mm);
	  \draw (j1) +(90:3mm) arc (90:150:3mm);
	  \draw (k1) +(225:3mm) arc (225:270:3mm);
	  \draw (i2) +(-65:3mm) arc (-65:-5:3mm);
	  \draw (j2) +(40:3mm) arc (40:115:3mm);
	  \draw (k2) +(220:3mm) arc (220:180:3mm);
\end{tikzpicture}      
\vspace{-20pt}
\caption{Illustration of constructing the affinity tensor.
Each entry is computed by comparing the angles of two triangles formed by three candidate correspondences
$(i_1, i_2), (j_1, j_2)$ and $(k_1, k_2)$.
}
\label{fig:tensor_construction}
\end{figure}
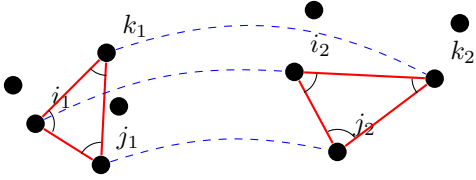

Hypergraphs are powerful in matching problems because they allow modeling of relations involving groups of more than two vertices 
so that \cblue{higher-order} information can be integrated. 
We concentrate our study on $3$-uniform hypergraphs, \ie each hyperedge describes a relation between up to $3$ vertices. 
Every $3$-uniform hypergraph can be represented by a \cblue{third-order} tensor.
Let \cblue{$V$ and $V'$} be two point sets with $n_1= |V|\leq n_2= |V'|$. 
The matching problem consists of finding a binary assignment matrix $X \in \{0,1\}^{n_1 \times n_2}$ 
such that $X_{ij}=1$ if $v_{i} \in V$ matches $v'_{j} \in V',$ and $X_{ij}=0$ \cgram{otherwise}. 
The set of all possible assignment matrices that assign each vertex of $V$ to exactly one vertex of $V'$ is given by
\[ M=\big\{X \in \{0,1\}^{n_1 \times n_2} \,\big|\, \sum_{i=1}^{n_1} X_{ij}\leq 1, \quad \sum_{j=1}^{n_2} X_{ij} = 1\big\}.\]
Suppose that we have a function $\FC: (V \times V')^3\to\R_+$ which maps each pair of triples 
$\{v_{i_1},v_{i_2},v_{i_3}\}\subset V$ and $\{v'_{j_1},v'_{j_2},v'_{j_3}\} \subset V'$ to its similarity weight 
$\FC_{(i_1,j_1),(i_2,j_2),(i_3,j_3)}$. 
For example, $\FC$ is calculated as the similarity of the corresponding angles of the two triangles $\bigtriangleup i_1 j_1 k_1$ and
$\bigtriangleup i_2 j_2 k_2$ as shown in Fig. \ref{fig:tensor_construction}.
Then, the score $S(X)$ of a matching $X\in M$ is defined as follows \cite{DucEtAl2011}
\[ S(X)=\!\!\!\!\!\sum_{i_1,i_2,i_3=1}^{n_1} \sum_{j_1,j_2,j_3=1}^{n_2} \!\!\!\!\FC_{(i_1,j_1),(i_2,j_2),(i_3,j_3)} X_{i_1 j_1} X_{i_2 j_2} X_{i_3 j_3}.\]
For ease of notation, we introduce a linear ordering in $V \times V'$  so that each $X\in M$ 
can be rewritten as a vector $\bx \in \{0,1\}^{n}$ with $n=n_1 n_2$ and $\FC$ as a tensor in $\R^{n \times n \times n}$. 
With this convention, the score function $S:\R^n \to\R$ can be computed in the following way
\begin{equation}\label{eq:HGM}
    S(\bx)=\sum_{i,j,k=1}^n \FC_{ijk}\, \bx_i\, \bx_j\, \bx_k.
\end{equation}
We recall that a \cblue{third-order} tensor $\TC\in\R^{n\times n\times n}$ is called symmetric 
if its entries $\TC_{ijk}$ are invariant under any permutation of $\{i,j,k\}$. 
In particular, the tensor 
$\TC_{ijk} \bydef \bx_i \bx_j \bx_k$ 
is symmetric and thus the non-symmetric part of $\FC$ is ``averaged'' out in the computation of $S(\bx)$. 
\cgram{Therefore, without loss of generality}, we can assume that $\FC$ is symmetric\footnote{Note that $\FC$ can always be symmetrized (without changing the score function) as follows
$\tilde{\FC}_{ijk}=\frac{1}{3!}\sum_{\sigma \in\mathfrak{S}_3}\FC_{\sigma(i)\sigma(j)\sigma(k)}$, where $i,j,k=1,\ldots,n$ and 
$\mathfrak{S}_3$ is the permutation group of three elements.}. 
In this paper, we consider only terms of order $3$, \ie $\FC_{ijk}=0$ if $i=j$, $i=k$ or $j=k$. 
However, note that it is possible to integrate \cblue{first-order} terms on the main diagonal $\FC_{iii}$, 
and pairwise potentials $\FC_{ijj}$ for $i\neq j$.
This is also \crevision{the} major advantage of hypergraph matching over conventional graph matching methods
because one can combine local appearance similarities, pair-wise similarities (\eg distances)
and higher-order invariants (\eg angles of triplets of points) in a unified framework.

\section{Mathematical Foundations for Multilinear Optimization Framework}
In this section, we derive the basis for our tensor block coordinate ascent framework for solving hypergraph matching problems.
The general idea is to optimize instead of the score function $S$
\crevision{
the associated third-order multilinear form $\MF\colon\R^n \times\R^n\times \R^n \to\R$ which is defined as
\begin{equation}\label{eq:def_mult}
   \MF(\bx,\by,\bz) = \sum_{i,j,k=1}^n \FC_{ijk}\,  \bx_{i}\,  \by_j\,  \bz_k .
\end{equation}
}

It turns out that the score function in Eq. \eqref{eq:HGM} is a special case of the multilinear form when all the arguments are 
the same, \crevision{\ie $S(\bx) = \MF(\bx,\bx,\bx).$}
\crevision{
The multilinear form is called symmetric if it is invariant under any permutation of its arguments, 
\ie $\MF(\bx^1, \bx^2, \bx^3) = \MF(\bx^{\sigma(1)}, \bx^{\sigma(2)}, \bx^{\sigma(3)})$ 
for all $\bx^1, \bx^2, \bx^3\in \RR^n$ and $\sigma$ a permutation of $\set{1,2,3}.$
Note that if the tensor $\FC$ is symmetric then $\MF$ is also symmetric.
}
In the following, we write $\MF(\,\cdot\,,\by,\bz)$ to denote a vector in $\RR^n$ such that 
$\MF(\,\cdot\,,\by,\bz)_{i} = \sum_{j,k=1}^n \FC_{ijk} \by_j \bz_k$ for all $1 \leq i \leq n,$
and write $\MF(\,\cdot\,,\,\cdot\,,\bz)$ to denote a matrix in $\RR^{n\times n}$ such that 
$\MF(\,\cdot\,,\,\cdot\,,\bz)_{ij} = \sum_{k=1}^n \FC_{ijk} \bz_k$ for all $1 \leq i,j \leq n.$
Note that the positions of the dots do not matter in the case of a symmetric multilinear form
because the function is invariant under any permutation of its arguments.

The hypergraph matching problem can be written as the maximization of a third-order score function 
$S$ subject to the assignment constraints,
\begin{eqnarray}
    \max\limits_{\bx\in M} S(\bx) = \max\limits_{\bx\in M} \MF(\bx,\bx,\bx) 
    = \max\limits_{\bx\in M} \sum_{i,j,k=1}^n \FC_{ijk}\, \bx_i\, \bx_j\, \bx_k .
\end{eqnarray}
In \cite{Quynh2015} we propose to solve instead the equivalent problem defined as
\begin{eqnarray}\label{eq:new_problem}
    \max\limits_{\bx\in M} S^4(\bx) = \max\limits_{\bx\in M} \MF^4(\bx,\bx,\bx,\bx) \cblue{\bydef} \sum_{i,j,k,l=1}^n \FC^4_{ijkl} \bx_i \bx_j \bx_k \bx_l
\end{eqnarray} 
where the new symmetric fourth-order tensor $\FC^4$ is obtained by lifting the third
order tensor $\FC$ via 
\begin{equation}\label{eq:lifting_step}
    \FC^4_{ijkl} = \FC_{ijk} + \FC_{ijl} + \FC_{ikl} + \FC_{jkl}.
\end{equation}
The reason for this lifting step is discussed below.
The main idea of \cite{Quynh2015} is then to optimize instead of the score function the associated multilinear form. 
The following theorem establishes equivalence of these problems under the assumption that the score function $S^4$ is convex on $\R^n$.
\begin{theorem}[Nguyen \etal \cite{Quynh2015}] \label{theo:Nguyen15} {
    Let $\FC^4$ be a symmetric fourth-order tensor and suppose the corresponding score function $S^4\colon\R^n \to \R$ is convex. 
    Then it holds for any compact constraint set $D \subset \R^n$,
    \begin{eqnarray*}
    \max\limits_{\bx \in D} \MF^4(\bx,\bx,\bx,\bx)\!\!\!
	    &=& \!\!\!\max\limits_{\bx,\by \in D} \MF^4(\bx,\bx,\by,\by) \\
	    &=& \!\!\!	\max\limits_{\bx,\by,\bz,\bt \in D} \MF^4(\bx,\by,\bz,\bt) .
	  \end{eqnarray*}
    \label{theo:main_order4}}
\end{theorem}
Note that a given score function $S^4$ need not be convex. In \cite{Quynh2015} 
we show that every \cblue{fourth-order} score function can be modified
so that it becomes convex, while solving \eqref{eq:new_problem} for the original and the modified score function turns out to be equivalent.
The maximization of the multilinear form $\MF^4(\bx,\by,\bz,\bt)$ can then be done in two ways. 
Either one fixes all but one or all but two \cgram{arguments} 
and maximizes the remaining arguments over the set of assignment matrices $M$.
As shown in \cite{Quynh2015}, this scheme boils down to solving a sequence of Linear Assignment Problems (LAPs)
or a sequence of (approximate) QAPs which can be done efficiently by existing lower-order methods.

The convexity of the score function is crucial for the proof of Theorem \ref{theo:Nguyen15}. 
The lifting step \eqref{eq:lifting_step} has been proposed in \cite{Quynh2015} as non-trivial third-order score functions are not convex
\crevision{on $\RR^n$} as the following lemma shows. 
\begin{lemma}[Nguyen \etal \cite{Quynh2015}] \label{lem:Nguyen15}
    Let $S$ be a third-order score function defined as in Eq. \eqref{eq:HGM}. 
    If $S$ is not constant zero, then $S:\R^n \to\R$ is not convex.
\end{lemma}
In the following we show that a statement similar to Theorem \ref{theo:Nguyen15} holds also in the \cblue{third-order} case. 
This \cgram{allows us} to propose
algorithms directly for the \cblue{third-order} case without the lifting step, \cgram{leading to two times faster algorithms.}
The new key insight to establish this result is that convexity on whole $\R^n$ is not required. 
If one just requires that the \cblue{third-order} score function is 
convex on the positive orthant,  then Lemma \ref{lem:Nguyen15} is no longer true. 
In fact convexity is only required in a ``pointwise'' fashion which we make precise in Theorem \ref{theo:main}. 
\crevision{
Similar to our previous Theorem \ref{theo:Nguyen15},
we state the main Theorem \ref{theo:main} of this paper for general compact sets $D \subset \R^n$ 
which includes as a special case the set of assignment matrices. 
Although this general form of the theorem may allow generalizations of our proposed framework to other applications, 
in this paper we only focus on the case $D=M$ for the application of hypergraph matching.
}
\begin{theorem}\label{theo:main}
    Let $D\subset\RR^n$ be a compact set,
    $\MF: \RR^n\times\RR^n\times\RR^n \to \RR$ a symmetric third-order multilinear form and 
    \crevision{
    $S: \RR^n \to \RR$
    }
    the associated third-order score function
    such that $S(\bx)=F(\bx,\bx,\bx),$
    then we have the following chain of implications:  \crevision{1) $\Rightarrow$ 2) $\Rightarrow$ 3) $\Rightarrow$ 4).}
    \begin{enumerate}
	\item $\nabla^2 S(\bx)$ is positive semidefinite for all $\bx\in D.$
	\item $\MF(\bx,\by-\bz,\by-\bz) \geq 0$ for all $\bx,\by,\bz \in D.$
	\item 
	It holds for all $\bx,\by,\bz\in D$
	\begin{equation}\label{eq:main_inequality}
	    \MF(\bx,\by,\bz) \leq \max\limits_{\bu\in\set{\bx,\by,\bz}} \MF(\bu,\bu,\bu) .
	\end{equation}
	\item The optimization of the multilinear form is equivalent to the optimization of its associated score function
	$$
	    \!\!\!\!\max\limits_{\bx,\by,\bz\in D} \MF(\bx,\by,\bz) = \max\limits_{\bx,\by\in D} \MF(\bx,\bx,\by) 
	    = \max\limits_{\bx\in D} \MF(\bx,\bx,\bx) .
	$$
    \end{enumerate}
\end{theorem}
As a twice continuously differentiable function
is convex if and only if its Hessian is positive semi-definite on its domain, 
the first requirement is much weaker than requiring convexity on whole $\R^n$. 
Thus Theorem \ref{theo:main} allows us to establish for $D=M$ \cred{inequality \eqref{eq:main_inequality}}
for a nontrivial class of third-order score functions, which
is the key result necessary to establish the equivalence of the maximization of the score function and the maximization of the multilinear form.
\crevision{This is the key requirement to show monotonic ascent for our block-coordinate optimization scheme.}
In Proposition \ref{prop:convexification} we show how to modify an existing \cblue{third-order} score function so that it satisfies condition 1) for $D=M$. 
Again this modification
turns out to be constant on $M$ and thus leads to equivalent optimization problems.
This \cred{altogether} leads to two new algorithms for the hypergraph matching problem \eqref{eq:HGM} 
similar to our previous work \cite{Quynh2015}.
The key idea of both algorithms is to use block-coordinate ascent updates,
where all but one argument or all but two arguments of the multilinear form are fixed and 
one optimizes over the remaining ones in either case.
The inequality \eqref{eq:main_inequality} then allows us to connect the solutions of the two optimization problems.
In both variants, we directly optimize over the discrete set $M$ of possible matchings, that is, there is no relaxation involved.
Moreover, we theoretically prove monotonic ascent for both methods. 
In all our experiments, the proposed third-order methods achieve competitive matching \cblue{score} and matching accuracy
to the fourth-order ones \cite{Quynh2015} while outperforming other state-of-the-art approaches.
On the other hand, we achieve a speed up of factor $2$ over the fourth-order algorithms
\cgram{due to the ability of working directly} with \cblue{the} original tensor.

Compared to \cite{Quynh2015} we avoid the lifting step while maintaining the same theoretical guarantees. 
Moreover, we use the weaker conditions on the score function in Theorem \ref{theo:main} 
to propose two variants of our algorithms based on homotopy methods as presented in Section \ref{sec:adapt_bcagm3}.

\crevision{
\cgram{In general, the multilinear form $\MF$} might not fulfill the conditions of Theorem \ref{theo:main}. 
Therefore, we propose to modify $\MF$ by adding to it some multilinear form $\MG$ 
so that the first statement of Theorem \ref{theo:main} is satisfied for the new function.
At the same time the modification does not affect the solution of our problem \cgram{since}
the added term is constant on the set of assignment matrices.
This will be detailed later on in Proposition \ref{prop:convexification}.
While finding the best possible modification is difficult, one can consider some canonical choices. 
As the first choice, one can implicitly modify $\MF$ by considering a new score function defined as
$\cblue{S_\alpha(\bx)} = S(\bx) + \alpha \norm{\bx}_2^2,$
and then show that there exists some $\alpha_0 \in \RR$ 
such that $\nabla^2 S_\alpha$ is positive semidefinite everywhere in $M$ for all $\alpha \geq \alpha_0.$
However, the  modified score function $S_\alpha$ then becomes in-homogeneous and thus there does not exist any multilinear form $F_\alpha$
such that $\cblue{S_\alpha(\bx)=F_\alpha(\bx,\bx,\bx)}.$ 

A homogeneous modification would be
$\cblue{S_\alpha(\bx)} = S(\bx) + \alpha \norm{\bx}_3^3$. 
Indeed, one can associate to $\norm{\bx}_3^3$ the multilinear form
$\MG(\bx,\by,\bz) = \sum_{i=1}^n \inner{{\be}_i,\bx} \inner{{\be}_i,\by} \inner{{\be}_i,\bz} ,$
where $\be_i$ is the $i$-th standard basis vector of $\RR^n.$ 
Moreover, it holds that $\MG(\bx,\bx,\bx) = \norm{\bx}_3^3.$ In this paper, we consider a family of modifications
which includes the one above by perturbing the set of basis vectors as
$\bar{\be}_i = \eps\ones + (1-\eps)\be_i$ with $\ones$ being the all-ones vector.
Note that $\norm{\bx}_3^3$ is included in this family as a special case for $\eps=0.$
The main idea is that one can later on optimize $\eps$ to find the best lower bound on $\alpha.$
That is to say, we aim at a minimal amount of $S$
such that the conditions from Theorem \ref{theo:main} are satisfied.
As shown in the proof of Proposition \ref{prop:convexification}, $\eps=1/3$ is the optimal value
for which the best bound on $\alpha$ can be obtained.
Thus, we use this value in the rest of the paper.
In particular, let $\bar{\be}_i = \frac{1}{3}\ones + \frac{2}{3}\be_i$ 
we define the symmetric third-order tensor $\GC$ as
}
\begin{equation}\label{eq:mod_tensor}
    \GC = \sum_{i=1}^n \bar{\be}_i \otimes \bar{\be}_i \otimes \bar{\be}_i , 
\end{equation} and the associated multilinear form as
\begin{equation}\label{eq:mod_mf}
	\MG(\bx,\by,\bz) = \sum_{i=1}^n \inner{\bar{\be}_i,\bx} \inner{\bar{\be}_i,\by} \inner{\bar{\be}_i,\bz} .
\end{equation}
\begin{proposition}\label{prop:convexification}
    Let $\MF: \RR^n\times \RR^n \times \RR^n \to \RR$ be a symmetric third-order multilinear form
    and $\MG: \RR^n\times \RR^n \times \RR^n \to \RR$ as defined in Eq. \eqref{eq:mod_mf}.
    We consider the new multilinear form $\MF_\alpha$ defined by 
    $$
	\MF_\alpha(\bx,\by,\bz) = \MF(\bx,\by,\bz) + \alpha \, \MG(\bx,\by,\bz).
    $$
    For all $\alpha \geq \frac{27}{4} \max\limits_{i=1,2,\ldots,n} \sqrt{\sum_{j,k=1}^n (\FC_{ijk})^2 }$ 
    it holds
    \begin{enumerate}
	\item $\MF_\alpha$ is a symmetric third-order multilinear form.
	\item $\nabla^2 S_\alpha(\bx)$ is positive \cred{semidefinite} for all $\bx\in M.$
	\item The new problem is equivalent to the original one
	$$
	    \argmax_{\bx\in M} S_\alpha(\bx) = \argmax_{\bx\in M} S(\bx) .
	$$
    \end{enumerate}
\end{proposition}

\textbf{Discussion}. 
In \cite{Joa2003} a general convexification strategy for
arbitrary score functions has been proposed where, similar to our modification, the added term is constant on the
set of assignment matrices $M$. 
However, as the added term is in-homogeneous, 
it cannot be extended to a symmetric multilinear form and thus does not fit to our framework. 
In \cblue{second-order} graph matching also several methods use convexified score functions
in various ways \cite{ZasEtl2009, ZhoTor2012, ZhoTor2013}. 
However, for none of these methods it is obvious how to extend it to a \cblue{third-order} approach for hypergraph matching. 

Once $\MF_\alpha$ is symmetric and $\nabla^2 S_\alpha(\bx)$ is positive semidefinite at every $\bx$ in $M$, 
it follows from Theorem \ref{theo:main} that
$$
    \max\limits_{\bx,\by,\bz\in M} \MF_\alpha(\bx,\by,\bz) 
    = \max\limits_{\bx\in M} S_\alpha(\bx) 
    \equiv \max\limits_{\bx\in M} S(\bx).
$$
Therefore, block-coordinate ascent schemes that optimize $\MF_\alpha$ over assignment constraints can be derived 
in a similar fashion to \cite{Quynh2015}.
In particular, we propose below two variants, one for solving 
$$
    \max\limits_{\bx,\by,\bz\in M} \MF_\alpha(\bx,\by,\bz)
$$ which leads to Algorithm \ref{algo:bcagm3}
and the other for solving
$$
    \max\limits_{\bx,\by\in M} \MF_\alpha(\bx,\by,\by) 
$$ which leads to Algorithm \ref{algo:bcagm3_psi}.
Both variants come along with theoretical guarantees including strict monotonic ascent and finite convergence.

\subsection{Tensor Block Coordinate Ascent via a Sequence of Linear Assignment Problems}
\label{sec:bcagm3}
The first algorithm uses a block coordinate ascent scheme to optimize the multilinear function $\MF_\alpha(\bx,\by,\bz)$
where two arguments are fixed at each iteration and the function is maximized over the remaining one. 
This boils down to solving a sequence of LAPs, 
which can be solved to global optimality by the Hungarian method \cite{Kuh1955,BurAmiMar2012}.
As we are optimizing the multilinear form, 
we need to guarantee that our algorithm produces finally a homogeneous solution, \ie $\bx=\by=\bz$.
Moreover, we require that our algorithms achieve ascent not only in the multilinear form but also in the original score function.
These are \cgram{the} two main reasons why we use the inequality \eqref{eq:main_inequality} at step 7) of Algorithm \ref{algo:bcagm3}
to get ascent in the score function.
The following theorem summarizes the properties of Algorithm \ref{algo:bcagm3}.

\begin{theorem}\label{theo:bcagm3}
    Let $\FC$ be a symmetric third-order tensor 
    \crevision{and $\alpha\geq\frac{27}{4}\max\limits_{i=1,2,\ldots,n} \sqrt{\sum_{j,k=1}^n (\FC_{ijk})^2 }$ .}
    Then the following holds for Algorithm \ref{algo:bcagm3}:
    \begin{enumerate}
	\item The sequence $\MF_\alpha (\bx^{k},\by^{k},\bz^{k})$ for $k=1,2,\ldots$ is strictly monotonically increasing or terminates.
	\item The sequence of scores $S(\bu^m)$ for $m=1,2,\ldots$ is strictly monotonically increasing or terminates. 
	For every $m$, $\bu^m \in M$ is a valid assignment matrix.
	\item The algorithm terminates after a finite number of iterations.
    \end{enumerate}
\end{theorem}

\begin{algorithm}
\begin{algorithmic}[1]
    \STATE Input: $\alpha = \frac{27}{4}\max\limits_{i=1,2,\ldots,n} \sqrt{\sum_{j,k=1}^n (\FC_{ijk})^2 }, \crevision{(\bx^0,\by^0,\bz^0)\in\RR^n}, \newline k = 0, m=0$
    \STATE Output: $\bx^* \in M$
	  \LOOP
		\STATE $\tbx^{k+1} = \arg\max\limits_{\bx \in M} \MF_\alpha(\bx,\by^{k},\bz^{k})$ 
		\STATE $\tby^{k+1} = \arg\max\limits_{\by \in M} \MF_\alpha(\tbx^{k+1},\by,\bz^{k})$ 
		\STATE $\tbz^{k+1} = \arg\max\limits_{\bz \in M} \MF_\alpha(\tbx^{k+1},\tby^{k+1},\bz)$ 
		
		\IF {$\MF_\alpha (\tbx^{k+1},\tby^{k+1},\tbz^{k+1}) = \MF_\alpha (\bx^{k},\by^{k},\bz^{k})$}
		    \STATE $\bu^{m+1} = \argmax\limits_{\bu \in \{ \tbx^{k+1},\tby^{k+1},\tbz^{k+1} \}} \MF_\alpha(\bu,\bu,\bu)$
		    \IF {$\MF_\alpha (\tbx^{k+1},\tby^{k+1},\tbz^{k+1}) < \MF_\alpha(\bu^{m+1},\bu^{m+1},\bu^{m+1})$} 
			\STATE $\bx^{k+1} = \by^{k+1} = \bz^{k+1} = \bu^{m+1}$
			\STATE $m = m + 1$
		    \ELSE
			\STATE \crevision{return $\bu^{m+1}$}
		    \ENDIF
		\ELSE 
		    \STATE $\bx^{k+1} = \tbx^{k+1}$, $\by^{k+1} = \tby^{k+1}$, $\bz^{k+1}=\tbz^{k+1}$
		\ENDIF
		\STATE $k = k + 1$
	    \ENDLOOP {}
      \caption{BCAGM3} 
      \label{algo:bcagm3}
\end{algorithmic}
\end{algorithm}

We would like to note that all statements of Theorem \ref{theo:bcagm3} remain valid for $\alpha=0$ 
if one of the statements 1)-3) in Theorem \ref{theo:main} holds for $D=M$.
In practice, this condition might be already satisfied for some constructed affinity tensors.
Thus we adopt the strategy in \cite{Quynh2015} 
by first running Algorithm  \ref{algo:bcagm3} with $\alpha=0$ until we get no further ascent 
and only then we set $\alpha = \frac{27}{4} \max\limits_{i=1,2,\ldots,n} \sqrt{\sum_{j,k=1}^n (\FC_{ijk})^2 }.$ 
It turns out that in our experiments often the first phase with $\alpha=0$ 
leads automatically to a homogeneous solution.
In this case, the \cred{algorithm} can be stopped as the following lemma shows.
\begin{lemma}\label{claim}
   Suppose Algorithm \ref{algo:bcagm3} runs with $\alpha=\alpha_0$ for some $\alpha_0\in\RR_+$.
	If $\tbx^{k+1}=\tby^{k+1}=\tbz^{k+1}=:\bu$ holds at some iteration $k$ then 
        for all $\alpha\geq\alpha_0$, it holds
	$$
	    \bu = \argmax\limits_{\bz\in M} \MF_\alpha(\bz,\bu,\bu) .
	$$ 
\end{lemma}
Lemma \ref{claim} shows that we can stop Algorithm \ref{algo:bcagm3}
whenever the first phase leads already to a homogeneous solution as no further ascent is possible.
In particular, there is no need to go for a new phase with larger value of $\alpha$ 
since the current iterate $(\bu,\bu,\bu)$ is already optimal to all local updates of Algorithm \ref{algo:bcagm3},
\ie steps (4)-(6), for all $\alpha\geq\alpha_0=0$.
Also, the fact that the iterate is already homogeneous 
implies that no further possible ascent can be achieved at step (7).

\subsection{Tensor Block Coordinate Ascent via Alternating between Quadratic and Linear Assignment Problems}
\label{sec:bcagm3_psi}
The second algorithm uses a block coordinate ascent scheme to optimize the multilinear form $\MF_\alpha$
where now one resp. two arguments are fixed and we optimize over the remaining ones in an alternating manner.
The resulting scheme alternates between QAPs and LAPs.
While the LAP can be efficiently solved using the Hungarian algorithm as before, 
the QAP is NP-Hard.  Thus a globally optimal solution is out of reach.
However, our algorithm does not require the globally optimal solution \cblue{at each step} in order to maintain the theoretical properties.
It is sufficient that the sub-routine method, called $\Psi$ yields monotonic ascent w.r.t. the current variable, that is,
\begin{equation}\label{subroutine}
    \bz = \Psi(A, \by^k) \quad \textrm{ s.t. } \quad \inner{\bz,A\bz}\geq \inner{\by^k,A\by^k}
\end{equation}
where $\by^k$ is the current iterate and the nonnegative symmetric $A \in \R^{n\times n}$ is 
$\MF_\alpha(\tbx^{k+1},\,\cdot\,,\,\cdot\,)$ in our algorithm.
As in Algorithm \ref{algo:bcagm3}, we go back to the optimization of the score function in step 6) 
using inequality \eqref{eq:main_inequality}. 
The following theorem summarizes the properties of Algorithm \ref{algo:bcagm3_psi}.

\begin{theorem}\label{theo:bcagm3_psi}
    Let $\FC$ be a symmetric \cblue{third-order} tensor
    \crevision{and $\alpha\geq\frac{27}{4}\max\limits_{i=1,2,\ldots,n} \sqrt{\sum_{j,k=1}^n (\FC_{ijk})^2 }$ .}
    Let $\Psi$ be an algorithm for the QAP which yields monotonic ascent w.r.t. the current iterate.
    The following holds for Algorithm \ref{algo:bcagm3_psi}:
    \begin{enumerate}
	  \itemsep0em
	  \item The sequence $\MF_\alpha (\bx^{k},\by^{k},\by^{k})$ for $k=1,2,\ldots$ 
	  is strictly monotonically increasing or terminates.
	  \item The sequence of scores $S(\bu^m)$ for $m=1,2,\ldots$ is strictly monotonically increasing or terminates. 
		For every $m$, $\bu^m \in M$ is a valid assignment matrix.
    	  \item The algorithm terminates after a finite number of iterations.
        \end{enumerate}
    \label{theo:quadratic}
\end{theorem}

\begin{algorithm}
\begin{algorithmic}[1]
    \STATE Input: $\alpha = \frac{27}{4}\max\limits_{i=1,2,\ldots,n} \sqrt{\sum_{j,k=1}^n (\FC_{ijk})^2 }, \crevision{(\bx^0,\by^0)\in\RR^n}, \newline k = 0, m=0$, 
    \\ $\bz=\Psi(A,\by^k)$ is a sub-routine method that provides monotonic ascent for the QAP, 
    \ie $\inner{\bz,A\bz}\geq \inner{\by^k,A\by^k}$
    \STATE Output: $\bx^* \in M$
	  \LOOP
		\STATE $\tbx^{k+1} = \arg\max\limits_{\bx \in M} \MF_\alpha(\bx,\by^{k},\by^{k})$ 
		\STATE $\tby^{k+1} = \Psi\left( \MF_\alpha(\tbx^{k+1},\,\cdot\,,\,\cdot\,), \by^k \right)$ 
		
		\IF {$\MF_\alpha (\tbx^{k+1},\tby^{k+1},\tby^{k+1}) = \MF_\alpha (\bx^{k},\by^{k},\by^{k})$}
		    \STATE $\bu^{m+1} = \argmax\limits_{\bu \in \{ \tbx^{k+1},\tby^{k+1} \}} \MF_\alpha(\bu,\bu,\bu)$
		    \IF {$\MF_\alpha (\tbx^{k+1},\tby^{k+1},\tby^{k+1}) < \MF_\alpha(\bu^{m+1},\bu^{m+1},\bu^{m+1})$} 
			\STATE $\bx^{k+1} = \by^{k+1} = \bu^{m+1}$
			\STATE $m = m + 1$
		    \ELSE
			\STATE \crevision{return $\bu^{m+1}$}
		    \ENDIF
		\ELSE 
		    \STATE $\bx^{k+1} = \tbx^{k+1}$, $\by^{k+1}=\tby^{k+1}$
		\ENDIF
		\STATE $k = k + 1$
	    \ENDLOOP {}
      \caption{BCAGM3+$\Psi$} 
      \label{algo:bcagm3_psi}
\end{algorithmic}
\end{algorithm}

In analogy to Theorem \ref{theo:bcagm3}, all statements of Theorem \ref{theo:bcagm3_psi} remain valid for $\alpha=0$ 
if one of the statements 1)-3) in Theorem \ref{theo:main} holds for $D=M$..
Thus we use the same initialization strategy with $\alpha=0$ as described above.
There are several methods available which we could use for the sub-routine $\Psi$ in the algorithm. 
We decided to use the recent max pooling algorithm \cite{ChoEtAl2014} and the IPFP algorithm \cite{LeoHebSuk2009},
and \cgram{then} use the Hungarian algorithm to turn their output into a valid assignment matrix in $M$ at each step. 
It turns out that the combination of our tensor block coordinate ascent scheme using
their algorithms as a sub-routine yields very good performance on all datasets.
In case of Algorithm \ref{algo:bcagm3_psi} a statement as in Lemma \ref{claim} is not possible,
as the subroutine $\Psi$ usually can only deliver a local solution to the subproblem and Lemma \ref{claim}
relies on the fact that the subproblem can be solved \cblue{globally optimally}.
However, we observe in our experiments that almost always Algorithm \ref{algo:bcagm3_psi} 
does not achieve further ascent when its iterate is homogeneous, 
thus we recommend to stop \cred{Algorithm \ref{algo:bcagm3_psi}} in this case.

\subsection{A Homotopy Tensor Block Coordinate Ascent Scheme}
\label{sec:adapt_bcagm3}
Both algorithms consist of two phases -- the first phase uses $\alpha=0$ as initialization
and the second phase uses $\alpha = \frac{27}{4} \max\limits_{i=1,2,\ldots,n} \sqrt{\sum_{j,k=1}^n (\FC_{ijk})^2 }.$
However, if the value of $\alpha$ is too large for the second phase, often no further improvement is achieved.
This phenomenon can be explained by the fact that 
a large modification term tends to homogenize the variables more quickly, \ie ~$\bx^k=\by^k=\bz^k$,
which makes the algorithm get stuck faster at a critical point.

To fix this, one first observes that the inequality \eqref{eq:main_inequality} lies at the heart of both algorithms.
On one hand, it serves as a sufficient condition for the equivalence of maximizing score functions and multilinear form.
On the other hand, it might help the algorithms to jump to better solutions in case they reach a stationary state.
Both methods guarantee this inequality to be satisfied for all the tuples $\set{\bx,\by,\bz}\in M$ at the same time
by using a rather large value of $\alpha.$
However, this might be too conservative as the algorithm itself converges typically rather quickly and thus visits only
a small number of feasible tuples.
Thus, we propose below to satisfy the inequality \eqref{eq:main_inequality} as the algorithm proceeds.
This is done by updating $\alpha$ accordingly during the algorithm which yields a homotopy method.

\begin{proposition}\label{prop:adapt}
    Let $\MF: \RR^n\times \RR^n \times \RR^n \to \RR$ be a symmetric third-order multilinear form
    and $\MG: \RR^n\times \RR^n \times \RR^n \to \RR$ as defined in Eq. \eqref{eq:mod_mf}.
    For each $\alpha \in \RR$, let $\MF_\alpha$ be defined as
    \begin{equation*}
	\MF_\alpha(\bx,\by,\bz) = \MF(\bx,\by,\bz) + \alpha\,\MG(\bx,\by,\bz) . 
    \end{equation*}
    Then the following holds 
    \begin{enumerate}
	\item $\MF_\alpha$ is a symmetric multilinear form.
	\item 
	For all non-homogeneous tuple $(\bx,\by,\bz) \in M\, (\ie \bx\neq\by \textrm{ or } \by\neq\bz \textrm{ or } \bz\neq\bx),$ 
	the following inequality holds 
	\begin{equation}\label{eq:adapt_inequality}
	    \MF_\alpha(\bx,\by,\bz) \leq \max\limits_{\bu\in\set{\bx,\by,\bz}} \MF_\alpha(\bu,\bu,\bu)
	\end{equation}
	if and only if
	$$
	    \alpha \geq \Lambda(\bx,\by,\bz) \bydef \frac{ \MF(\bx,\by,\bz) - \max\limits_{\bu\in\set{\bx,\by,\bz}} \MF(\bu,\bu,\bu) } 
	    { \MG(\bx,\bx,\bx) - \MG(\bx,\by,\bz) } .
	$$
	\item For every $\alpha \in \RR$, it holds
	$$
	    \argmax\limits_{\bx\in M} S_\alpha(\bx) = \argmax\limits_{\bx\in M} S(\bx) .
	$$
    \end{enumerate}
\end{proposition}
Proposition \ref{prop:adapt} suggests that if $\alpha$ is chosen such that
\begin{equation}\label{eq:minimum_bound}
    \alpha \geq \max\limits_{\substack{\bx,\by,\bz\in M \\ \bx\neq\by \textrm{ or } \by\neq\bz \textrm{ or } \bz\neq\bx}} 
    \Lambda(\bx,\by,\bz) ,
\end{equation}
then the inequality \eqref{eq:main_inequality} is satisfied for all $\bx,\by,\bz \in M$,
for which, it follows from Theorem \ref{theo:main}
$$
    \max\limits_{\bx,\by,\bz\in M} \MF_\alpha(\bx,\by,\bz) = \max\limits_{\bx\in M} S_\alpha(\bx) 
    \equiv \max\limits_{\bx\in M} S(\bx) .
$$
Thus, Algorithm \ref{algo:bcagm3} or \ref{algo:bcagm3_psi} can be applied again to solve the problem.
However, the computation of the optimal bound given in \eqref{eq:minimum_bound} 
is not feasible as the number of feasible tuples grows exponentially with the size of the problem.
Therefore, we adaptively update $\alpha$ as follows.
First, we use block coordinate ascent steps as in the previous methods with $\alpha=0$.
When such a step achieves no further ascent and inequality \eqref{eq:main_inequality} is violated,
we slightly increase $\alpha$ so that the inequality becomes satisfied for the current iterate,
and the algorithm is continued.
The whole scheme is shown in Algorithm \ref{algo:adapt_bcagm3}.
Note that a small $\xi>0$ is used in step 13) to prove the convergence, which
is analyzed below.

\begin{algorithm}
\begin{algorithmic}[1]
    \STATE Input: $\alpha = 0, \crevision{(\bx^0,\by^0,\bz^0)\in\RR^n}, k = 0, m=0, \xi>0$
    \STATE Output: $\bx^* \in M$
	  \LOOP
		\STATE $\tbx^{k+1} = \arg\max\limits_{\bx \in M} \MF_\alpha(\bx,\by^{k},\bz^{k})$ 
		\STATE $\tby^{k+1} = \arg\max\limits_{\by \in M} \MF_\alpha(\tbx^{k+1},\by,\bz^{k})$ 
		\STATE $\tbz^{k+1} = \arg\max\limits_{\bz \in M} \MF_\alpha(\tbx^{k+1},\tby^{k+1},\bz)$ 
		
		\IF {$\MF_\alpha (\tbx^{k+1},\tby^{k+1},\tbz^{k+1}) = \MF_\alpha (\bx^{k},\by^{k},\bz^{k})$}
		    \STATE $\bu^{m+1} = \argmax\limits_{\bu \in \{ \tbx^{k+1},\tby^{k+1},\tbz^{k+1} \}} \MF_\alpha(\bu,\bu,\bu)$
		    \IF {$\MF_\alpha (\tbx^{k+1},\tby^{k+1},\tbz^{k+1}) < \MF_\alpha(\bu^{m+1},\bu^{m+1},\bu^{m+1})$} 
			\STATE $\bx^{k+1} = \by^{k+1} = \bz^{k+1} = \bu^{m+1}$
			\STATE $m = m + 1$
		    \ELSIF {\crevision{$\alpha$ does not satisfy Eq. \eqref{eq:minimum_bound}}}
			\STATE $\alpha = \Lambda(\tbx^{k+1},\tby^{k+1},\tbz^{k+1}) + \xi$
			\STATE $\bx^{k+1} = \tbx^{k+1}$, $\by^{k+1}=\tby^{k+1}$, $\bz^{k+1}=\tbz^{k+1}$
		    \ELSE 
			\crevision{\STATE \crevision{return $\bu^{m+1}$}}
		    \ENDIF
		\ELSE 
		    \STATE $\bx^{k+1} = \tbx^{k+1}$, $\by^{k+1}=\tby^{k+1}$, $\bz^{k+1}=\tbz^{k+1}$
		\ENDIF
		\STATE $k = k + 1$
	    \ENDLOOP {}
      \caption{Adapt-BCAGM3} 
      \label{algo:adapt_bcagm3}
\end{algorithmic}
\end{algorithm}

Note that $\alpha$ is always strictly increasing whenever Algorithm \ref{algo:adapt_bcagm3} enters step 12).
This can be seen as follows. 
If the current value of $\alpha$ is greater than or equal to $\Lambda(\tbx^{k+1},\tby^{k+1},\tbz^{k+1})$, 
then inequality \eqref{eq:adapt_inequality} holds for the current tuple $\set{\tbx^{k+1},\tby^{k+1},\tbz^{k+1}}$, 
which implies the algorithm should not have entered step 12), leading to a contradiction.

For any fixed value of $\alpha$, 
Algorithm \ref{algo:adapt_bcagm3} works exactly in the same fashion as Algorithm \ref{algo:bcagm3},
thus, it also yields strict monotonic ascent.
The only difference is that the first algorithm updates $\alpha$ only with a potentially large value,
whereas the new one splits this update into multiple phases.

In case $\alpha$ exceeds the bound given in \eqref{eq:minimum_bound} at some iteration,
the ``strict'' inequality will hold for all $\bx,\by,\bz \in M$, in which case $\alpha$ will no longer be updated.
\cgram{This}, combined with the fact that the algorithm yields strict monotonic ascent for fixed $\alpha$ 
and $\MF_\alpha$ is bounded from above, guarantees the whole scheme converges in a finite number of iterations. 


It should be emphasized that for any value of $\alpha$, 
the optimization of $S_\alpha(\bx)$  and $S(\bx)$ on $M$ is always equivalent, but  the equivalence 
of the optimization of the multilinear form $\MF_\alpha(\bx,\by,\bz)$ 
and that of the score function $S_\alpha(\bx)$ is only guaranteed when $\alpha$ satisfies Eq. \eqref{eq:minimum_bound}.

Finally, we adopt the same homotopy approach as described above for Algorithm \ref{algo:bcagm3_psi}
to obtain Algorithm \ref{algo:adapt_bcagm3_psi}.
All the properties of Algorithm \ref{algo:adapt_bcagm3} transfer to Algorithm \ref{algo:adapt_bcagm3_psi}.
In particular, the algorithm achieves strict monotonic ascent within each phase and has finite termination.

\begin{algorithm}
\begin{algorithmic}[1]
    \STATE Input: $\alpha = 0, \crevision{(\bx^0,\by^0)\in\RR^n}, k = 0, m=0, \xi>0$
    \STATE Output: $\bx^* \in M$
	  \LOOP
		\STATE $\tbx^{k+1} = \arg\max\limits_{\bx \in M} \MF_\alpha(\bx,\by^{k},\by^{k})$ 
		\STATE $\tby^{k+1} = \Psi\left( \MF_\alpha(\tbx^{k+1},\,\cdot\,,\,\cdot\,), \by^k \right)$ 
		
		\IF {$\MF_\alpha (\tbx^{k+1},\tby^{k+1},\tby^{k+1}) = \MF_\alpha (\bx^{k},\by^{k},\by^{k})$}
		    \STATE $\bu^{m+1} = \argmax\limits_{\bu \in \{ \tbx^{k+1},\tby^{k+1} \}} \MF_\alpha(\bu,\bu,\bu)$
		    \IF {$\MF_\alpha (\tbx^{k+1},\tby^{k+1},\tby^{k+1}) < \MF_\alpha(\bu^{m+1},\bu^{m+1},\bu^{m+1})$} 
			\STATE $\bx^{k+1} = \by^{k+1} = \bu^{m+1}$
			\STATE $m = m + 1$
		    \ELSIF {\crevision{$\alpha$ does not satisfy Eq. \eqref{eq:minimum_bound}}}
			\STATE $\alpha = \Lambda(\tbx^{k+1},\tby^{k+1},\tby^{k+1}) + \xi$
			\STATE $\bx^{k+1} = \tbx^{k+1}$, $\by^{k+1}=\tby^{k+1}$
		    \ELSE 
			\crevision{\STATE \crevision{return $\bu^{m+1}$}}
		    \ENDIF
		\ELSE 
		    \STATE $\bx^{k+1} = \tbx^{k+1}$, $\by^{k+1}=\tby^{k+1}$
		\ENDIF
		\STATE $k = k + 1$
	    \ENDLOOP {}
      \caption{Adapt-BCAGM3+$\Psi$} 
      \label{algo:adapt_bcagm3_psi}
\end{algorithmic}
\end{algorithm}

\section{Computational Analysis}
\label{sec:complexity}
Even though all our algorithms are shown to converge in a finite number of iterations, 
it is difficult to bound the number of iterations. 
Therefore, we focus on analyzing the cost of each iteration for both methods.

For Algorithm \ref{algo:bcagm3}, each iteration requires solving three linear assignment problems, 
in which the gradient of the multilinear form needs to be computed, \eg , the iterate
$$
    \tbx^{k+1} = \arg\max\limits_{\bx \in M} \MF_\alpha(\bx,\by^{k},\bz^{k})
    = \arg\max\limits_{\bx \in M} \inner{\bx, \MF_\alpha(\,\cdot\,,\by^{k},\bz^{k}) }
$$
requires to compute the vector
$$
    \MF_\alpha(\,\cdot\,,\by^{k},\bz^{k})_r = \sum_{s,t=1}^n \FC_{rst} \, \by^k_s \, \bz^k_t, \quad \forall\, r = 1,2,\ldots,n.
$$
Without any knowledge of $\by^k$ and $\bz^k$, this computation can be done efficiently 
by passing through all the entries of $\FC$ once, \ie in $O(T)$ 
where $T$ is the number of nonzero entries of the tensor \cite{DucEtAl2011}. 
However, this approach potentially considers many unnecessary elements,
for instance, those entries indexed by $(r,s,t)$ for which $\by^k_s=0$ or $\by^k_t=0$ do not contribute anything to the final result,
thus, should be avoided during computation.
One special property of our algorithm is that it always delivers binary assignment matrices at each step,
thus, the sparsity of the variables can be taken into account at each step.
As $\by^k$ and $\bz^k$ are binary assignment matrices, they have exactly $n_1$ nonzero values. 
\cgram{Therefore}, we first go over the positions of all nonzero components in each variable.
In particular, for each pair $(s, t)$ such that 
$\by^k_s > 0$ and $\bz^k_t > 0$, we can access all the related entries $\FC_{rst}$ using a linked list or hashing method.
With that, the above computation can be done in $O\left( \min\set{T, n_1^2 K} \right)$, 
where $K \ll n=n_1n_2$ is the average number of tensor entries that are associated to each pair $(s,t)$.

For solving the LAP, we use the Hungarian method\footnote{We employ the implementation of Cyrill Stachniss due to its simplicity and efficiency: 
\url{http://www2.informatik.uni-freiburg.de/~stachnis/misc.html}}.
The Hungarian method for the  LAP has cubic complexity $O\left( \max\set{n_1, n_2}^3 \right).$
Therefore, the total complexity of each iteration is $O\left(\max\set{n_1, n_2}^3 + \min\set{T, n_1^2 K} \right)$.
For $K = \max\set{n_1, n_2}$ and $n_1^2 K < T$, 
the total complexity becomes $O\left( \max\set{n_1, n_2}^3 \right)$, 
which is roughly the same as the complexity of the Hungarian method.

For Algorithm \ref{algo:bcagm3_psi}, each iteration consists of one LAP and one QAP.
The complexity for solving LAP is the same as above, that is, $O\left(\max\set{n_1, n_2}^3 + \min\set{T, n_1^2 K} \right).$
For the QAP, the iterate
\begin{eqnarray*}
    \tby^{k+1} 	&=& \arg\max\limits_{\by \in M} \MF_\alpha(\tbx^{k+1},\by,\by) \\
		&=& \arg\max\limits_{\by \in M} \inner{\by, \MF_\alpha(\tbx^{k+1},\,\cdot\,,\,\cdot\,) \by} 
\end{eqnarray*}
requires to compute the matrix 
$$
    \MF_\alpha(\tbx^{k+1},\,\cdot\,,\,\cdot\,)_{st} = \sum_{r=1}^n \FC_{rst} \, \tbx^{k+1}_r, \quad \forall\, s,t=1,2,\ldots,n.
$$
Adopting the same technique as before, we first pass through all the nonzero components of $\tbx^{k+1}$,
and at each index $r$ where $\tbx^{k+1}_r > 0$, 
we access all the associated tensor entries $\FC_{rst}$ via the use of linked list. 
This costs $O\left( \min\set{T, n_1 L} \right)$, 
where $L \ll (n_1n_2)^2$ is the number of nonzero tensor entries associated to each index $r$ on average.
Given the above matrix, let us assume our quadratic solvers need $t$ iterations to solve the problem, 
where each iteration costs approximately the same as the computation of the matrix\footnote{This assumption is reasonable for those quadratic solvers that involve only matrix-vector multiplications at each step.}.
Then the computation of $\tby^{k+1}$ takes $O\left( t \min\set{T, n_1 L} \right)$.
The total complexity of each iteration is thus the sum of two complexities, one for solving LAP and the other for solving QAP,
that is, $O(\max\set{n_1, n_2}^3 + \min\set{T, n_1^2 K} + t \min\set{T, n_1 L}).$
For $L=\max\set{n_1, n_2}^2$ and $n_1 L < T$, 
the whole complexity is simplified to $O(t \max\set{n_1, n_2}^3)$,
which is equivalent to solving a sequence of LAPs using the Hungarian method.
Therefore, we observe in our experiments that 
often Algorithm \ref{algo:bcagm3_psi} is quite slower than Algorithm \ref{algo:bcagm3}
since the later only needs to solve three LAPs at each step. 

Finally, the cost per iteration of Algorithm \ref{algo:adapt_bcagm3} and Algorithm \ref{algo:adapt_bcagm3_psi}
is the same as the cost per iteration of Algorithm \ref{algo:bcagm3} and Algorithm \ref{algo:bcagm3_psi} respectively.

\section{Experiments}\label{sec:exp}
This section reports experimental results on standard benchmark datasets (one synthetic and three real)
and compares our algorithms with state-of-the-art hypergraph matching as well as several graph matching methods.

\textbf{Hypergraph Matching Methods}
Tensor Matching (TM) \cite{DucEtAl2011}, Hypergraph Matching via Reweighted Random Walks (RRWHM) \cite{LeeChoLee2011}
and probabilistic hypergraph matching method (HGM) \cite{ZasSha2008}. 

\textbf{Graph Matching Methods}
Max Pooling Matching (MPM) \cite{ChoEtAl2014} and Integer Projected Fixed Point (IPFP) \cite{LeoHebSuk2009},
Reweighted Random Walks for Graph Matching (RRWM) \cite{ChoLeeLee2010} 
and Spectral Matching (SM) \cite{LeoHeb2005}.
Since HGM, RRWM and SM are outperformed by other methods  \cite{Quynh2015, ChoEtAl2014, LeeChoLee2011},
we do not show their performance for saving space.

We denote our  Algorithm \ref{algo:bcagm3} as BCAGM3 which 
uses the Hungarian method at each iteration, and  Algorithm \ref{algo:bcagm3_psi}  as BCAGM3+MP resp. BCAGM3+IPFP
which uses MPM \cite{ChoEtAl2014} and IPFP \cite{LeoHebSuk2009} respectively as subroutine.
MPM has recently outperformed other graph matching algorithms in the presence of a large number of outliers. 
Our homotopy algorithms will be denoted as Adapt-BCAGM3  (Algorithm \ref{algo:adapt_bcagm3})
and Adapt-BCAGM3+MP and Adapt-BCAGM3+IPFP  (Algorithm \ref{algo:adapt_bcagm3_psi}).
We also compare our third-order methods with the fourth-order counterparts from \cite{Quynh2015}
that we denote as BCAGM, BCAGM+MP and BCAGM+IPFP accordingly.
Note that we differentiate between the proposed third-order algorithms and their fourth-order counterparts in \cite{Quynh2015}
by adding a $3$ in their names. For all the algorithms, we use the authors' original implementation.

In the experiments below, we use the all-ones vector as a starting point for all the algorithms
\crevision{as done in the previous work \cite{DucEtAl2011, ChoEtAl2014, LeeChoLee2011} .
}
Moreover, we use the Hungarian method to turn the output of all algorithms into a proper matching.

\textbf{Generation of affinity tensor/matrix:}

To build the affinity tensor for the \cblue{third-order} algorithms, we follow the approach of Duchenne \etal \cite{DucEtAl2011}.
For each triple of points we compute a feature vector $f$ 
from the angles of the triangle formed by those three points as illustrated in Fig. \ref{fig:tensor_construction}.  
Let $f_{i_1,i_2,i_3}$ and $f_{j_1,j_2,j_3}$ denote the feature vectors of two triples $(i_1,i_2,i_3)$ and $(j_1,j_2,j_3)$,
we compute the \cblue{third-order} affinity tensor as:
\begin{equation}
    \FC_{(i_1,j_1),(i_2,j_2),(i_3,j_3)} = \exp(-\gamma \norm{f_{i_1,i_2,i_3} - f_{j_1,j_2,j_3}}_2^2)
\end{equation} where $\gamma$ is the inverse of the mean of all squared distances.
As shown by Duchenne \etal \cite{DucEtAl2011} 
this higher-order affinity \cblue{is} more efficient than pairwise distances 
which have been used in previous graph matching methods \cite{LeoHeb2005, ShiEtal2007, LeoHebSuk2009, ZhoTor2012}.
In particular, it can deal with scale \cgram{transformations} that is not possible for affinities based on distances.
Moreover, more complicated higher-order affinities can also be derived to achieve (partial) 
invariance with respect to other transformations 
such as affine and projective transformations \cite{DucEtAl2011}.
This explains why \cblue{higher-order} approaches lead typically to superior results compared to second-order approaches.
However, one should take into account the computational cost of constructing the affinity tensor.
Given two images with $n_1$ and $n_2$ points respectively, then the number of candidate correspondences is $n=n_1 n_2$.
Thus a third-order affinity tensor consists of $\binom{n^3}{3}$ entries, which is not only infeasible to compute in reasonable time 
but also requires huge memory resources even for moderate $n_1$ and $n_2$.
Therefore we consider only sparse tensors as in \cite{DucEtAl2011, LeeChoLee2011}.
In particular, we randomly sample $n^2$ triples of points from the first point set, 
and find for each of them $k$ nearest neighbors in the second point set using approximate nearest neighbor search algorithm 
\cite{AryEtal}. Each pair of \cgram{tuples} then determines a non-zero entry of the tensor to be calculated
while the remaining entries are zero.
In principle, a large value of $k$ would increase the chance that correct matches are captured by these pairs of tuples
which potentially improves the result, however, this comes at additional searching cost.
Therefore, we empirically set $k=300$ in all our experiments to achieve reasonable running time for all the methods.

For \cblue{second-order} methods, we construct the affinity matrix by following previous work \cite{ChoEtAl2014, ChoLeeLee2010},
which estimates the pairwise similarity as
\begin{equation}
    \exp(- {| d_{i_1 i_2}^P - d_{j_1 j_2}^Q |}^2 / \sigma_s^2)
\end{equation} 
where $d_{i_1 i_2}^P$ and $d_{j_1 j_2}^Q$ are the Euclidean distances between two points $i_1,\,i_2\in P$
and $j_1,j_2\in Q$ respectively, $\sigma_s$ is a normalization term that is specified later in each experiment.

\begin{figure}[!t]
    \centering
    \subfloat{\includegraphics[width=1.01in]{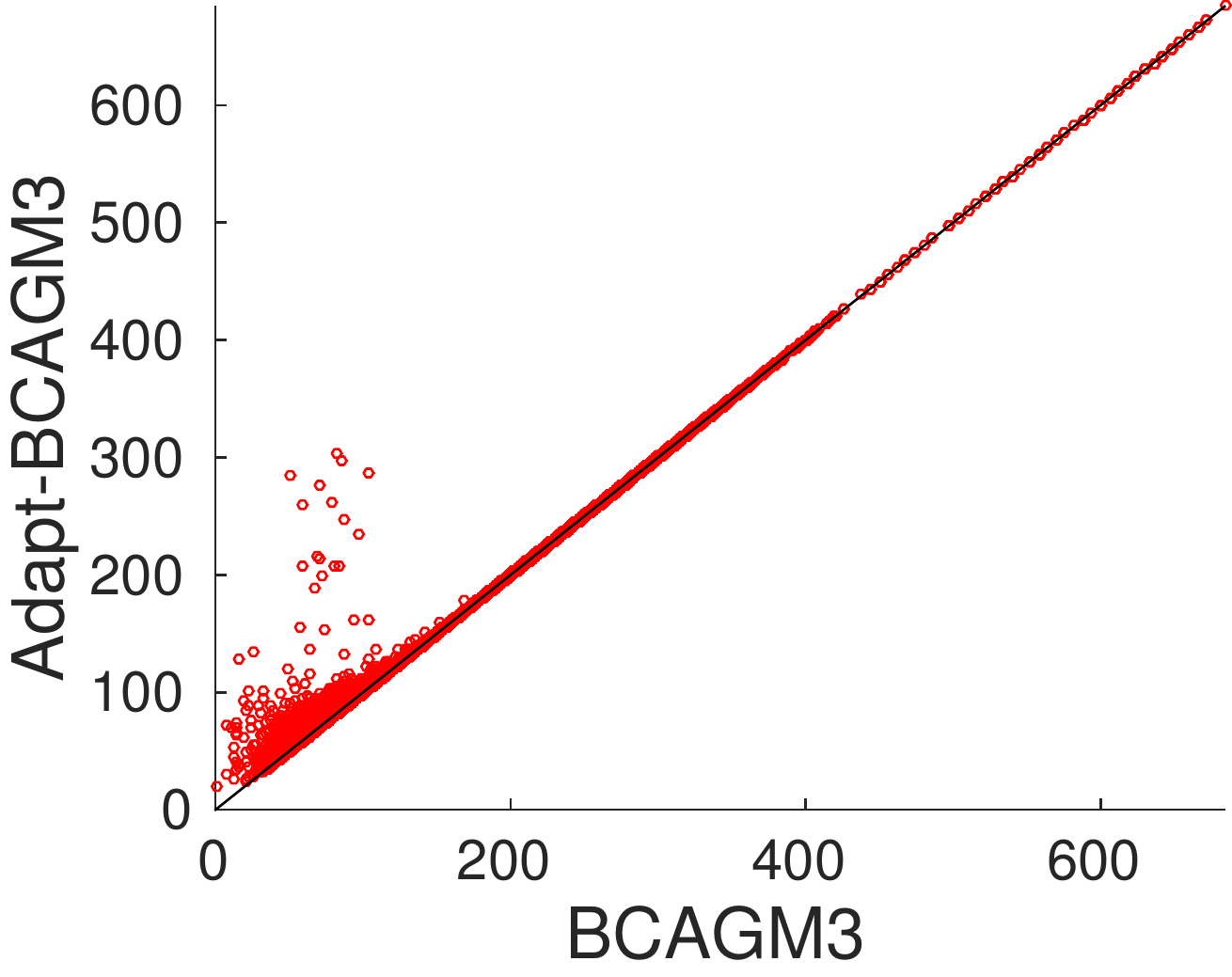}}
    \setcounter{subfigure}{0}
    \subfloat[Face]{\includegraphics[width=1.01in]{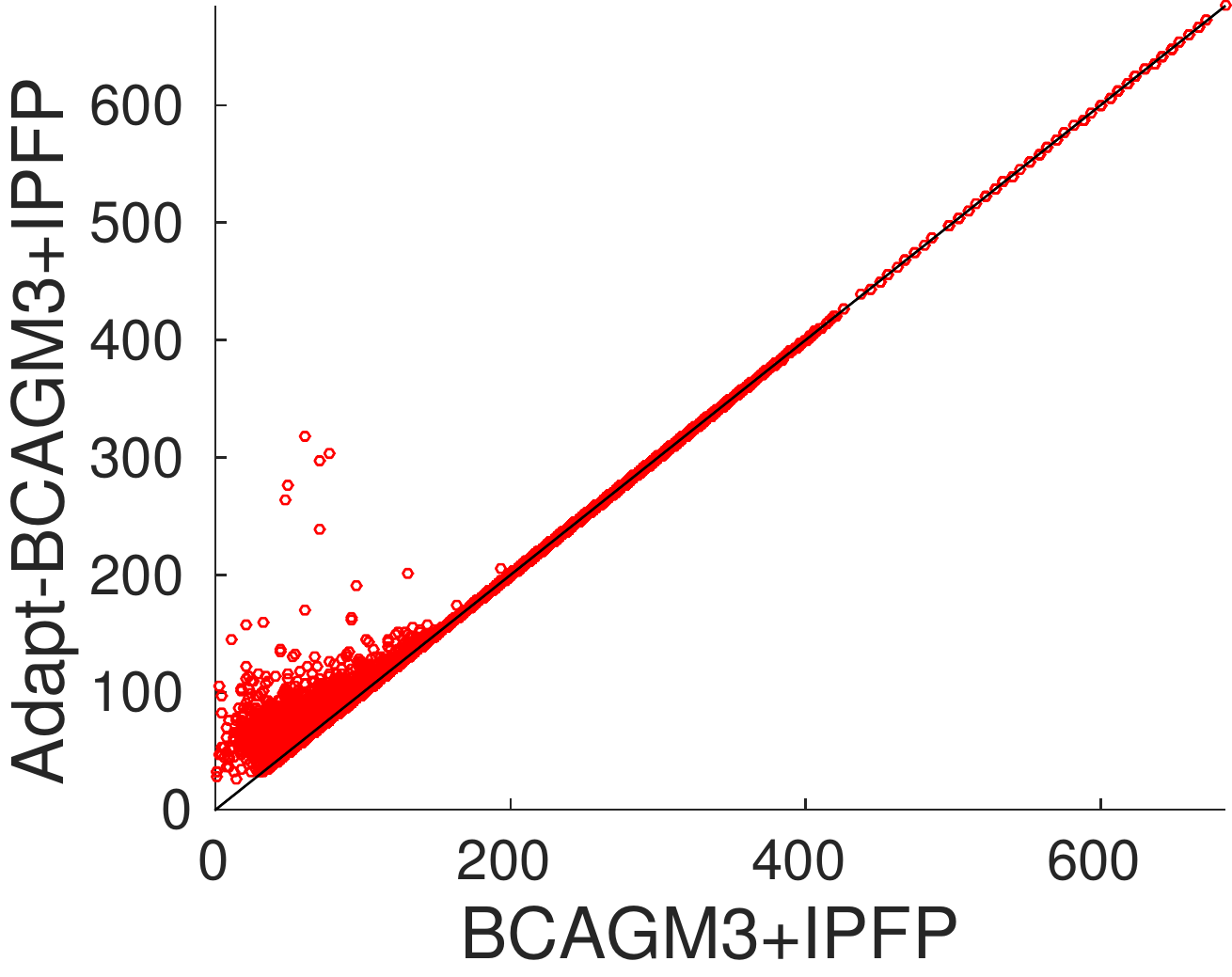}}
    \subfloat{\includegraphics[width=1.01in]{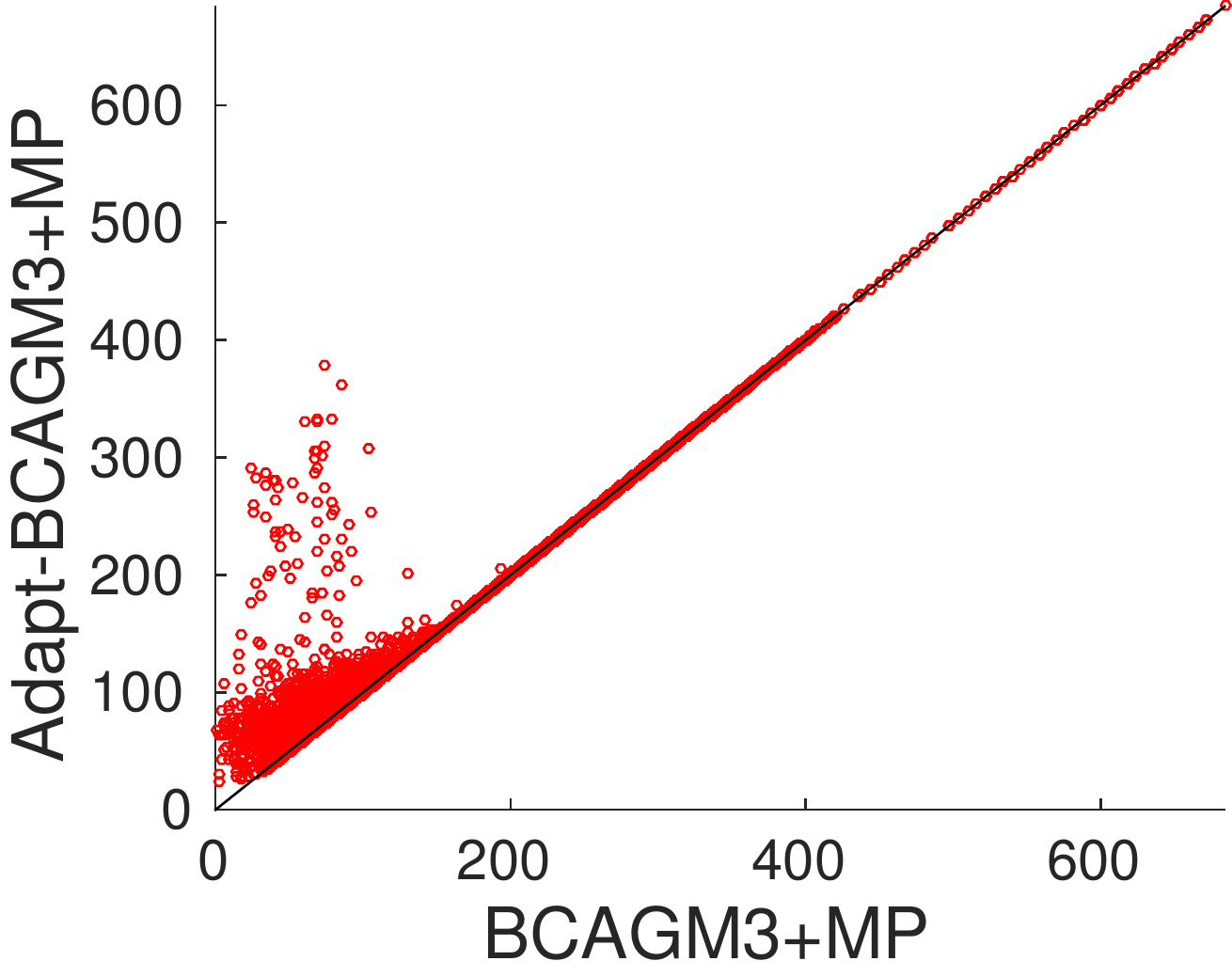}} 
    \\
    \setcounter{subfigure}{0}
    \subfloat{\includegraphics[width=1.01in]{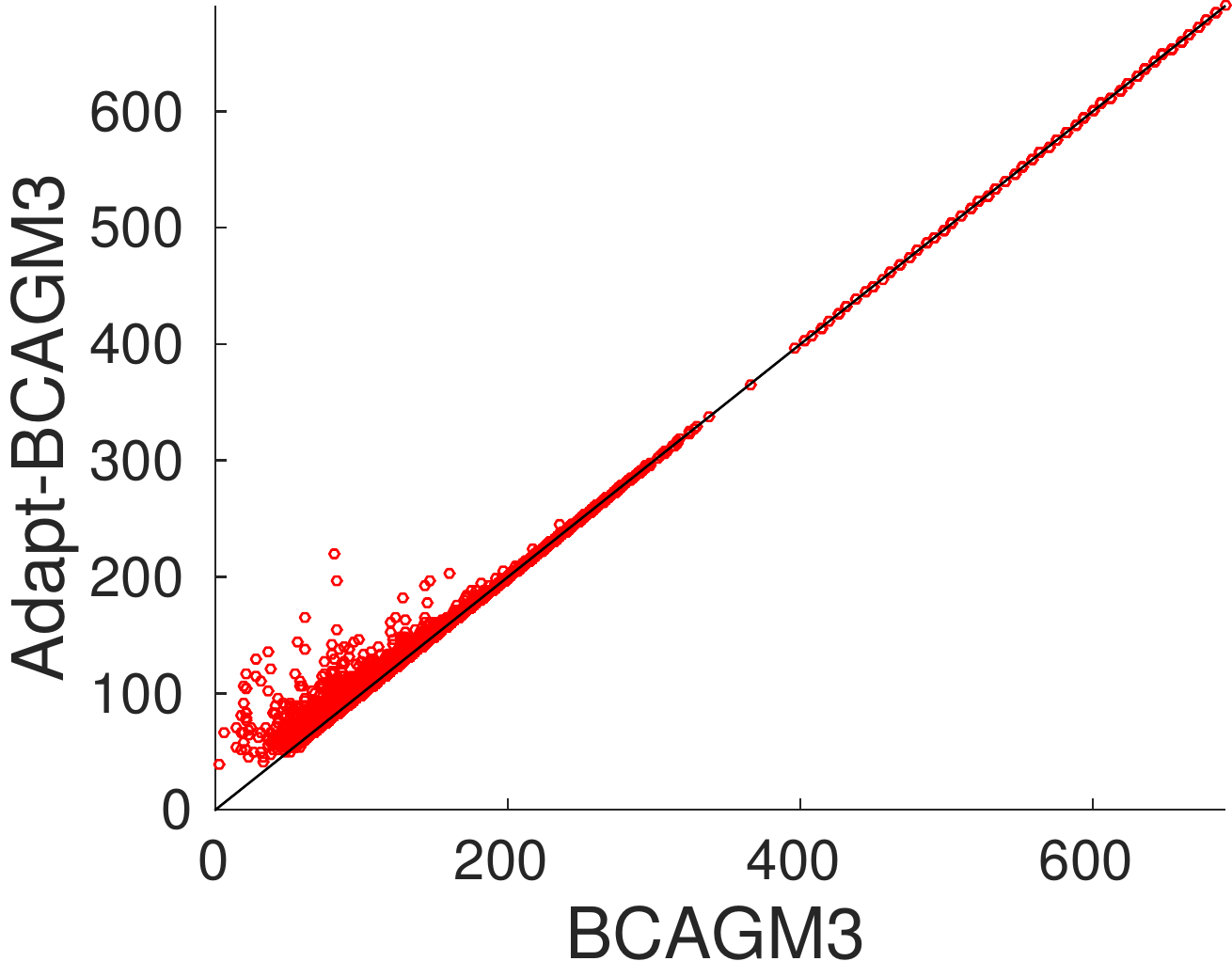}}
    \subfloat[Duck]{\includegraphics[width=1.01in]{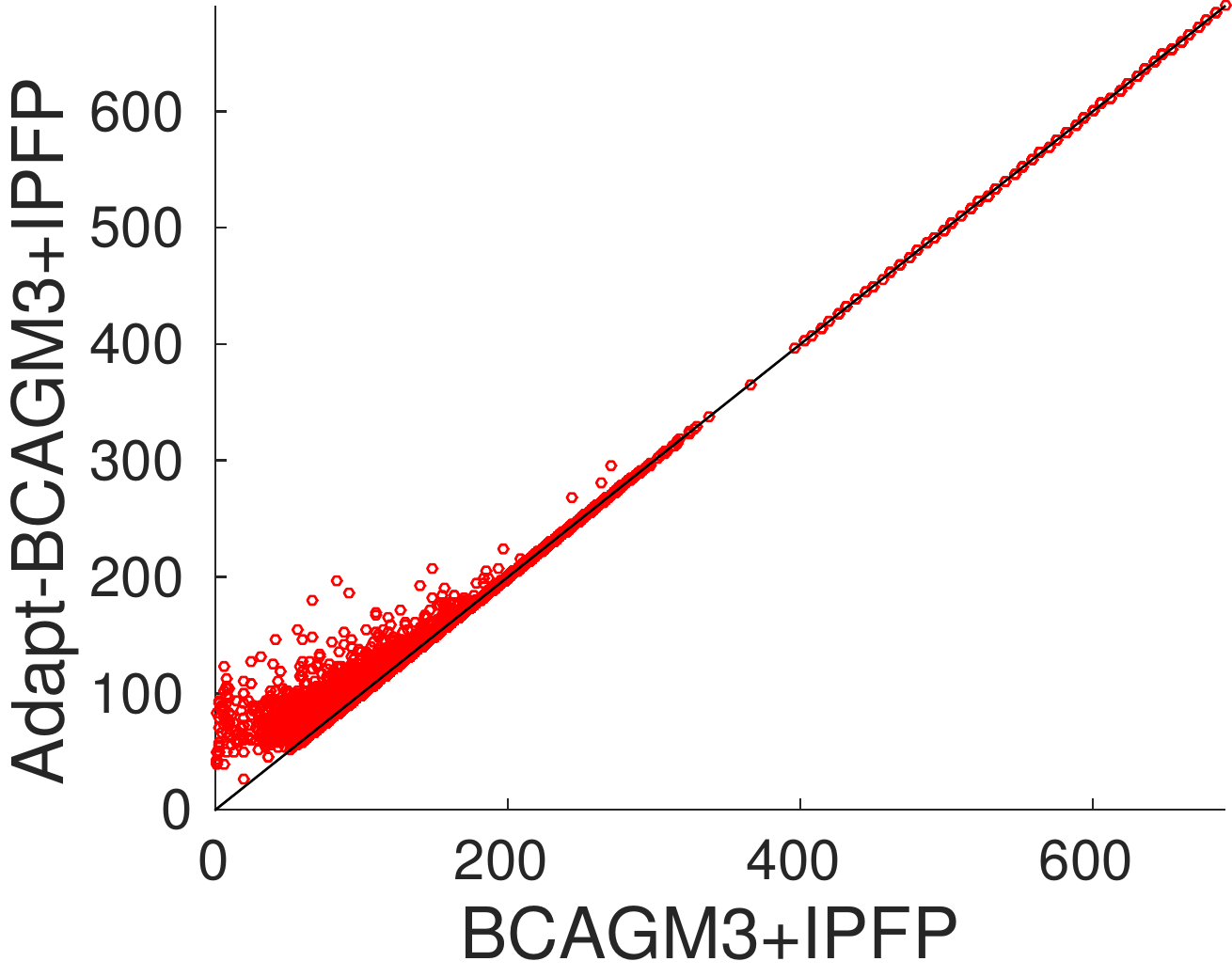}}
    \setcounter{subfigure}{0}
    \subfloat{\includegraphics[width=1.01in]{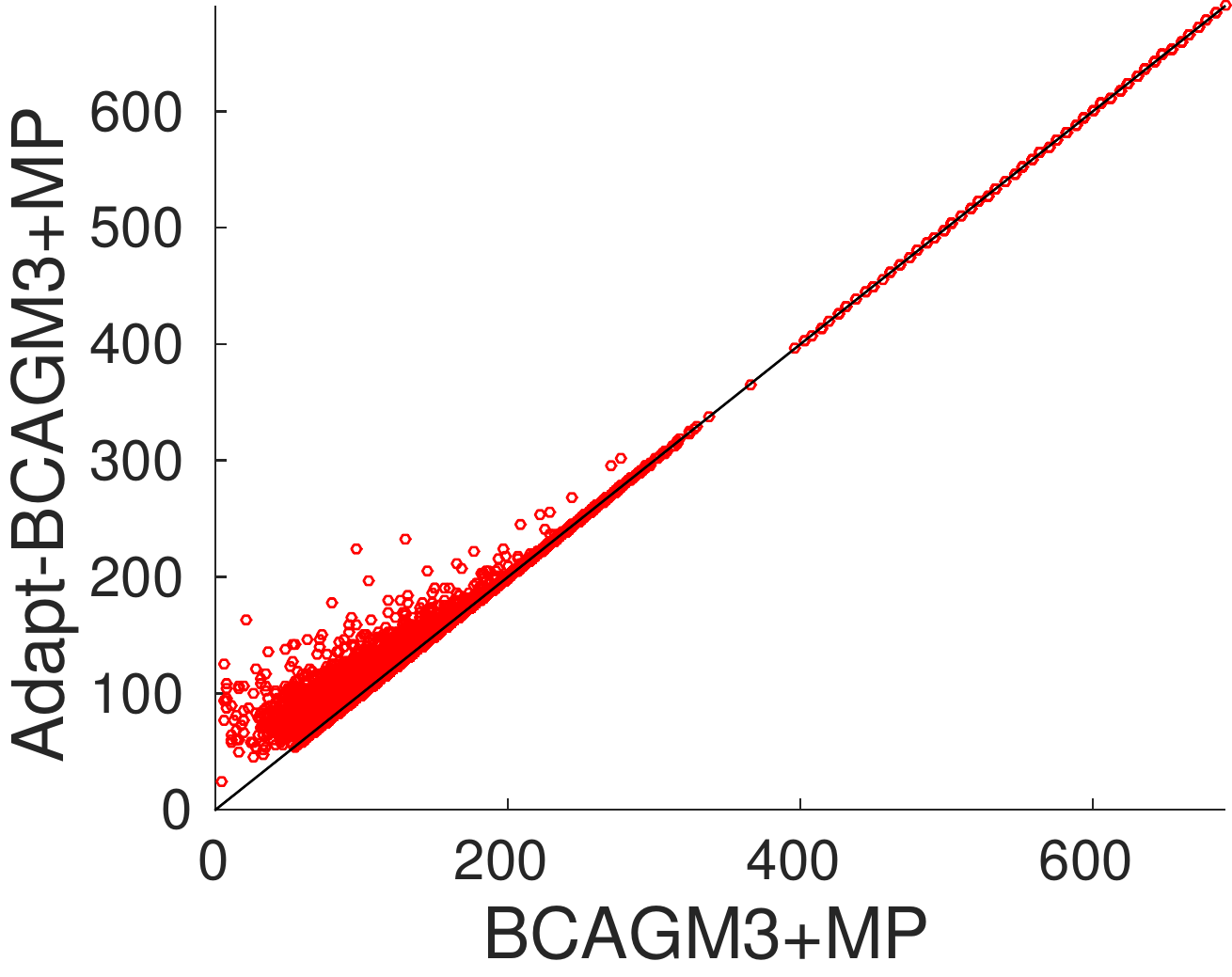}} 
    \\
    \subfloat{\includegraphics[width=1.01in]{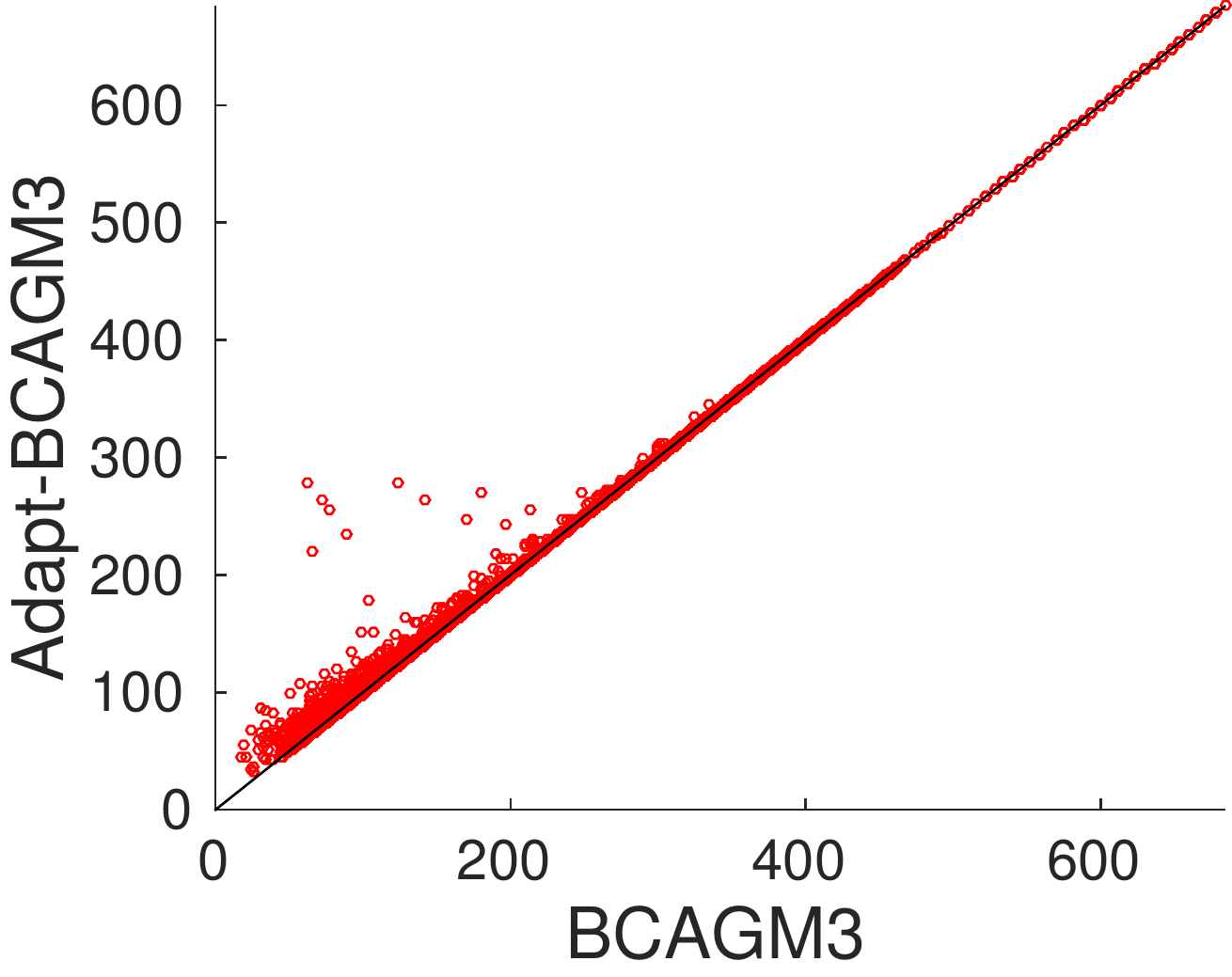}}
    \subfloat[Winebottle]{\includegraphics[width=1.01in]{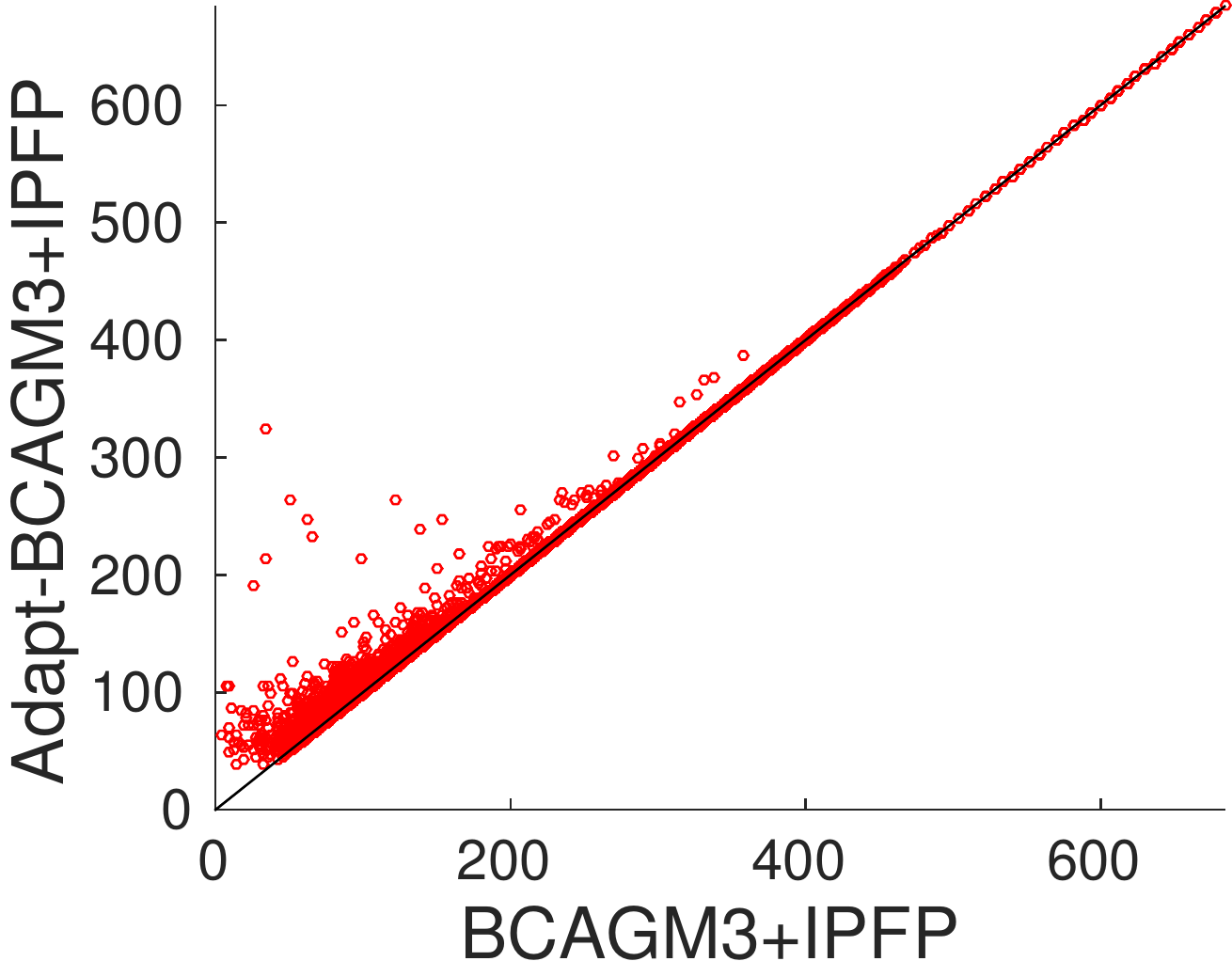}}
    \subfloat{\includegraphics[width=1.01in]{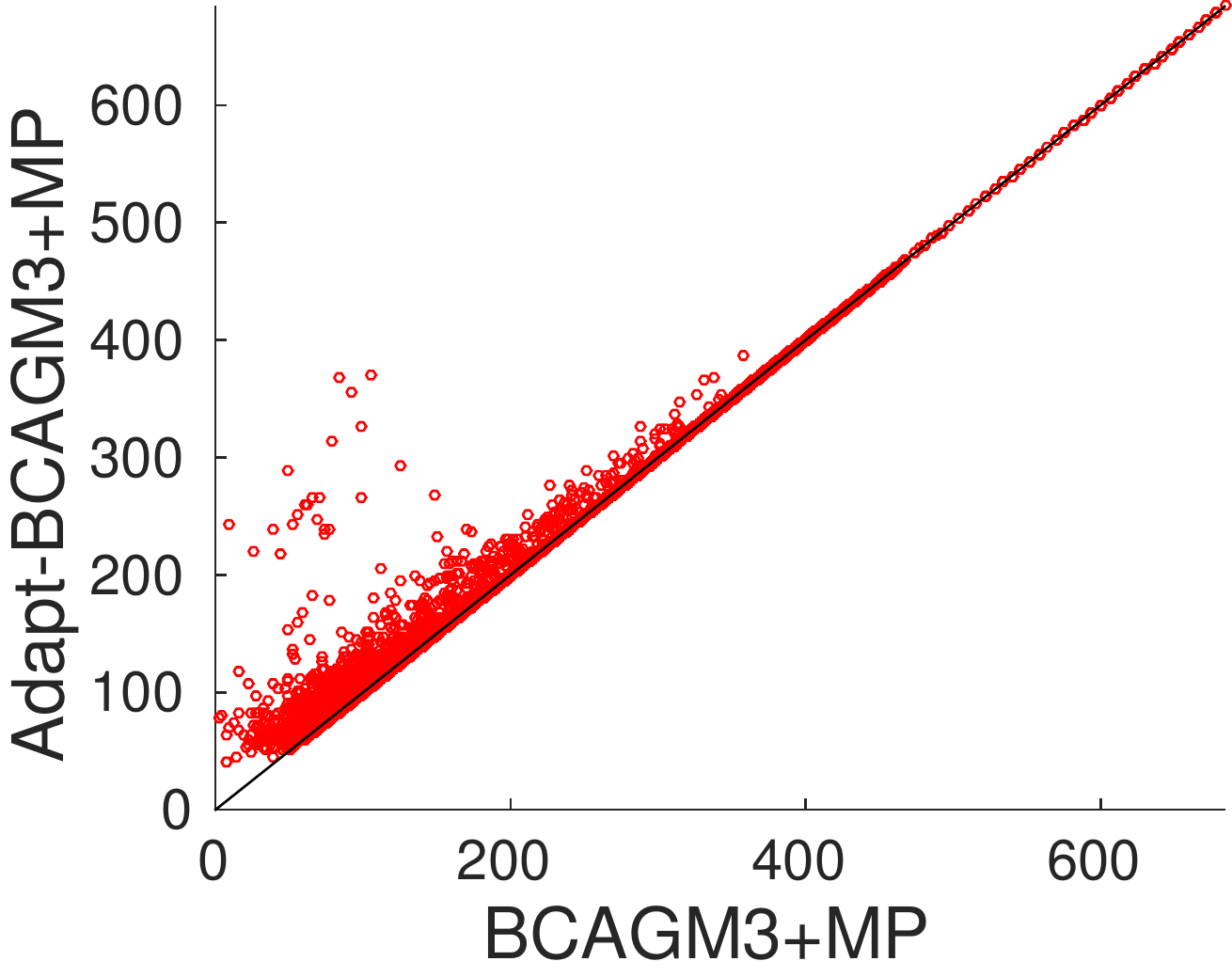}}
    \caption{Scatter plots showing matching score of adaptive vs. non-adaptive methods over different datasets.
    }
    \label{fig:scatter_adapt_nonadapt}
\end{figure}

\subsection{Advantages of adaptive methods}
\label{sec:adap3_vs_nonadapt3}
In this section, we compare our adaptive algorithms with the non-adaptive versions.
In particular, we conduct our experiments on a realistic dataset \cite{ChoEtal2013} 
which consists of three object classes: face, duck and wine bottle. 
Each class contains at least $50$ images with different instances.\footnote{The dataset consists of $109$ Face, $50$ Duck, $66$ Wine bottle and is available from \cite{willowObjectDataset}.}
Every image in the dataset is annotated with $10$ ground truth points for matching.
We perform our experiments separately on each object class, for which we test all our algorithms on 
$2000$ image pairs randomly selected from the same class.
In order to make the tasks more difficult, we randomly add to one image several points that are extracted using a SIFT detector.
This simulates realistic outlier points resulting from cluttered background in images.
In particular, the number of outliers is varied from $0$ to $20$ which yields $4200$ test cases per object class in total.
The experimental results are shown in the scatter plot of Fig. \ref{fig:scatter_adapt_nonadapt} where
the matching score of the non-adaptive method (x-axis) is plotted against the matching score of the adaptive version (y-axis).
One can clearly see that, \cgram{in all the experiments,} 
the adaptive methods always achieve similar or better matching score than the
non-adaptive methods, which confirms the advantage of gradually updating $\alpha.$

To better understand our algorithms especially in terms of matching accuracy,
in the next sections we will compare them with state-of-the-art methods over different datasets.
Since the proposed adaptive versions perform better than the non-adaptive approaches, 
we show below only the performance of the adaptive ones.
The appendix provides more detailed comparison between both approaches
as well as the fourth-order methods from \cite{Quynh2015}.

\begin{figure}
    \centering
    \subfloat{\includegraphics[width=0.47\linewidth]{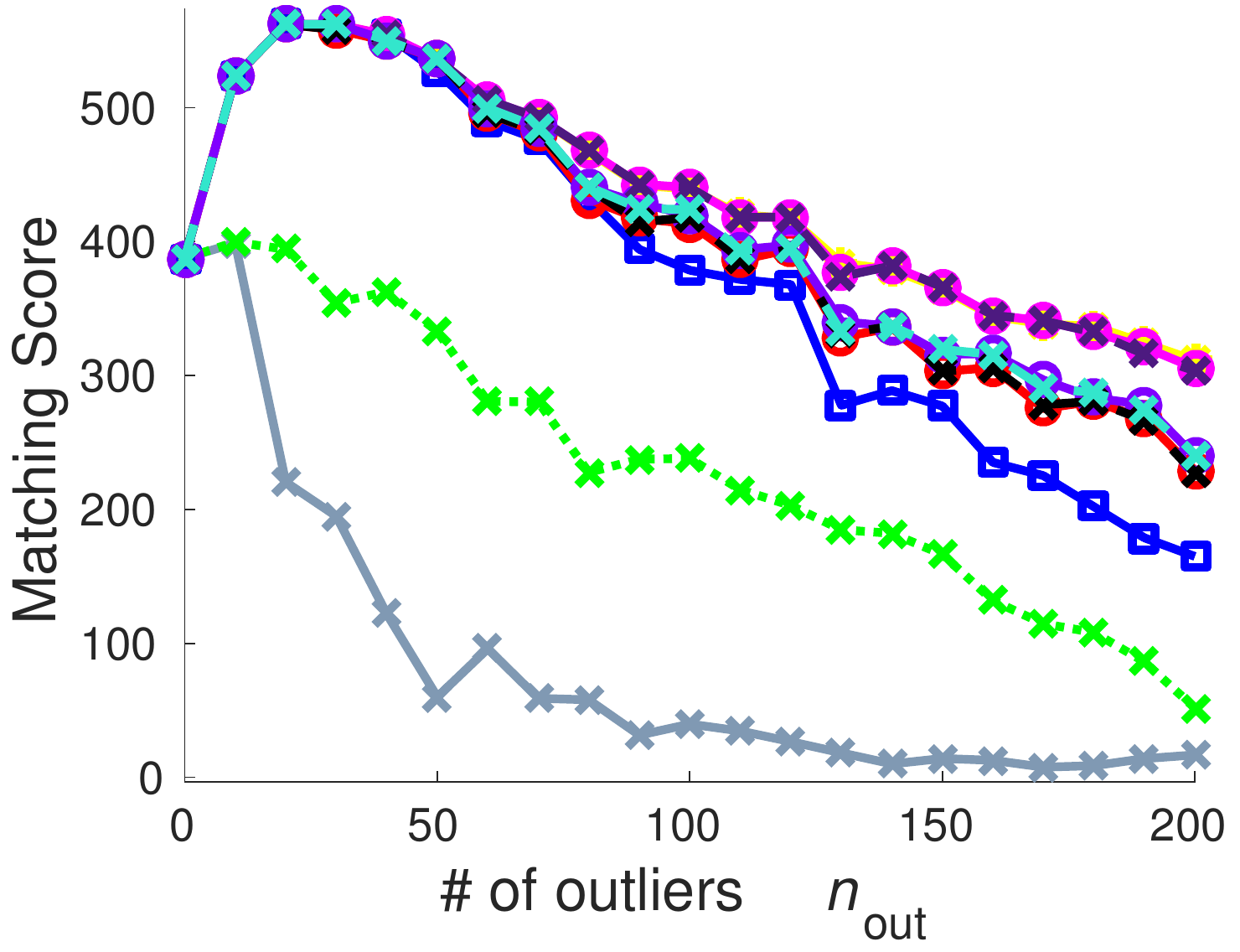} }
    \subfloat{\includegraphics[width=0.47\linewidth]{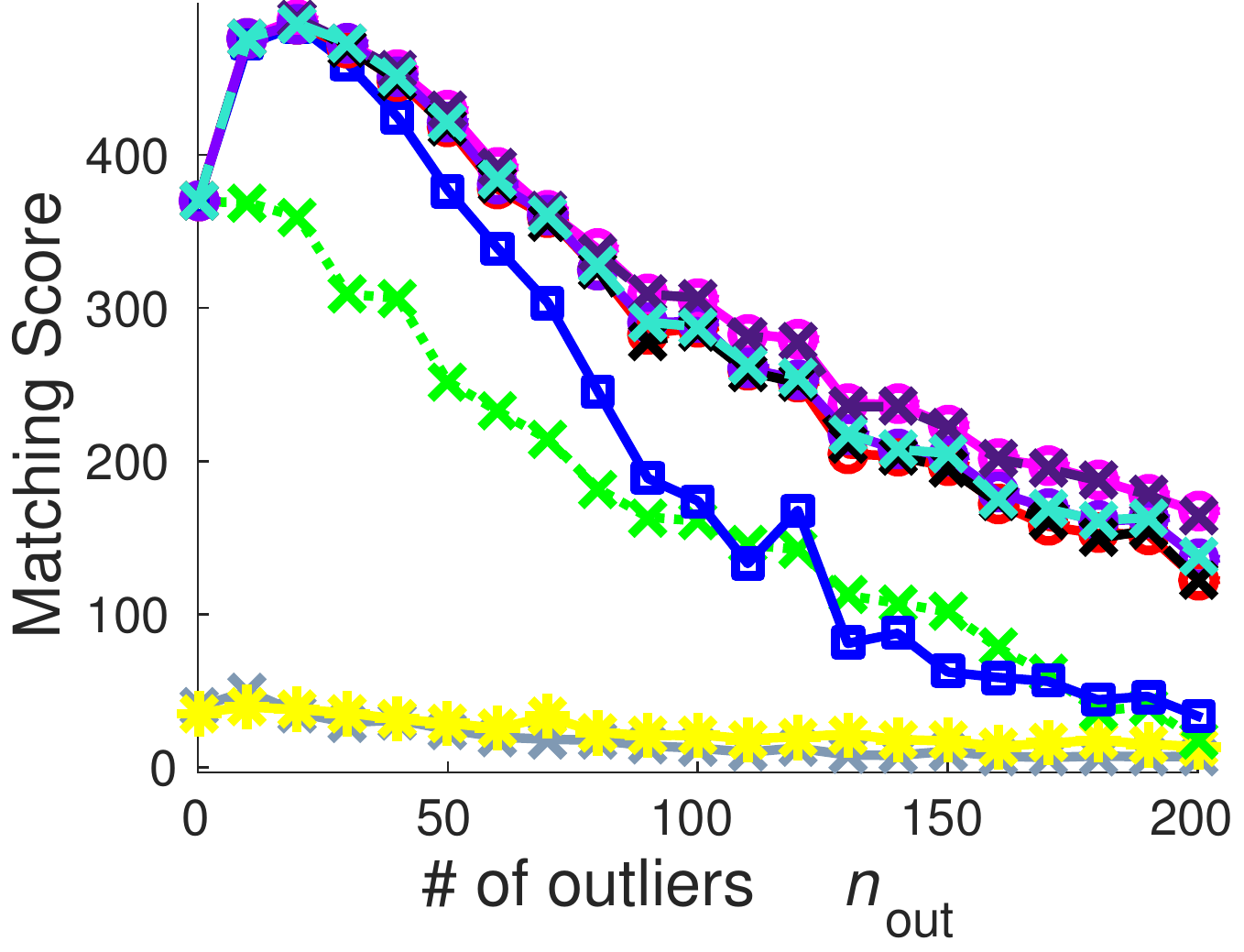} }
    \\
    \setcounter{subfigure}{0}
    \subfloat[$n_{in}=10, \sigma=0.01$]{
	\includegraphics[width=0.47\linewidth]{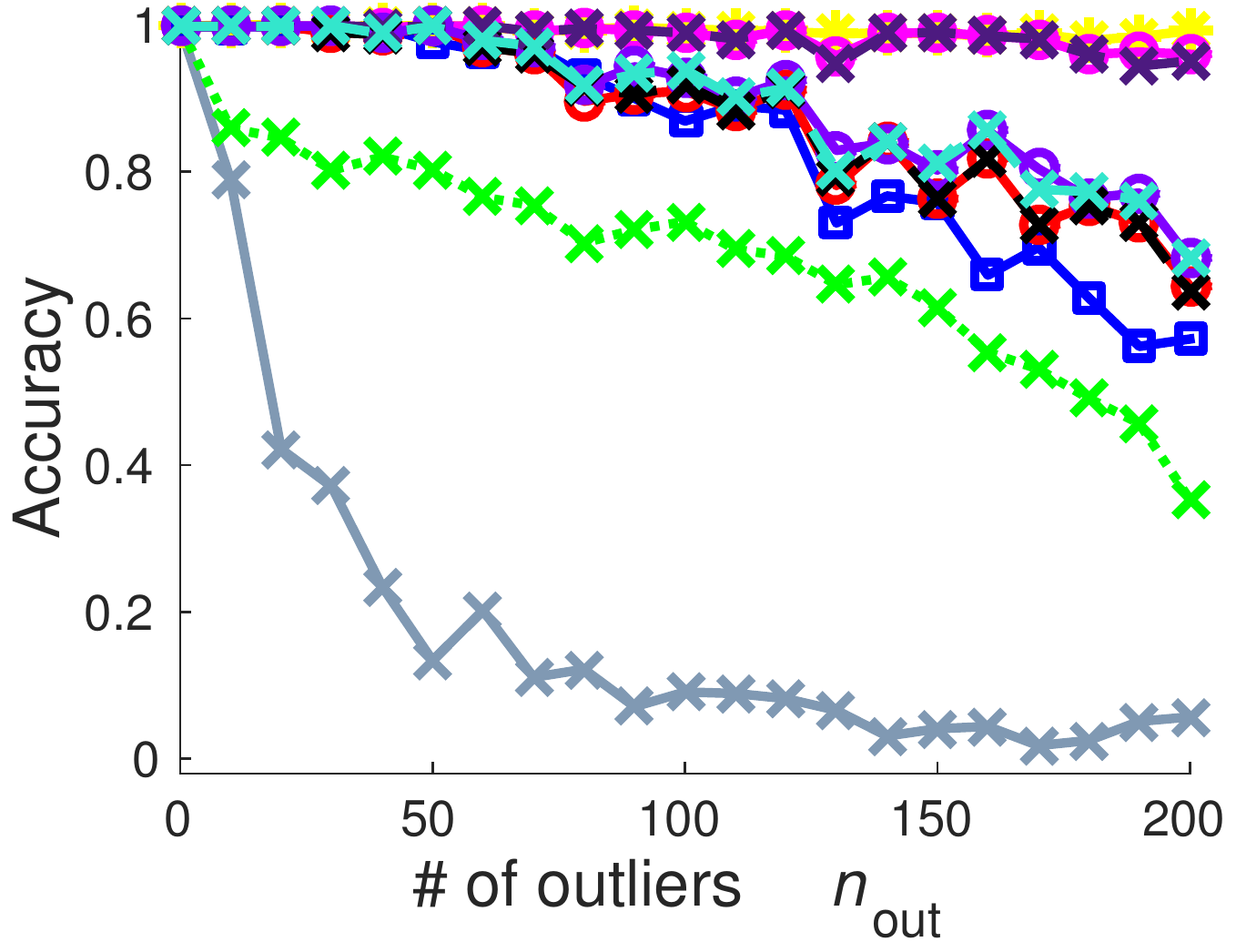}
    }
    \subfloat[$n_{in}=10, \sigma=0.03, \textrm{scale}=1.5$]{
	\includegraphics[width=0.47\linewidth]{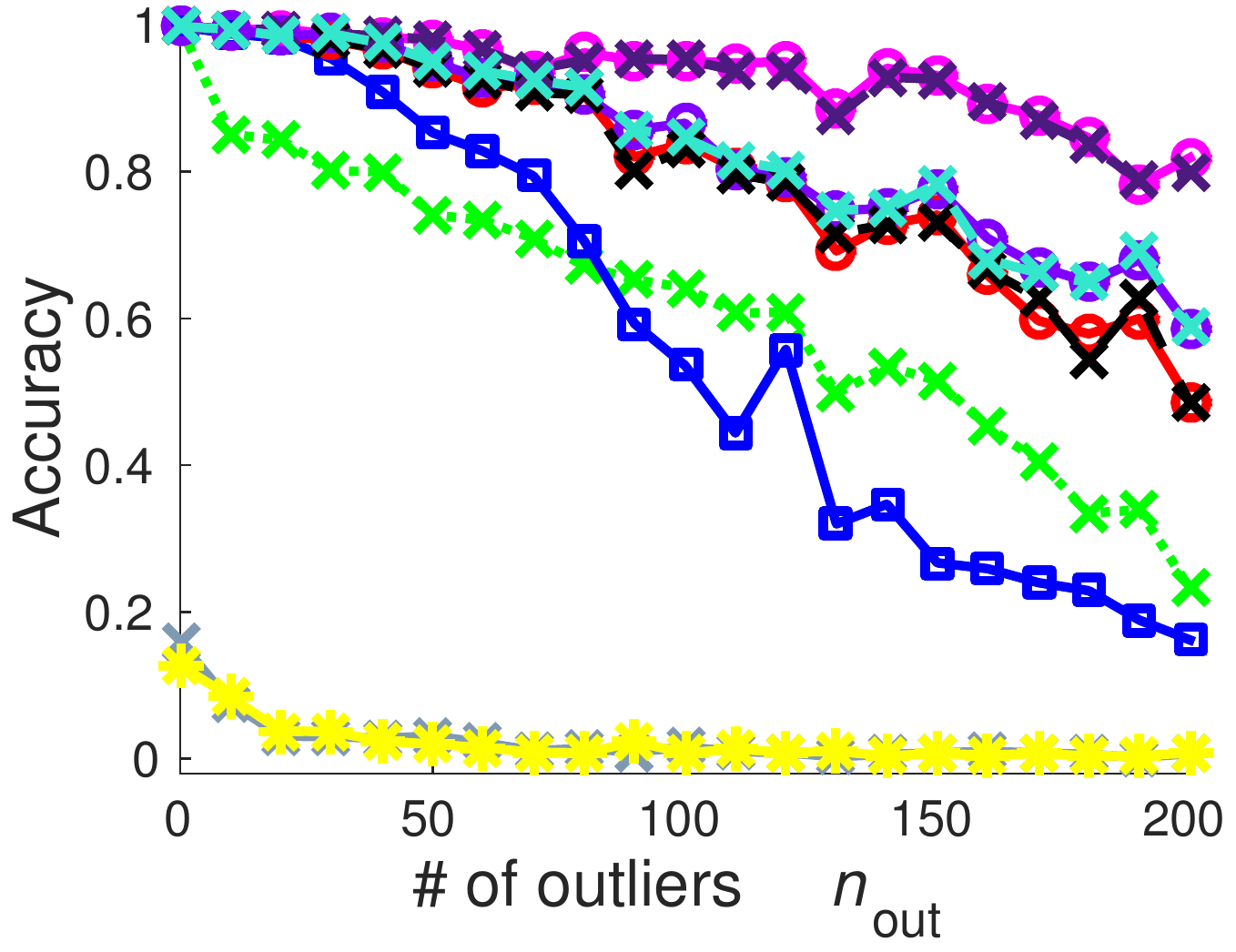}
    }
\caption{
Matching point sets in $\R^2$. The number of outliers is varied from $0$ to $200$. 
\crevision{See Fig. \ref{fig:def} for the legend.}
(Best viewed in color.)
}
\label{fig:exp_out}
\end{figure}

\begin{figure}
    \centering
    \subfloat{ \includegraphics[width=0.48\linewidth]{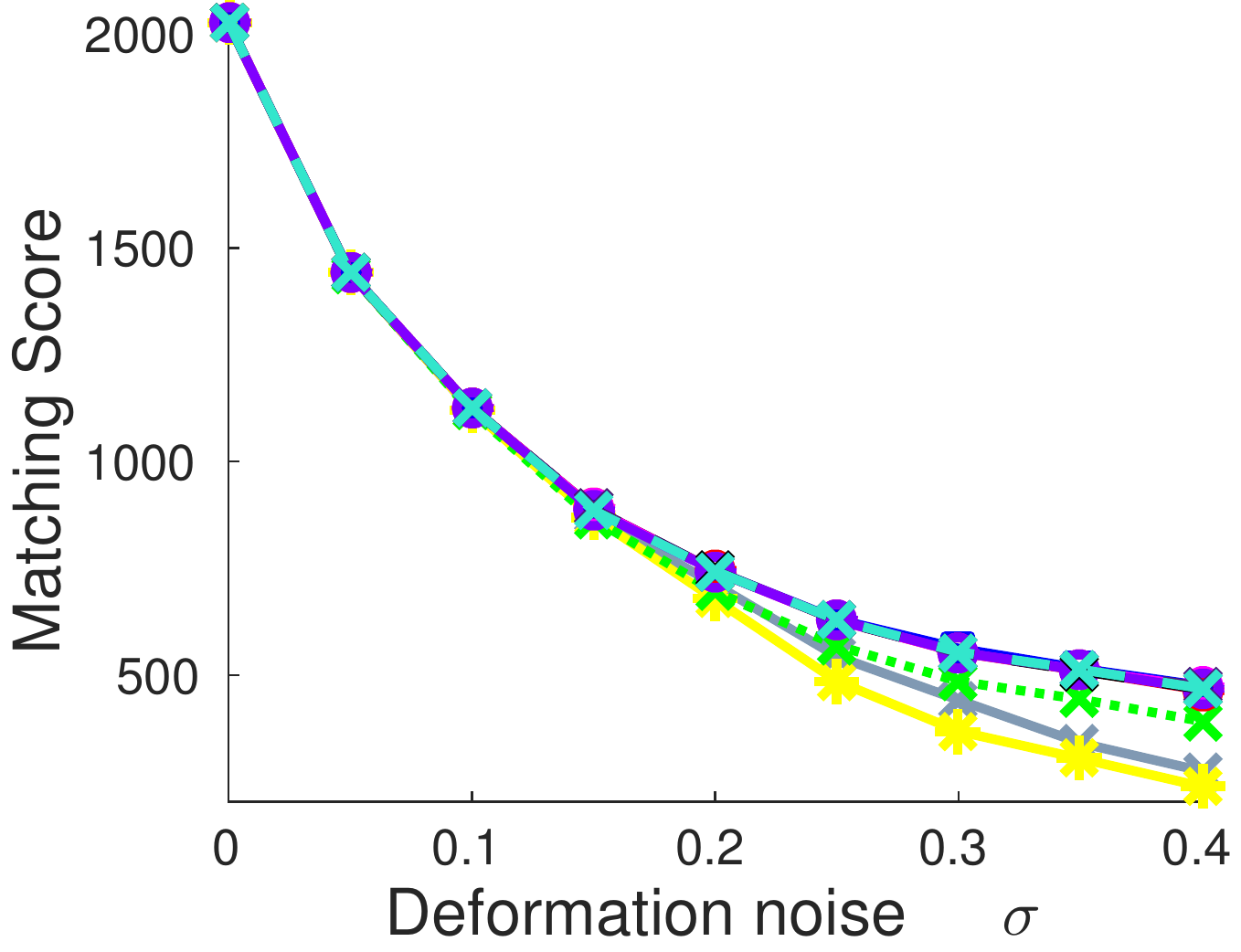} } 
    \subfloat{ \includegraphics[width=0.48\linewidth]{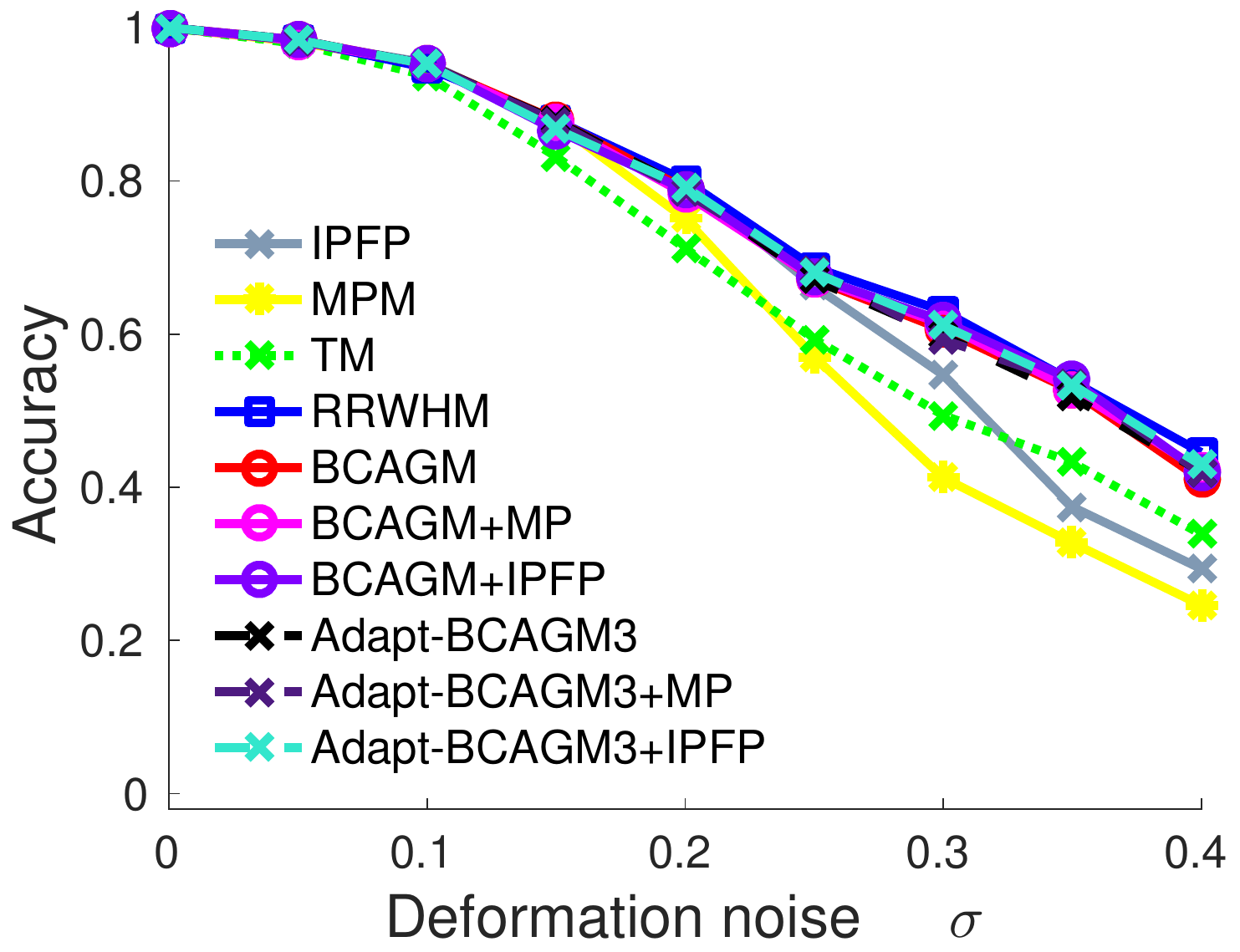} } 
\caption{
    Performance of all algorithms with varying deformation noise.
    The number of inlier points is fixed as $n_{in}=20.$ (Best viewed in color.)
}
\label{fig:def}
\end{figure}
\subsection{Synthetic Dataset}\label{subsec:exp_synthetic}
In this section we are trying to match point sets in $\R^2$ which is a standard benchmark for testing
graph and hypergraph matching methods \cite{ChoEtAl2014, LeeChoLee2011, ShiEtal2007, LeoHeb2009, ZhoTor2013}.
We follow a similar procedure \cblue{as} in \cite{ChoEtAl2014} for the creation of two point sets $P$ and $Q$.
In order to create $P$ we sample $n_{in}$ inlier points from a Gaussian distribution $\Nc(0,1)$.
These points are then copied to $Q$ and subsequently modified by several operations in the following order:
1) the coordinates of every point are multiplied with a $\textrm{scale}$ factor;
2) additive Gaussian noise $\Nc(0,\sigma^2)$ is added to every point where $\sigma$ is hereinafter referred to as deformation;
3) on top of that $n_{out}$ random points, that we refer to as outliers, are added to $Q$ by sampling from $\Nc(0,1)$.
Depending on different scenarios described below, 
some operations in the above chain might be absent but their order is the same.
Based on that, we test all matching algorithms under different changes in the data: outliers, deformation and scaling.
In this section, we consider two main settings: outlier and deformation setting.
In all plots in this section, each quantitative result was obtained by averaging over $100$ random trials. 
The accuracy is computed as the ratio between the number of correct matches and the number of inlier points.
We use $\sigma_s = 0.5$ in the affinity construction for the \cblue{second-order} methods as suggested by \cite{ChoEtAl2014}.

In the outlier setting, we perform two experiments.
In the first experiment, we vary the number of outliers from $0$ to very challenging $200$, 
which is $20$-times more than the number of inliers $n_{in}$, while $\sigma$ is set to $0.01$ 
to simulate noise in realistic settings and no scaling is used here (\ie $\textrm{scale}=1.0$).
In the second case, we slightly increase $\sigma$ to $0.03$ and scale all the points in $Q$ by a factor of $1.5$, 
making the problem more realistic and also much more difficult.
This experiment shows the robustness of higher-order methods over \cblue{second-order} ones w.r.t. scale transformations.
The results are shown in Fig. \ref{fig:exp_out}.
Overall, our higher-order algorithms show remarkable tolerance to the number of outliers
and significantly outperform previous higher-order as well as second-order methods 
in terms of both matching score and matching accuracy.
The better performance of higher-order approaches shows their robustness against geometric transformations
compared to second-order ones.
In the first experiment without scaling as shown in Fig. \ref{fig:exp_out} (a), 
most higher-order methods and MPM 
perform very well without being deteriorated much by outliers.
When the scale change is present in the data as shown in Fig. \ref{fig:exp_out}(b),
our algorithms significantly outperform the current state-of-the-art.
In particular, both MPM and IPFP fail in this case since they use pairwise distance similarities
which are clearly not invariant under scaling.
Previous higher-order \cgram{algorithms} like RRWHM or TM also show quite rapidly decreasing performance
as the number of outliers increases compared to our \cgram{methods}.
This experiment also features the difference between two groups of our algorithms, 
\ie the ones using MPM as subroutine including BCAGM+MP and Adapt-BCAGM3+MP perform the best in both settings.
Thus, one can state that our approach transfers the robustness of MPM as a second-order method 
to the higher-order case and thus is able to deal with scaling which is difficult for \cblue{second-order} methods.

For the deformation setting, $\sigma$ is varied from $0$ to challenging $0.4$ while 
the number of inliers $n_{in}$ is fixed to $20$. 
Since we want to test the tolerance of all methods only with respect to deformation, 
we do not add any \cgram{outliers} nor scale the points in this setting.
This type of experiment has been used in previous works \cite{ChoEtAl2014, LeeChoLee2011},
where $\sigma$ is varied from $0$ to $0.2.$
Fig. \ref{fig:def} shows the result where our algorithms are quite competitive to 
RRWHM which has been shown to be robust to deformation before \cite{LeeChoLee2011}. 
It is interesting to note that while MPM is very robust against outliers (without scaling), 
it is outperformed by other methods when the deformation is large.
Notably, our algorithms BCAGM+MP and Adapt-BCAGM3+MP using MPM as subroutine are not affected by this slight weakness of MPM.

Considering all experiments in this section, our algorithms outperform or are on par with all previous approaches
in terms of robustness to outliers, deformation and scaling. 
Moreover, our proposed third-order algorithms show on average quite similar performance compared to their fourth-order counterparts \cite{Quynh2015}.

\begin{figure}
    \centering
    \subfloat[30 pts vs. 30 pts]{ \includegraphics[width=0.48\linewidth]{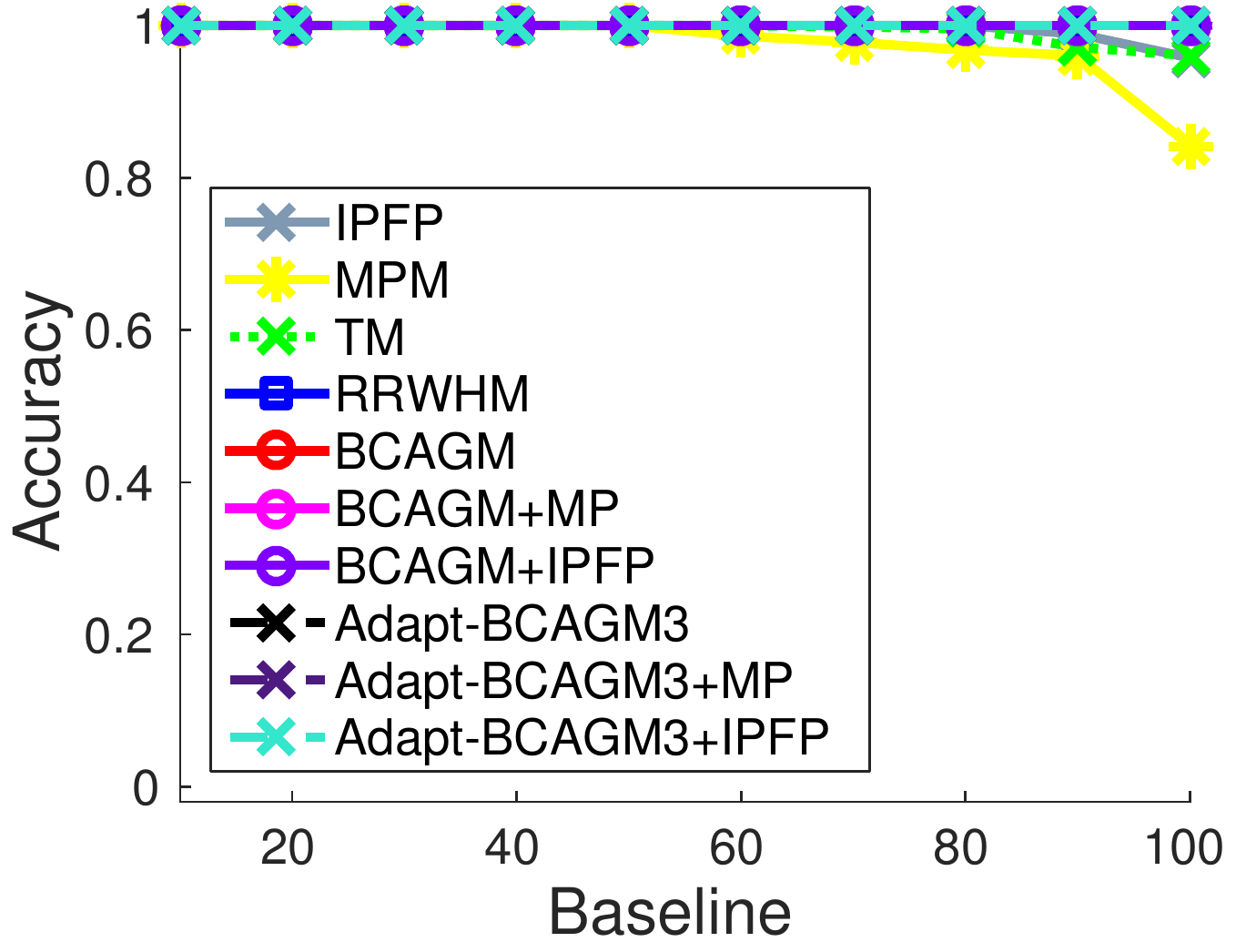} } 
    \subfloat[20 pts vs. 30 pts]{ \includegraphics[width=0.48\linewidth]{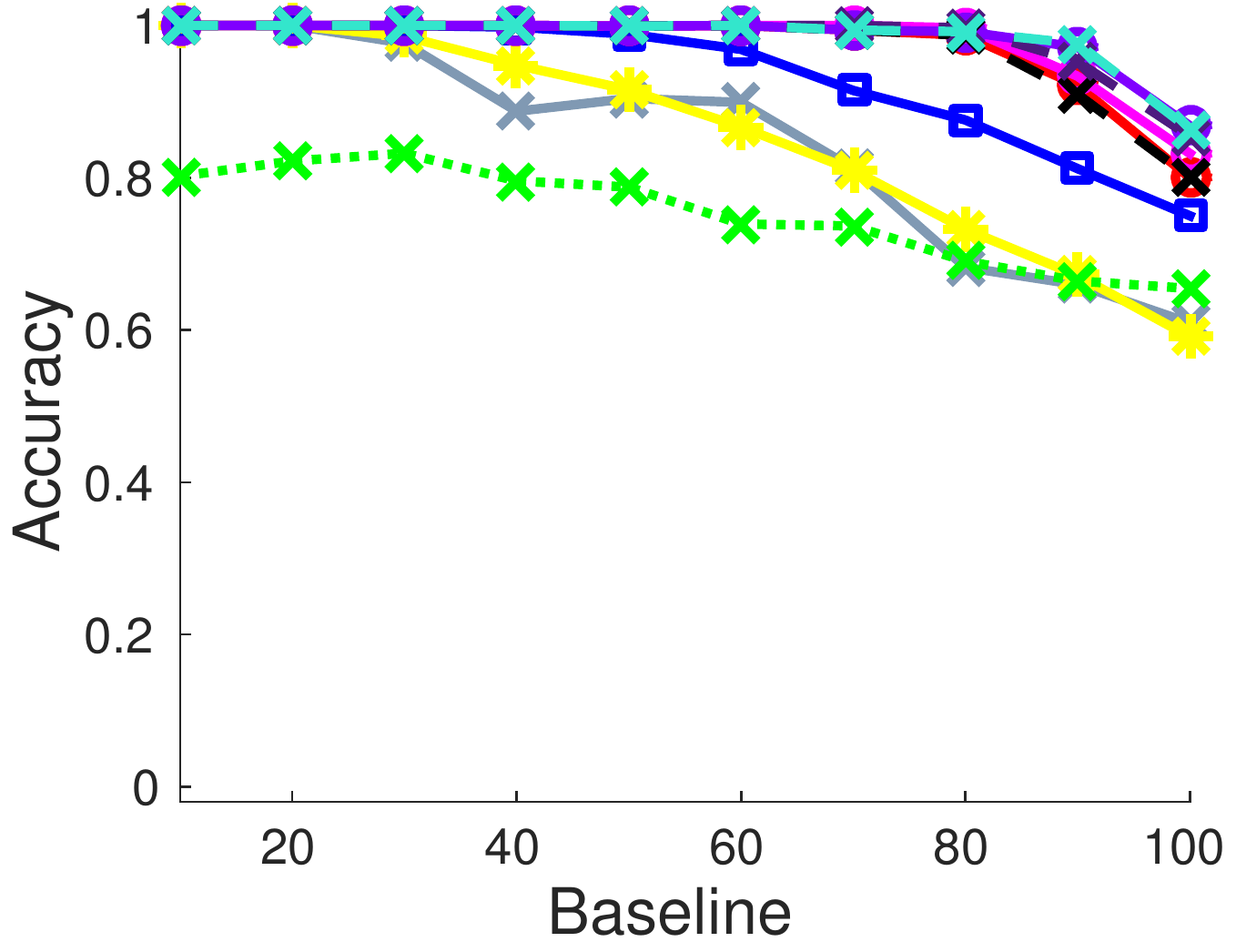} }
    \\
    \subfloat[10 pts vs. 30 pts]{ \includegraphics[width=0.48\linewidth]{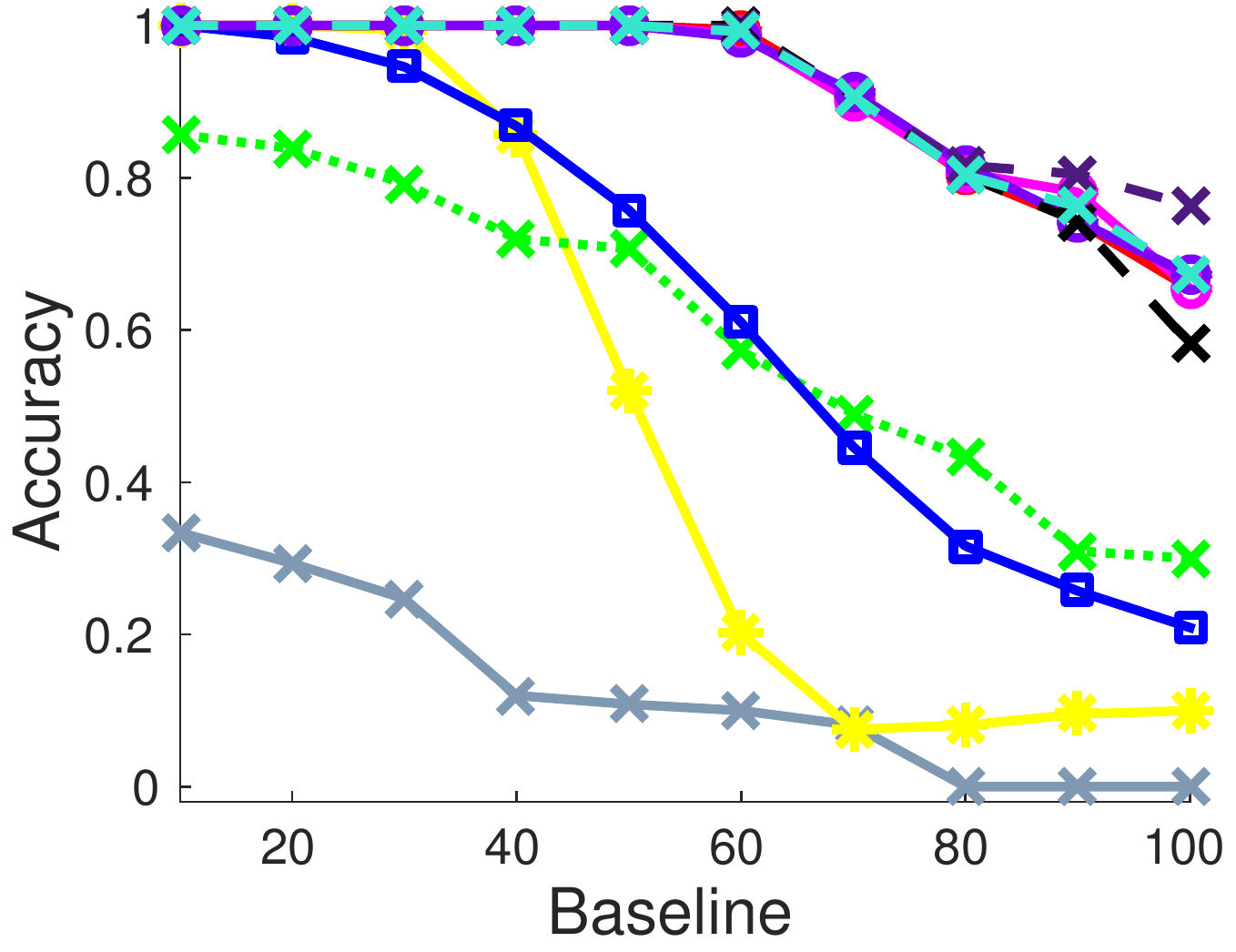} }  
    \subfloat[10 pts vs. 30 pts]{ \includegraphics[width=0.48\linewidth]{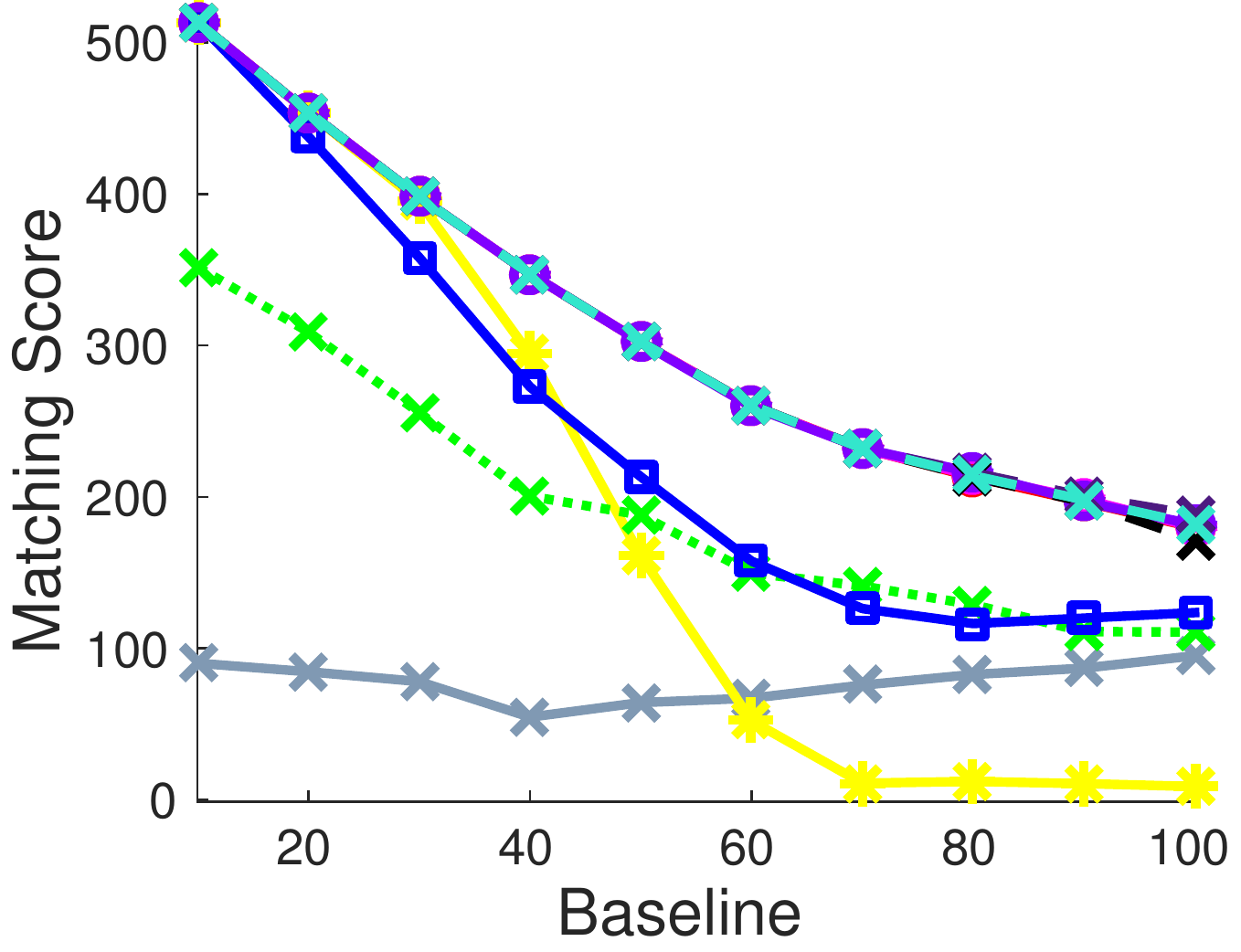} } 
\caption{
CMU house dataset with \crevision{different number of points in two images}. (Best viewed in color.)
}
\label{fig:exp_house}
\end{figure}

\subsection{CMU House Dataset}\label{subsec:exp_house}
The CMU house dataset has been widely used in previous work ~\cite{ChoLeeLee2010,LeeChoLee2011,DucEtAl2011,ZhoTor2013} 
to evaluate matching algorithms. 
In this dataset, 30 landmark points are manually tracked over a sequence of 111 images, 
which are taken from the same object under different view points.
In this experiment, ``baseline'' denotes the distance of the frames in the sequence and thus correlates
well with the difficulty to match the corresponding frames.

We match all possible image pairs with ``baseline'' of $10, 20, 30, \ldots, 100$ frames 
and compute the average matching accuracy for each algorithm. 
The algorithms are evaluated in three settings. In the first experiment, we match $30$ points to $30$ points.
Then we make the problem significantly harder by randomly removing points from one image, motivated by a scenario 
where one has background clutter and thus not all points can be matched. 
This results in two matching experiments, namely $10$ points to $30$ points, and $20$ points to $30$ points. 
For the choice of $\sigma_s$ in the affinity tensor for \cblue{second-order} methods, 
we follow ~\cite{ChoLeeLee2010,ZhoTor2013} by setting $\sigma_s = 2500$.

The experimental results are shown in Fig. \ref{fig:exp_house}. 
While most algorithms perform rather well on the $30$ to $30$ task, 
our methods perform significantly better than all other methods in the more difficult tasks, 
thus showing as for the synthetic datasets that our methods are quite robust
to different kinds of noise in the matching problem.
Notably, in the $10$ to $30$ task, Adapt-BCAGM3+MP outperforms BCAGM+MP for large baselines.

\begin{figure}
    \centering
    \subfloat[House]{ \includegraphics[width=0.48\linewidth]{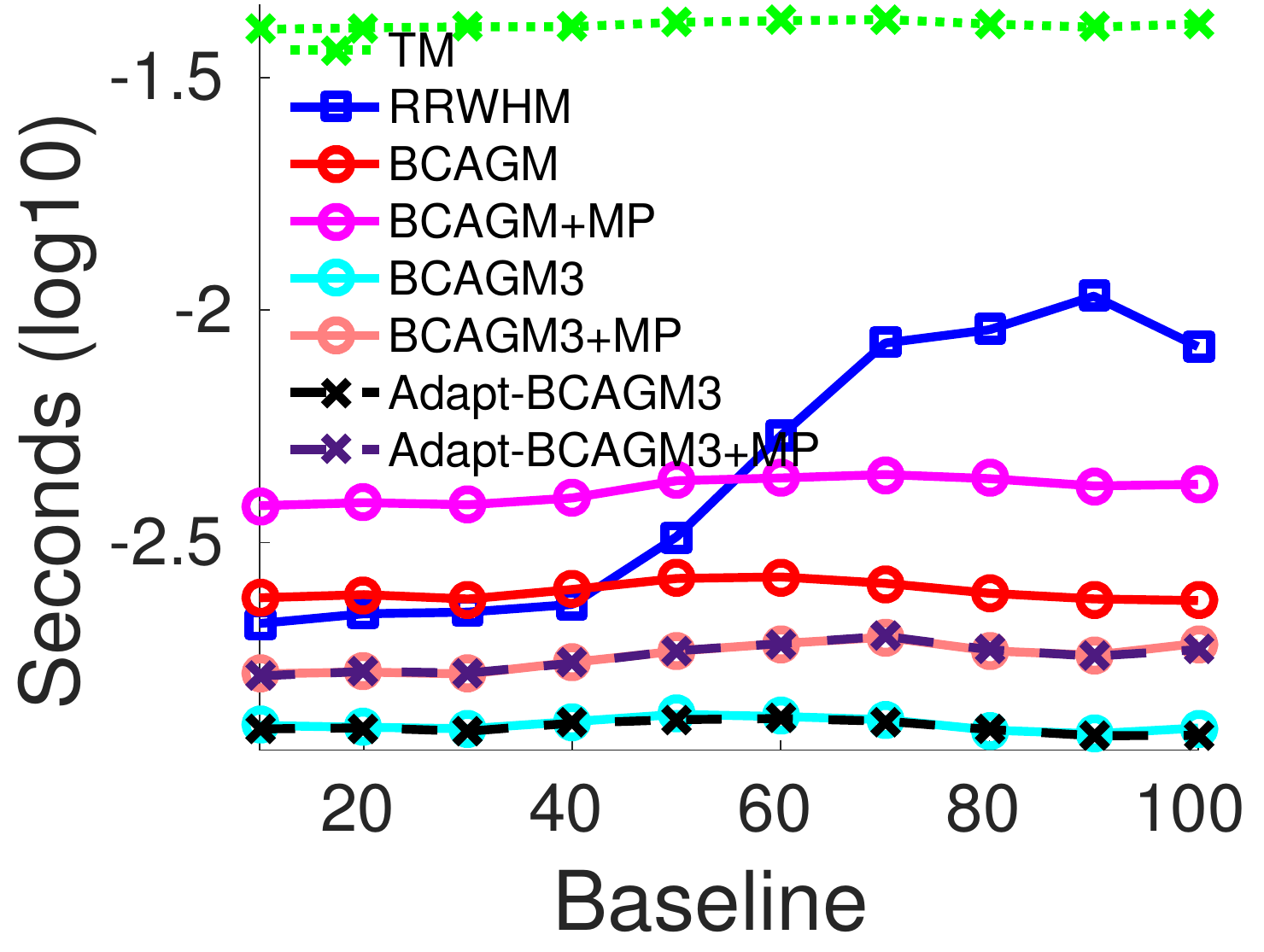} }  
    \subfloat[Willow dataset]{ \includegraphics[width=0.48\linewidth]{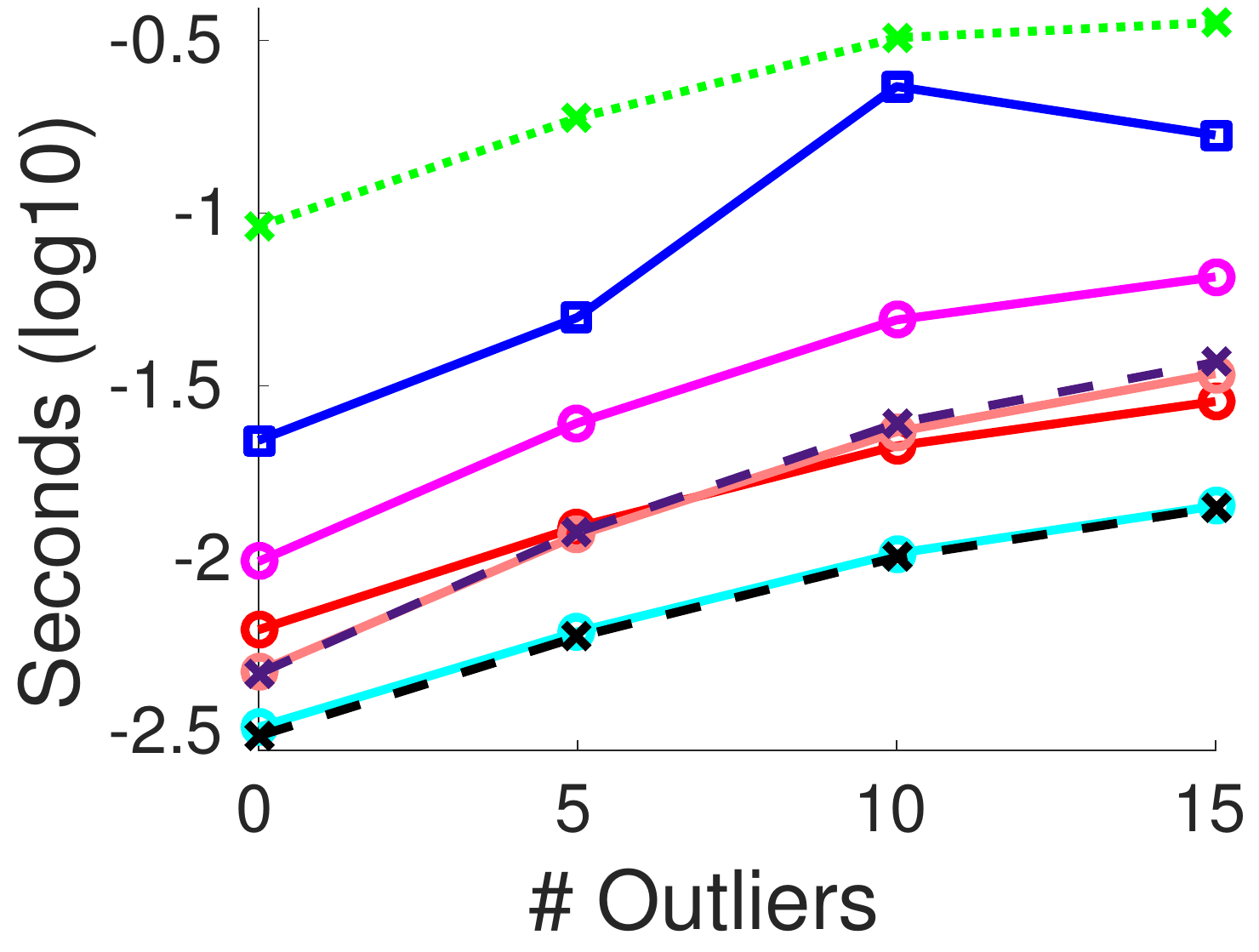} }  
\caption{
Running time of higher-order methods (Best viewed in color.)
}
\label{fig:exp_runtime_ours_others}
\end{figure}

\subsection{Willow Object Class Dataset}
\begin{figure*}
    \centering
    \subfloat{ \includegraphics[width=0.31\linewidth]{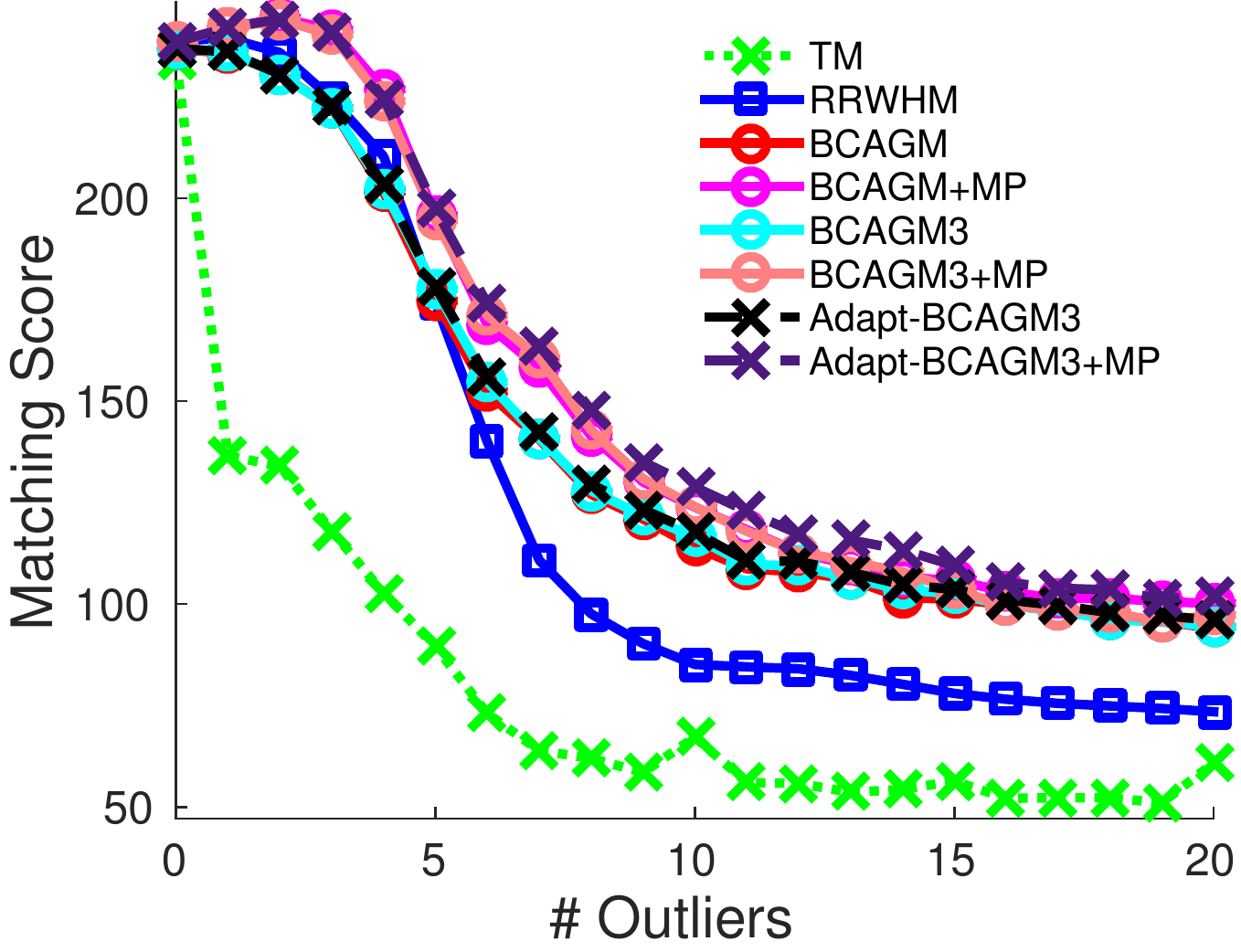} }  
    \subfloat{ \includegraphics[width=0.31\linewidth]{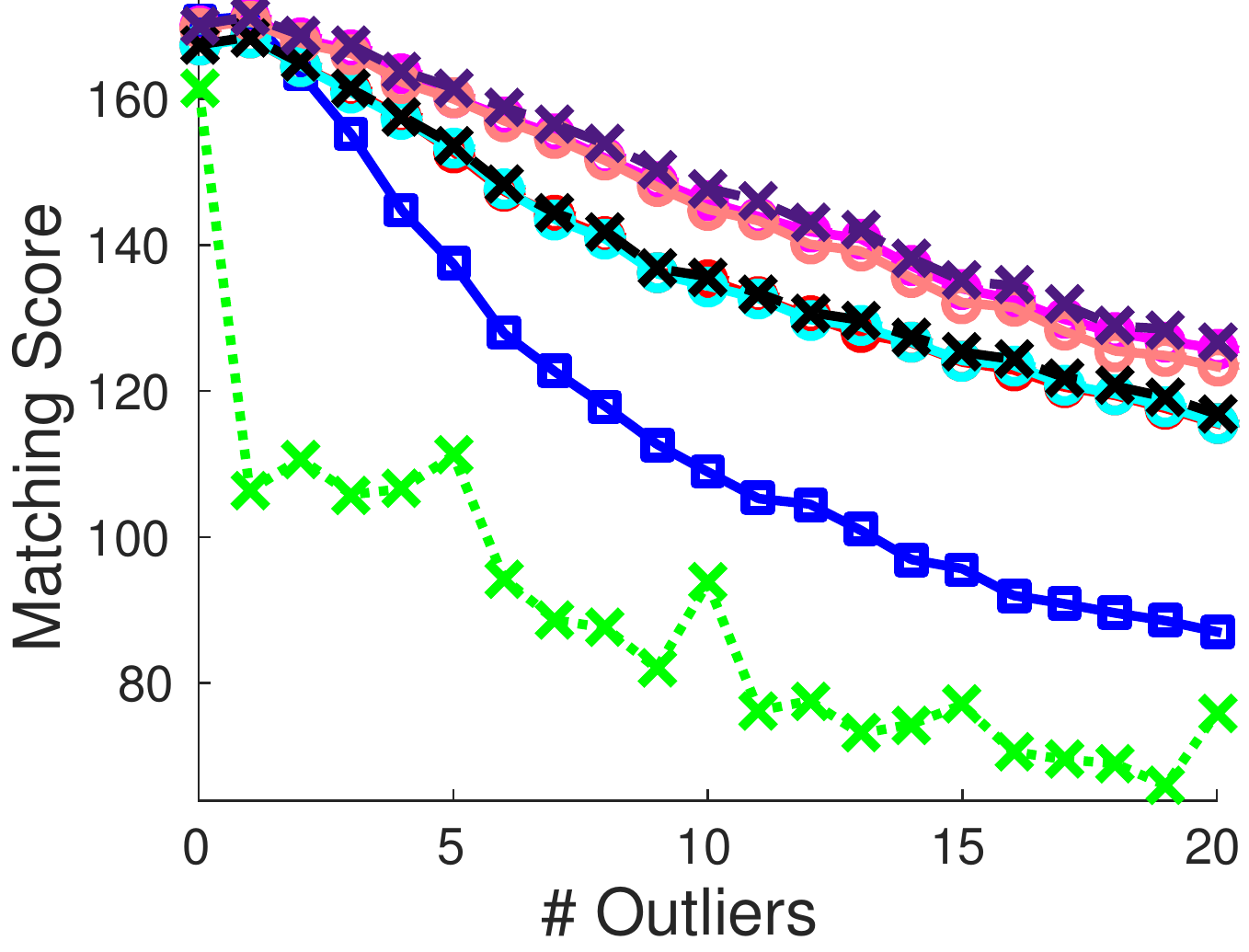} }  
    \subfloat{ \includegraphics[width=0.31\linewidth]{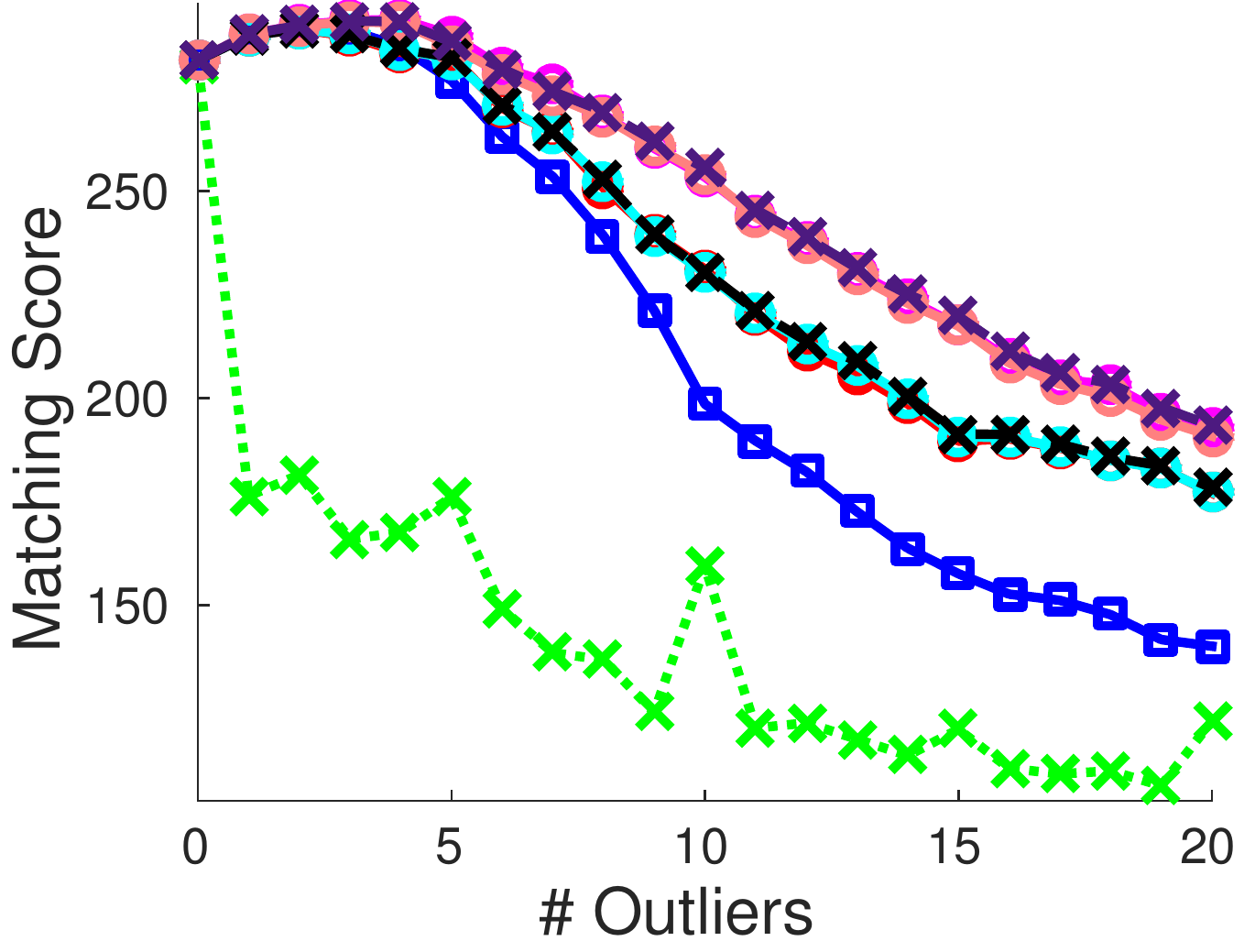} }  
    \\
    \setcounter{subfigure}{0}
    \subfloat[Face]{ \includegraphics[width=0.31\linewidth]{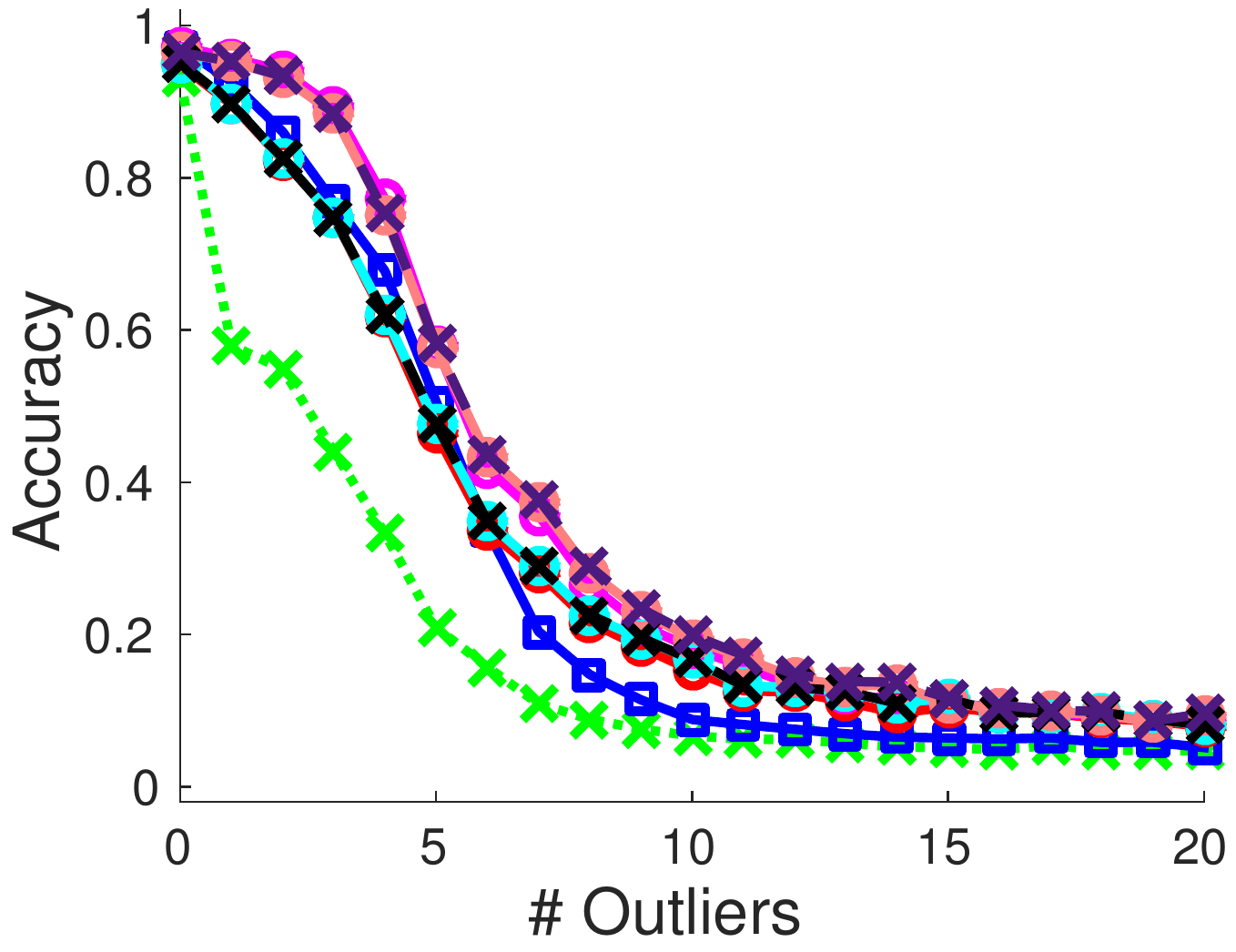} }  
    \subfloat[Duck]{ \includegraphics[width=0.31\linewidth]{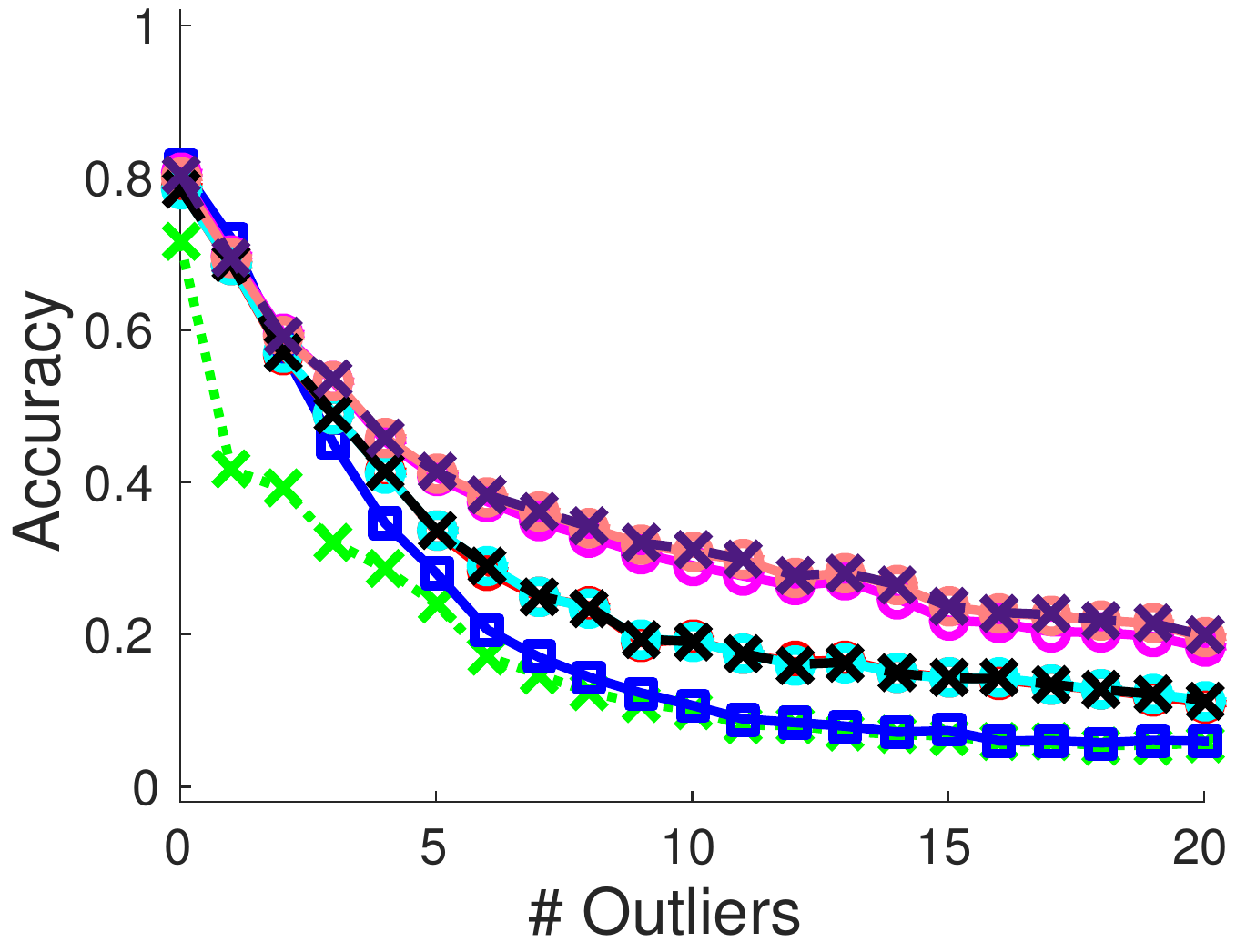} }  
    \subfloat[Winebottle]{ \includegraphics[width=0.31\linewidth]{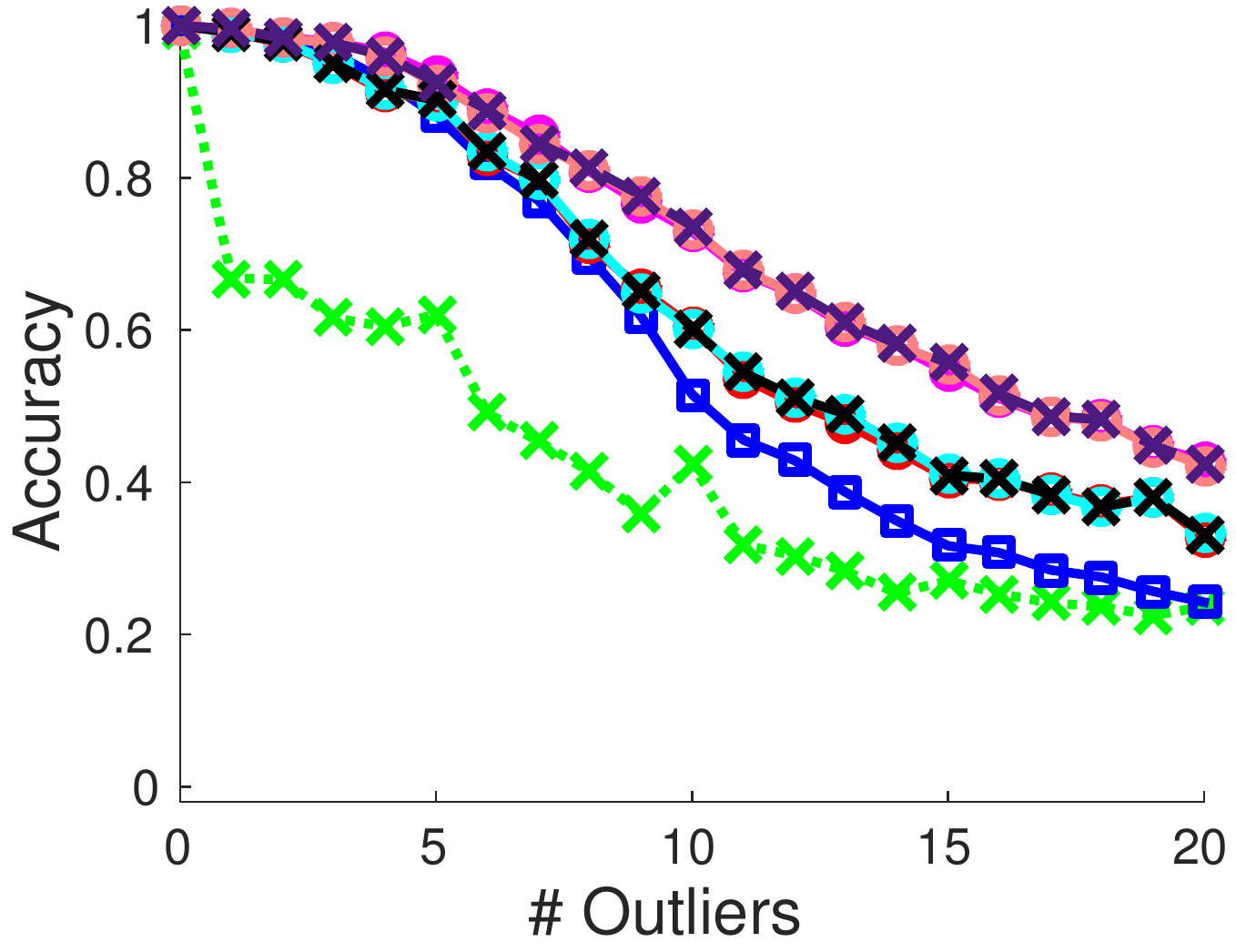} }  
\caption{
Matching score and accuracy of higher-order algorithms on Willow Object Dataset \cite{ChoEtal2013}. (Best viewed in color.)
}
\label{fig:exp_willow}
\end{figure*}
In this last experiment we use the Willow Object Class dataset \cite{willowObjectDataset} 
previously introduced in Section \ref{sec:adap3_vs_nonadapt3}.
We use exactly the same setting as in Section \ref{sec:adap3_vs_nonadapt3}.
We test all algorithms on $2000$ randomly selected image pairs from each object class,
where the number of outliers in one image is varied from $0$ to $20.$
For a given number of outliers, we report in Fig. \ref{fig:exp_willow} the average performance of all algorithms 
over all the $2000$ image pairs.
It can be seen that both our third-order and fourth-order methods consistently outperform the state-of-the-art 
for all the object classes even so the differences are a bit smaller.
In particular, for the Duck class, Adapt-BCAGM3+MP performs slightly better than BCAGM+MP even though this is hard to see
(see the Appendices for detailed comparisons) because the result has been averaged already over all pairs of images.

\subsection{Comparison between third-order methods and fourth-order methods}
\label{sec:third_vs_fourth}
\begin{table*}
    \centering
    \renewcommand{\arraystretch}{1.3}
    \begin{tabular}{|c|c|c|c|c|c|c|c|c|}
	\hline
	& \multicolumn{2}{c|}{Synthetic} & \multicolumn{2}{c|}{House} & \multicolumn{2}{c|}{Face+Duck+Winebottle} \\
	\hline \hline
	\textbf{Comparison} & No. & Avg(\%) 		& No. & Avg(\%) 	& No. & Avg(\%) \\ \hline 
	\hline
	\rowcolor{Gray}
	BCAGM3 $>$ BCAGM & 			241&51.9	&  		2&1.5 		& 		6277&30.0  \\ \hline
	BCAGM3 $<$ BCAGM & 			292&347.0 	&  		7&18.5 		& 		7130&27.0  \\ \hline
	BCAGM3 $=$ BCAGM & 			4567&0.0	&  		1671&0.0 	& 		49593&0.0  \\ \hline
	\hline
	\rowcolor{Gray}
	BCAGM3+IPFP $>$ BCAGM+IPFP & 		186&2.4		&  		19&7.5 		& 		8960&24.0  \\ \hline
	BCAGM3+IPFP $<$ BCAGM+IPFP & 		188&59.5 	&  		5&6.9 		& 		9834&41.0  \\ \hline
	BCAGM3+IPFP $=$ BCAGM+IPFP & 		4726&0.0	&  		1656&0.0 	& 		44206&0.0  \\ \hline
	\hline
	\rowcolor{Gray}
	BCAGM3+MP $>$ BCAGM+MP & 		194&43.4	&  		4&12.2 		& 		5695&32.0  \\ \hline
	BCAGM3+MP $<$ BCAGM+MP & 		243&237.1 	&  		6&27.6 		& 		7461&56.0  \\ \hline
	BCAGM3+MP $=$ BCAGM+MP & 		4663&0.0	&  		1670&0.0 	& 		49844&0.0  \\ \hline
    \end{tabular}
    \caption{Comparison between the standard third-order methods (BCAGM3, BCAGM3+$\Psi$) 
    and the standard fourth-order methods (BCAGM, BCAGM+$\Psi$ \cite{Quynh2015}) on different datasets.
    The first column (No.) shows the number of test cases where one algorithm achieves better matching score than the other.
    The second column (Avg(\%)) shows the average performance gain achieved by the better method in each case,
    which is computed using Eq. \eqref{eq:avg_gain}.
    }
    \label{tab:standard34}
\end{table*}

\begin{table*}
    \centering
    \renewcommand{\arraystretch}{1.3}
    \begin{tabular}{|c|c|c|c|c|c|c|c|c|}
	\hline
	& \multicolumn{2}{c|}{Synthetic} & \multicolumn{2}{c|}{House} & \multicolumn{2}{c|}{Face+Duck+Winebottle} \\
	\hline \hline
	\textbf{Comparison} & No. & Avg(\%) 		& No. & Avg(\%) 	& No. & Avg(\%) \\ \hline 
	\hline
	\rowcolor{Gray}
	Adapt-BCAGM3 $>$ Adapt-BCAGM & 		269&47.2	&  		3&1.6 		& 		6717&28.0  \\ \hline
	Adapt-BCAGM3 $<$ Adapt-BCAGM & 		239&51.6	&  		3&20.3 		& 		5621&23.0  \\ \hline
	Adapt-BCAGM3 $=$ Adapt-BCAGM & 		4592&0.0	&  		1674&0.0	& 		50662&0.0  \\ \hline
	\hline
	\rowcolor{Gray}
	Adapt-BCAGM3+IPFP $>$ Adapt-BCAGM+IPFP & 	207&2.3		&  		20&6.9		& 		10774&21.0  \\ \hline
	Adapt-BCAGM3+IPFP $<$ Adapt-BCAGM+IPFP & 	109&55.6 	&  		3&1.9 		& 		6318&26.0  \\ \hline
	Adapt-BCAGM3+IPFP $=$ Adapt-BCAGM+IPFP & 	4784&0.0	&  		1657&0.0 	& 		45908&0.0  \\ \hline
	\hline
	\rowcolor{Gray}
	Adapt-BCAGM3+MP $>$ Adapt-BCAGM+MP & 		217&39.0	&  		4&12.2 		& 		6311&29.0  \\ \hline
	Adapt-BCAGM3+MP $<$ Adapt-BCAGM+MP & 		177&59.0	&  		4&10.4 		& 		5603&26.0  \\ \hline
	Adapt-BCAGM3+MP $=$ Adapt-BCAGM+MP & 		4706&0.0	&  		1672&0.0	& 		51086&0.0  \\ \hline
    \end{tabular}
    \caption{
    Comparison between the adaptive third-order methods (Adapt-BCAGM3, Adapt-BCAGM3+$\Psi$) 
    and the adaptive fourth-order methods (Adapt-BCAGM, Adapt-BCAGM+$\Psi$) on different datasets.
    The first column (No.) shows the number of test cases where one algorithm achieves better matching score than the other.
    The second column (Avg(\%)) shows the average performance gain achieved by the better method in each case,
    which is computed using Eq. \eqref{eq:avg_gain}.}
    \label{tab:adapt34}
\end{table*}

This section presents a detailed comparison of our third-order approaches and fourth-order approaches.
In particular, we first compare the non-adaptive third-order algorithms 
(Algorithm \ref{algo:bcagm3} and Algorithm \ref{algo:bcagm3_psi})
to their non-adaptive fourth-order counterparts \cite{Quynh2015}.
Second, we extend the adaptive scheme derived for third-order tensors in Section \ref{sec:adapt_bcagm3} to fourth-order tensors,
and denote the corresponding adaptive fourth-order algorithms as Adapt-BCAGM, Adapt-BCAGM+IPFP and Adapt-BCAGM+MP 
with the same notation as before.
We then compare these algorithms with the adaptive third-order methods 
from Algorithm \ref{algo:adapt_bcagm3} and Algorithm \ref{algo:adapt_bcagm3_psi}.
The purpose is to test for advantages and disadvantages of third-order approaches versus fourth-order ones under both settings -- 
with and without adaptive updates. 

For each pair of algorithms, 
we report in  Table \ref{tab:standard34} the number of test cases where one algorithm performs better than the other one
in terms of matching score and how much improvement it achieves on average over the other method for the cases where the
performance is not equal.
More precisely, let $f_i$ and $g_i$ be the matching score of BCAGM3 and BCAGM respectively on the Synthetic dataset.
Then ``BCAGM3 $>$ BCAGM'' denotes the cases where BCAGM3 achieves a better objective value than BCAGM,
and we report this number in the first column (No.), that is, $\abs{\Setbar{i}{f_i > g_i}},$
and the average performance gain in the second column (Avg(\%)), that is,
\begin{equation}\label{eq:avg_gain}
    \frac{1}{\abs{\Setbar{i}{f_i > g_i}}} \sum\limits_{i:f_i > g_i} \frac{f_i - g_i}{g_i}.
\end{equation}
Table \ref{tab:standard34} shows that the standard fourth-order algorithms \cite{Quynh2015} 
have a better overall performance than the standard third-order ones in terms of matching score.
However, the differences are minor and obtaining better matching score does not necessarily imply the same for matching accuracy,
especially when the ground truth is not a global optimum of the hypergraph matching problem.
Fig. \ref{fig:exp_willow} (a) shows an example of this phenomenon
where third-order methods achieve better matching score than the fourth-order algorithms 
but their matching accuracy is very similar.

Table \ref{tab:adapt34} shows the same kind of comparison but for the adaptive schemes.
One can see that the situation has changed as the adaptive third-order schemes win more frequently 
over their fourth-order counterparts. 
Again, the differences are minor.
While on the synthetic dataset the adaptive fourth-order methods perform better, 
on the real dataset, it is the opposite.

All in all, one can state that the standard fourth-order algorithms are better than 
the standard third-order approaches in terms of achieving better matching score.
This indicates potential benefits of using fourth-order tensors as well as the lifting step as done in \cite{Quynh2015}.
However, the differences between the adaptive third- and fourth-order methods are negligible and are slightly on the side
of the adaptive \cblue{third-order} methods. As the third-order methods are two times faster (see next section), 
our recommendation is to use the adaptive third-order methods in practice.

\subsection{Running time}
Fig. \ref{fig:exp_runtime_ours_others} shows the running time of all hypergraph matching algorithms.
All our methods achieve competitive/smaller running time than the state-of-the-art, 
in particular, they are $4$ to $16$ times faster than TM while being competitive to RRWHM in case of BCAGM3+MP
or up to two times faster in case of BCAGM3 and Adapt-BCAGM3.
It can also be seen that the proposed third-order approaches are two times faster than their fourth-order counterparts on average.
\crevision{
Further experimental results on the running time of all higher-order algorithms can be found in 
Tab. \ref{tab:runtime_synthetic}, Tab. \ref{tab:runtime_house} and Tab. \ref{tab:runtime_willow} in the appendix.
}

\section{Demo of matching results}
\begin{figure}
    \subfloat[Image 1]{ \includegraphics[width=0.48\linewidth]{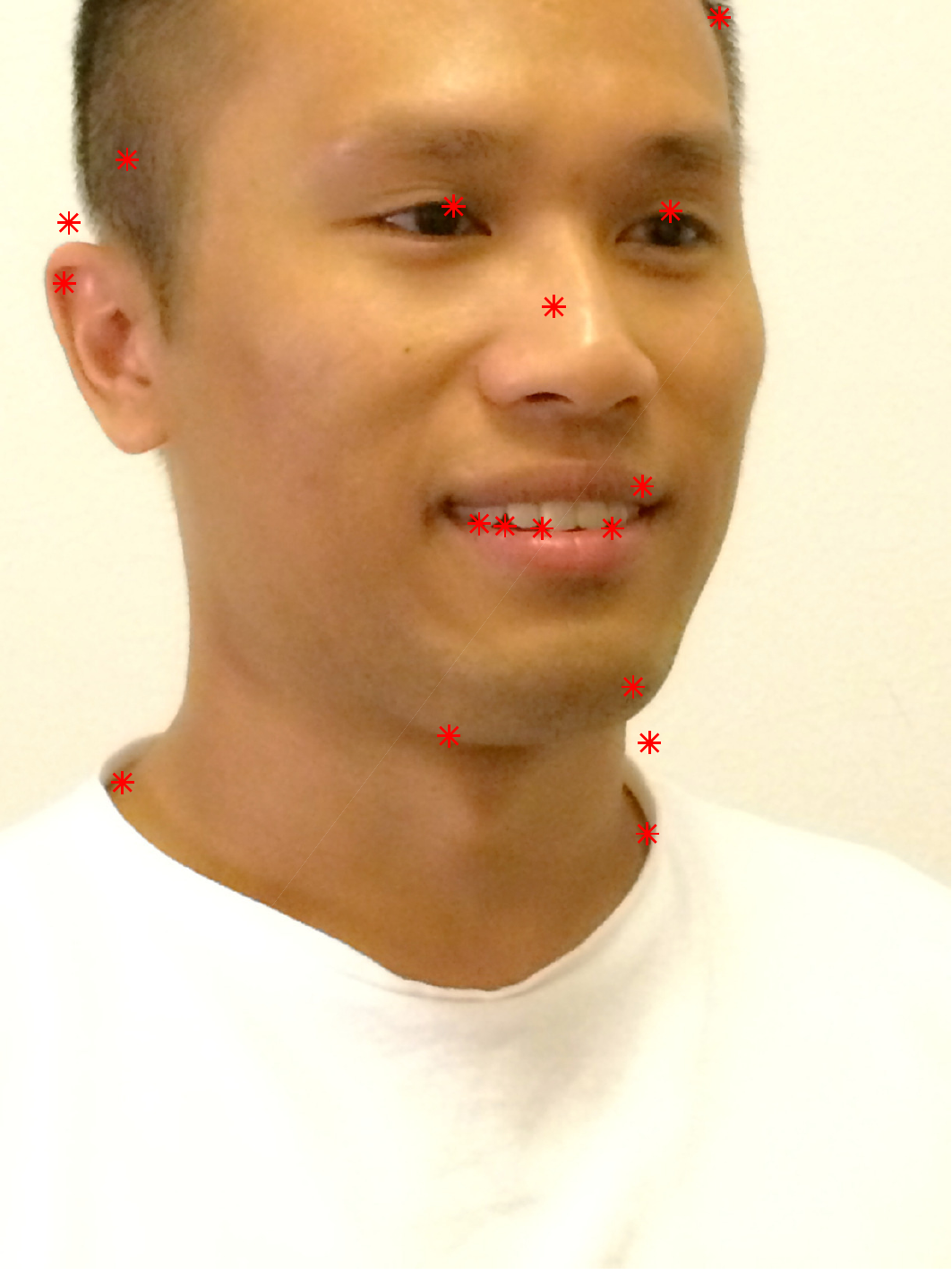} }  
    \subfloat[Image 2]{ \includegraphics[width=0.48\linewidth]{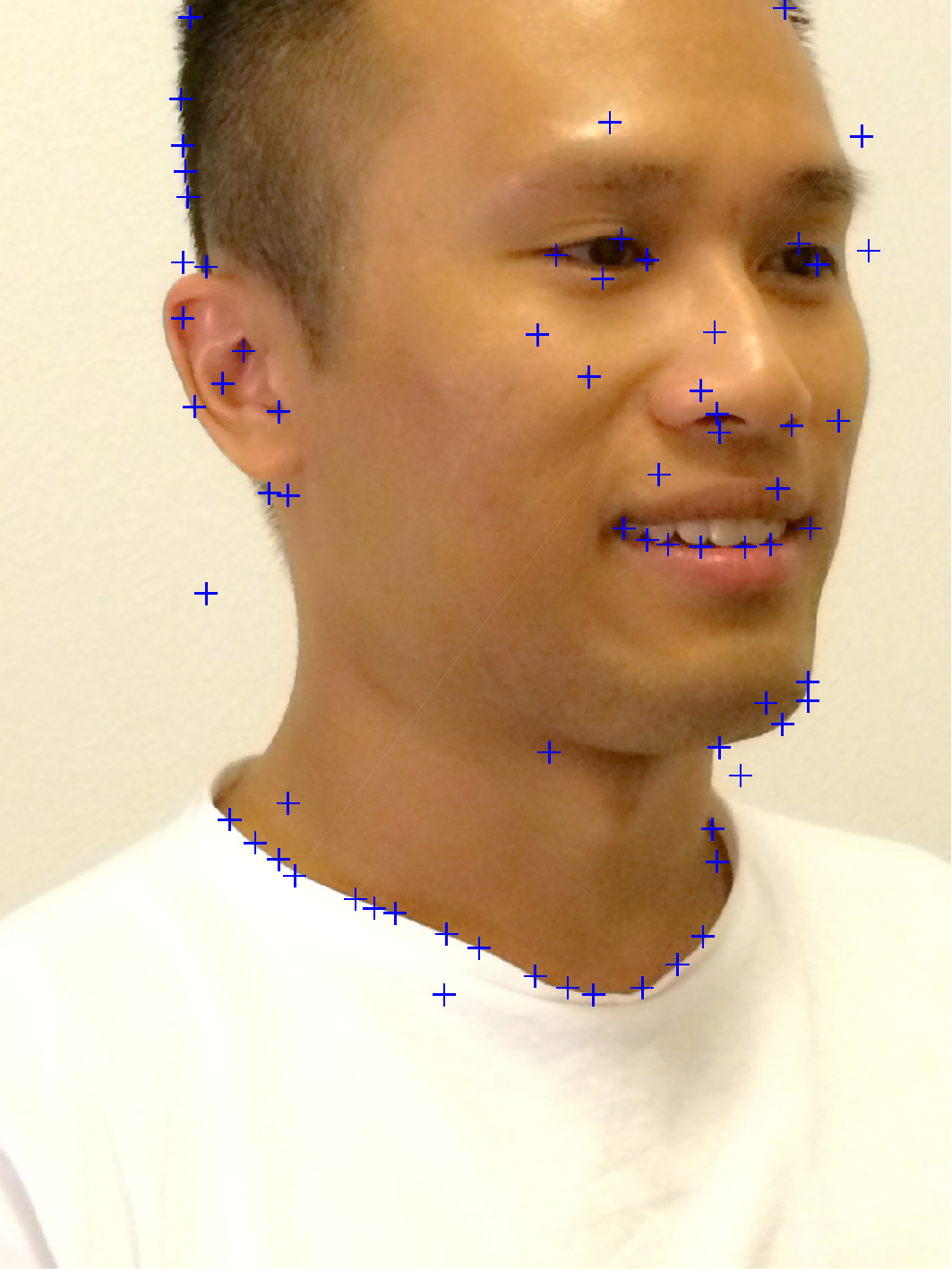} }  \\
    \subfloat[TM\cite{DucEtAl2011} (363)]{ \includegraphics[width=0.48\linewidth]{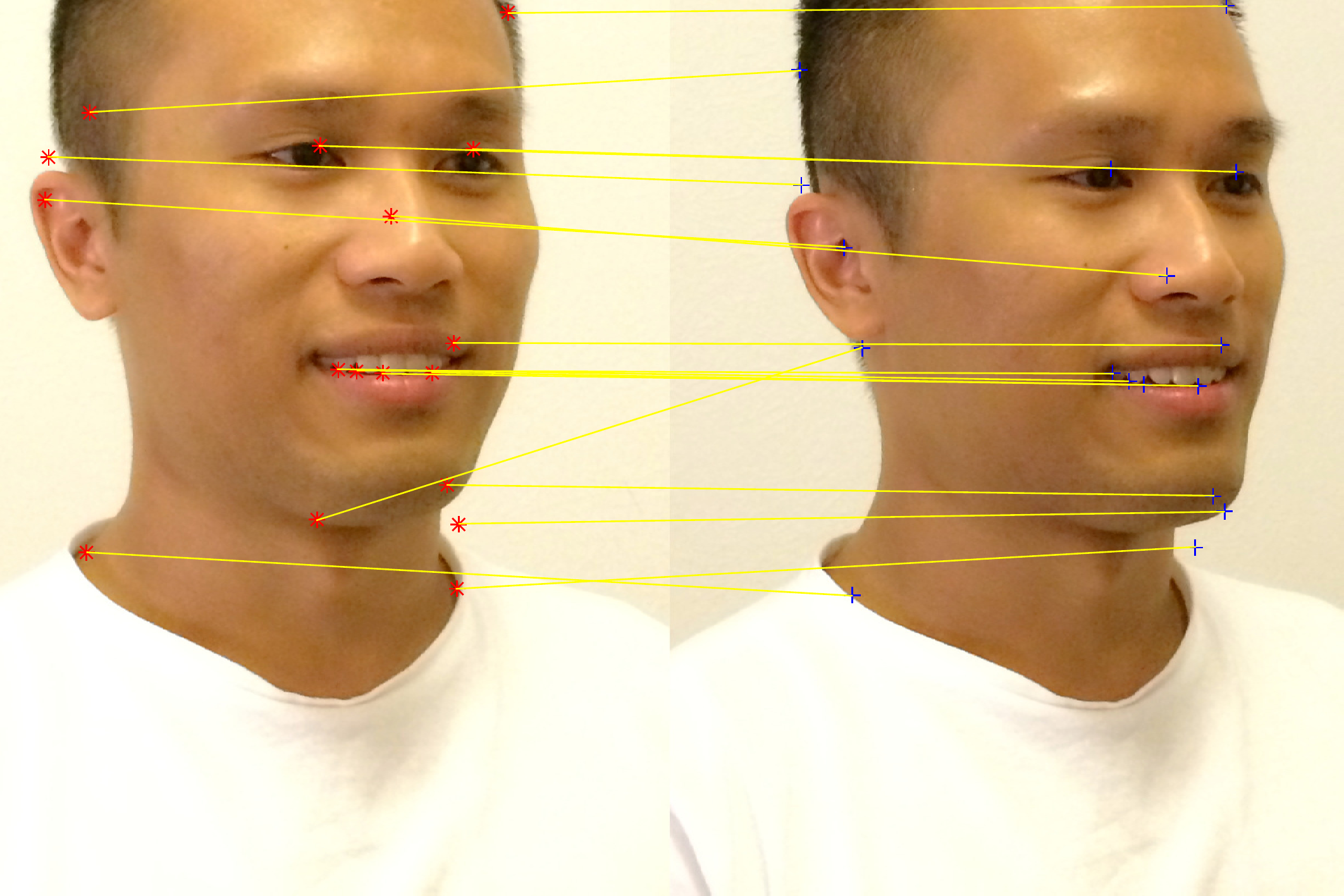} }  
    \subfloat[RRWHM\cite{LeeChoLee2011} (441)]{ \includegraphics[width=0.48\linewidth]{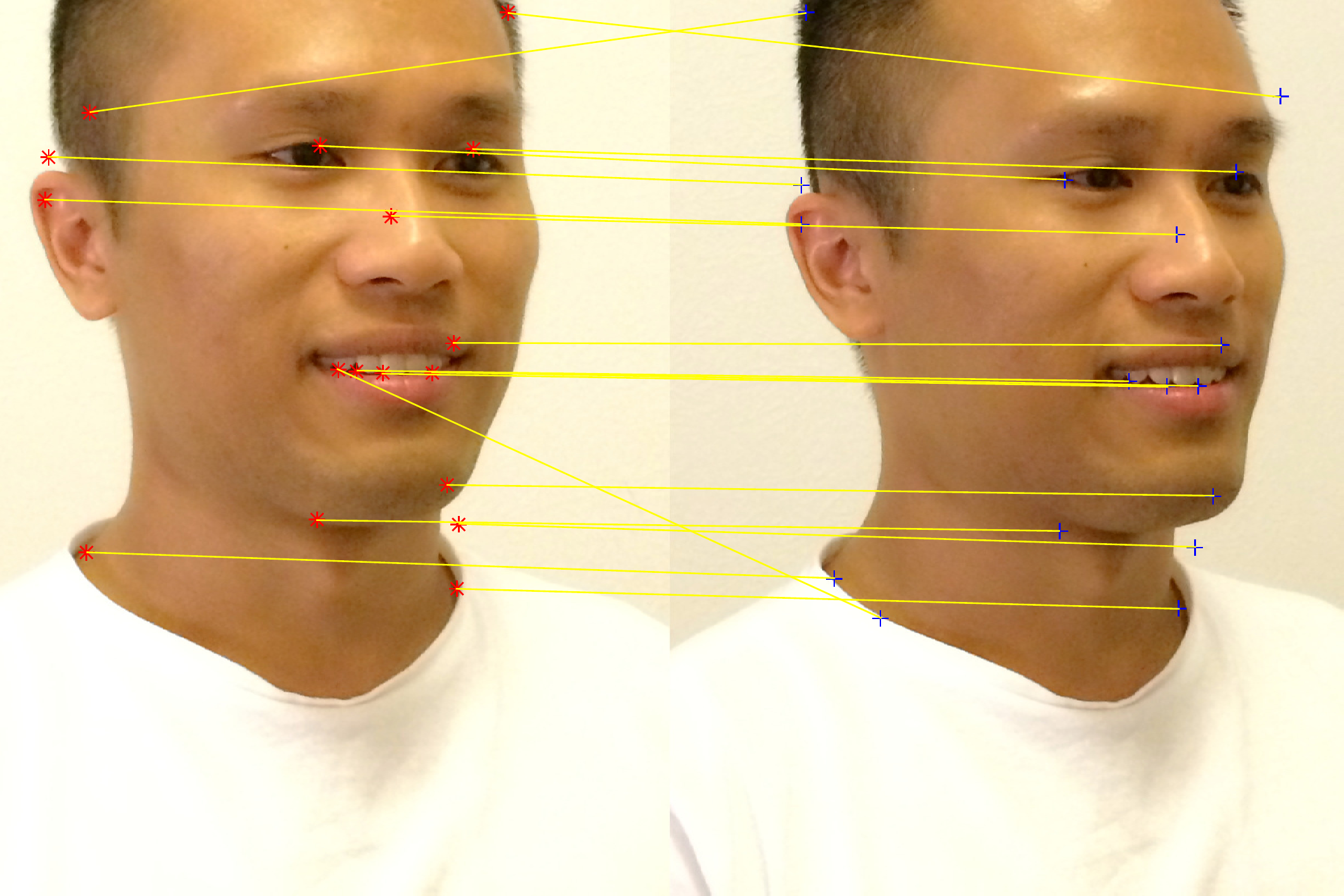} }  \\
    \subfloat[BCAGM\cite{Quynh2015} (599)]{ \includegraphics[width=0.48\linewidth]{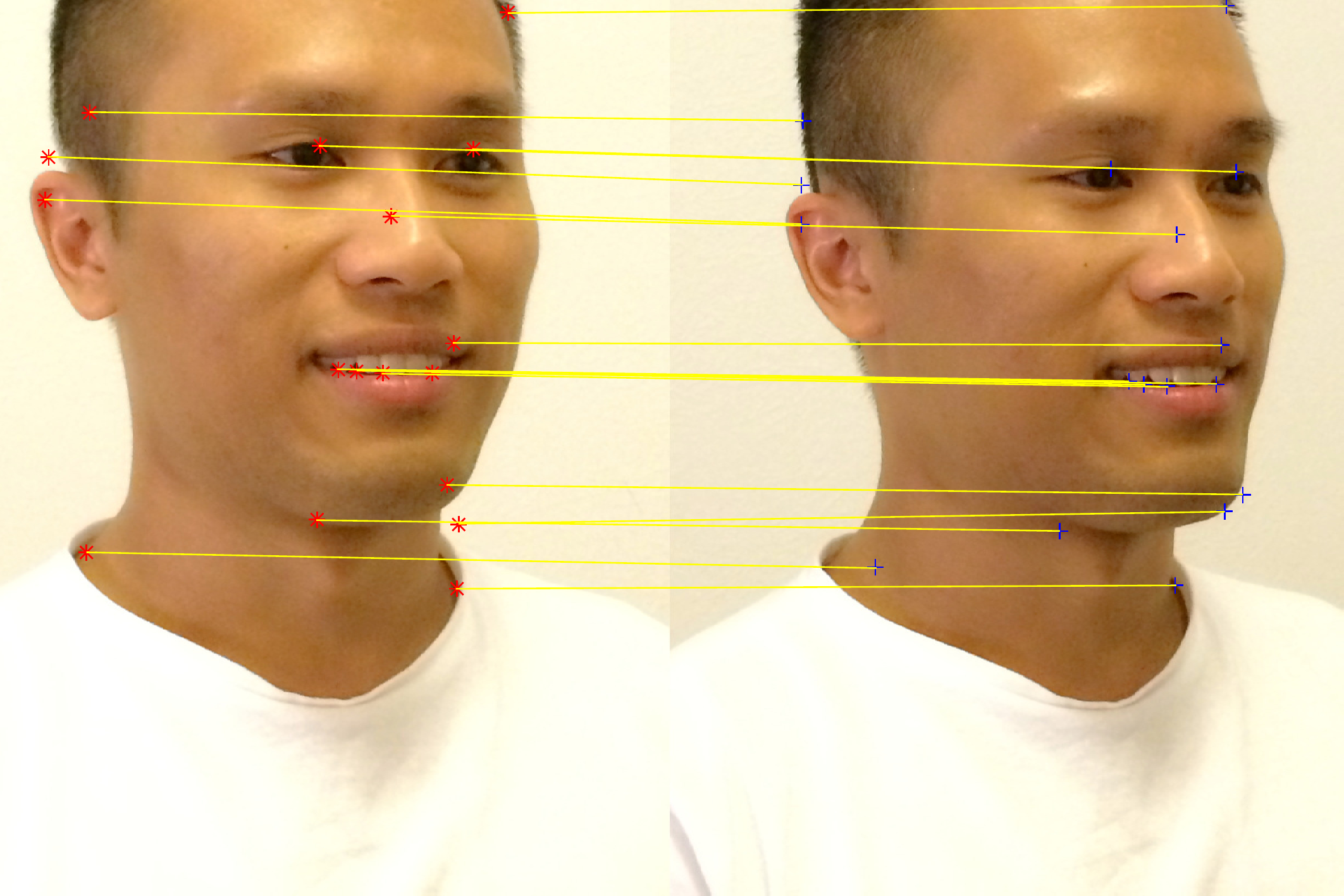} }  
    \subfloat[BCAGM+MP\cite{Quynh2015} (600)]{ \includegraphics[width=0.48\linewidth]{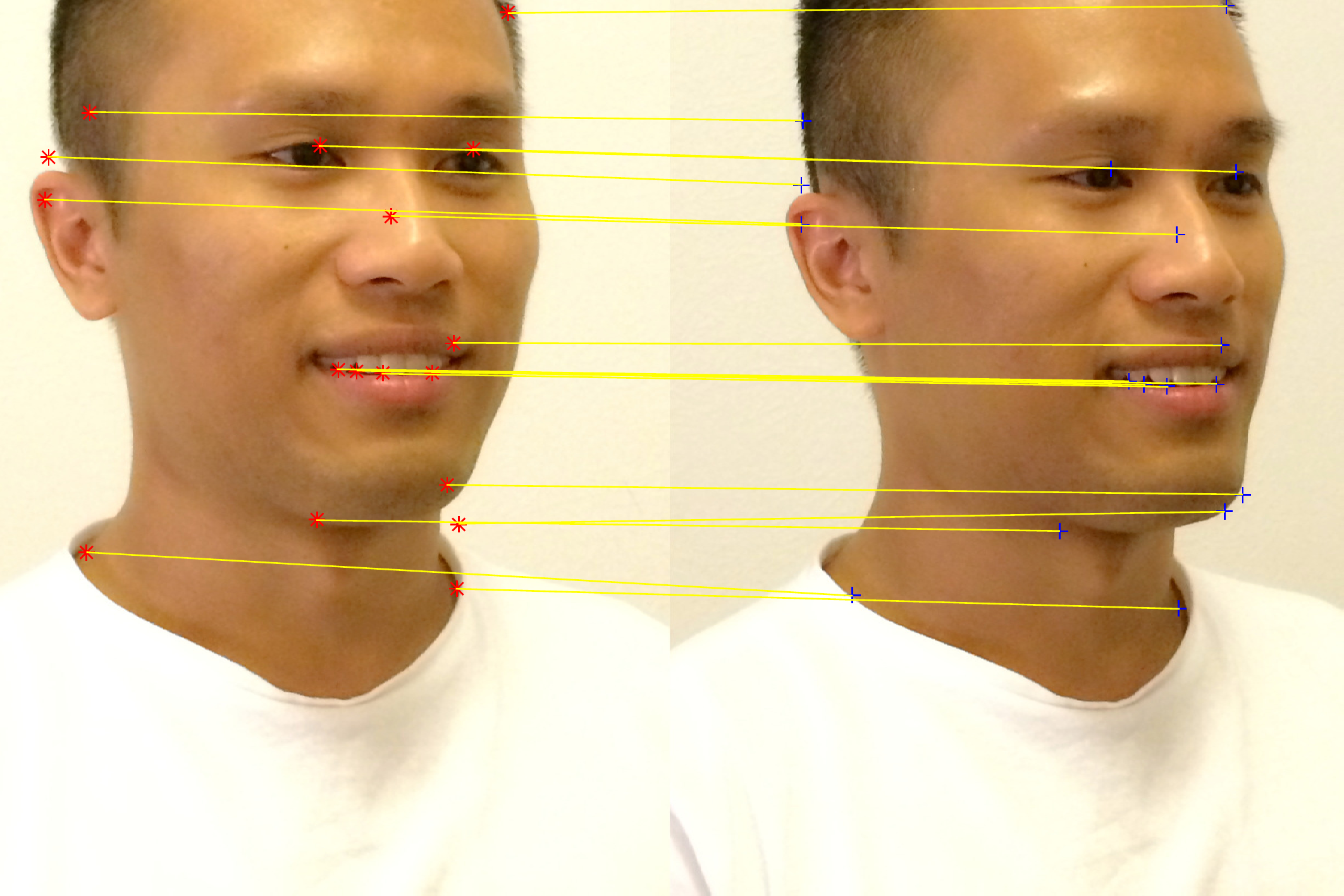} } \\  
    \subfloat[Adapt-BCAGM3 (600)]{ \includegraphics[width=0.48\linewidth]{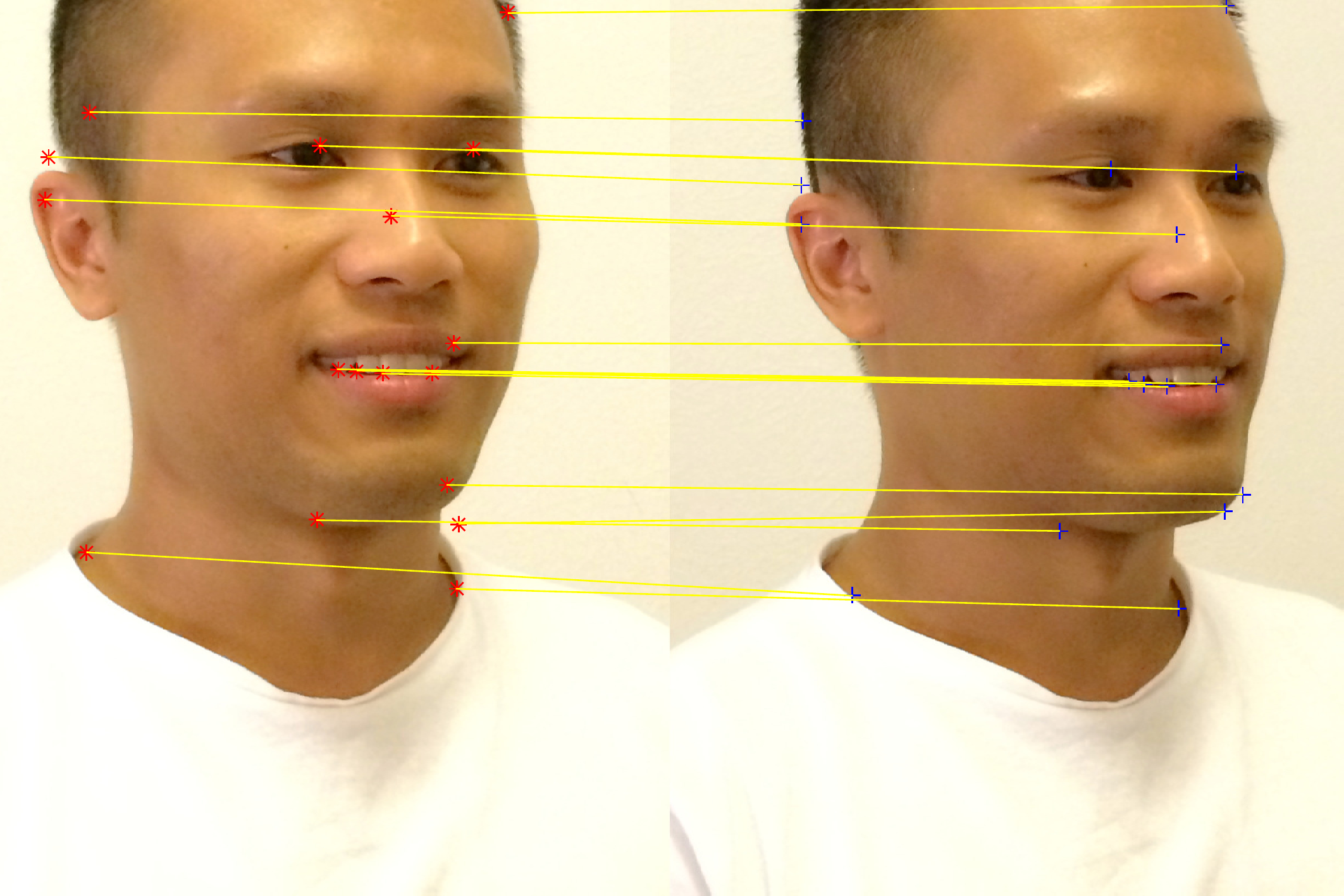} }  
    \subfloat[Adapt-BCAGM3+MP (600)]{ \includegraphics[width=0.48\linewidth]{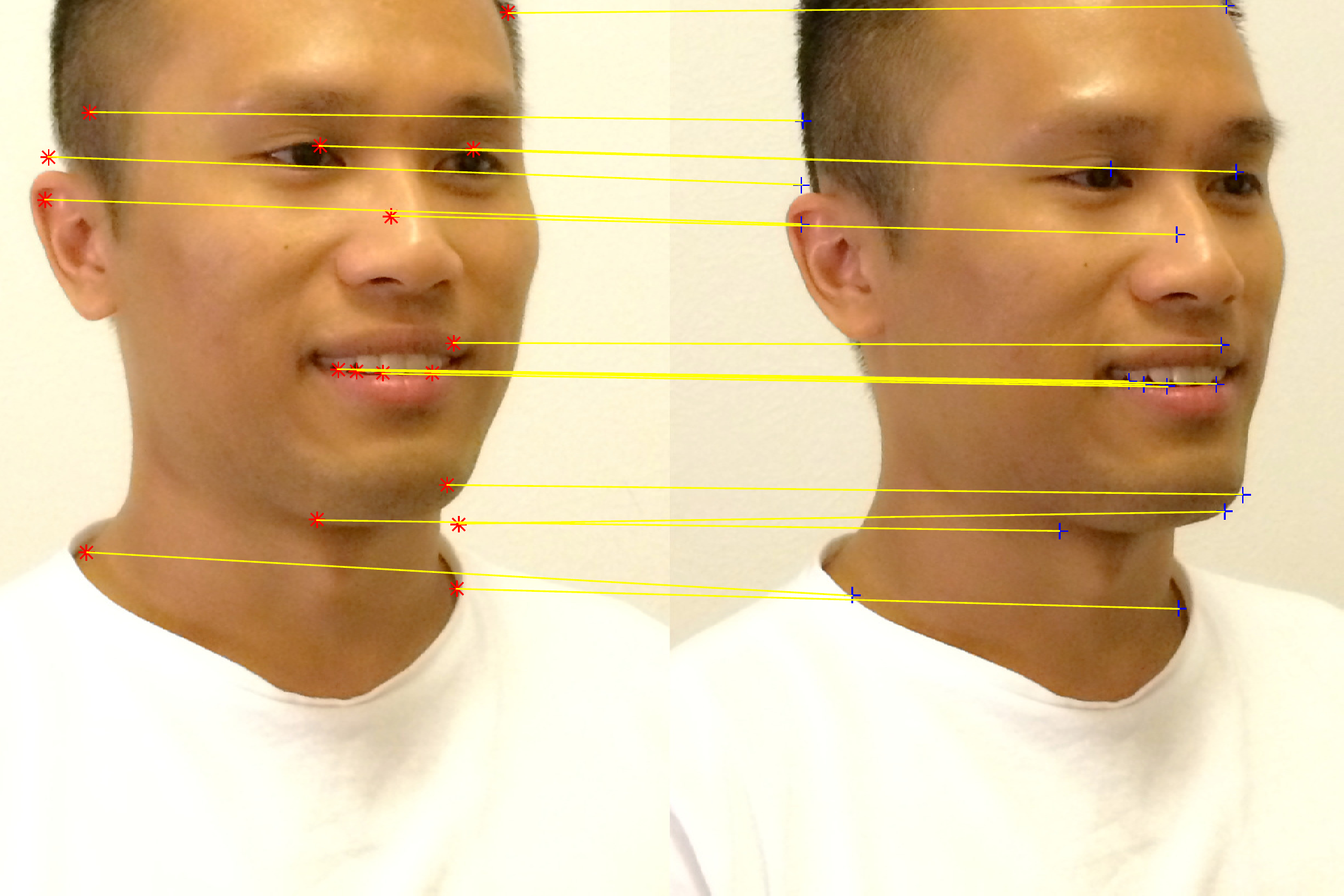} }
\caption{Demo of matching results. Matching score is reported for each method. (Best viewed in color.)}
\label{fig:example_matches}
\end{figure}

We show in this section a demo matching on realistic images.
Different from the previous sections, we run a SIFT detector on two realistic images, \cgram{hence}, 
there might be outliers in both point sets.
Since each point on one image might not have a corresponding point on the other image, 
the matching task \cgram{becomes} more difficult.
Fig. \ref{fig:example_matches} shows the matching results.
Our algorithms achieve more reasonable matches than the previous higher-order approaches
\crevision{while also achieving consistently better matching score than the others}.
Further examples can be found in the appendix.

\section{Conclusion and Outlook}\label{sec:concl}
In this work, we present a new optimization framework for hypergraph matching based on a multilinear reformulation.
In particular, we extend our theory from fourth-order tensors \cite{Quynh2015} to third-order tensors, 
proving \cgram{under mild conditions} the equivalence between the maximization of a third-order multilinear form and 
the maximization of its associated score function.
Based on that, we propose Algorithm \ref{algo:bcagm3} and Algorithm \ref{algo:bcagm3_psi} 
which both maximize the third-order multilinear form directly over the set of assignment constraints. 
These algorithms share the same attributes with their fourth-order counterparts \cite{Quynh2015}
in the sense that they achieve monotonic ascent in the original score function directly on the set of assignment matrices,
and both converge to a homogeneous solution in a finite number of steps.
Furthermore, we provide several sufficient conditions,
weaker than the convexity constraint used in \cite{Quynh2015},
to ensure the equivalence between the optimization problems.
We also propose two adaptive schemes which yield slight improvements over the standard approaches.
Overall, our algorithms, including the fourth-order and the third-order ones,
not only achieve superior performance in terms of matching score but also yield competitive or 
significantly better matching accuracy than previous work.
This holds even when there are a large number of outliers, deformation, scaling or other form of noise in the data.
Moreover, our methods are competitive or better in terms of runtime compared to previous work.
Our new adaptive third-order methods are competitive to 
the corresponding adaptive fourth-order ones but are on average two times faster and thus we recommend
\cgram{them} for usage in practice.

For future work it might be interesting to go beyond injective mappings which require
every point on one image to be mapped to exactly one point in the second image.
Non-injective mappings are of interest if not all points need to be matched
as for instance when outliers are present on both sides \cite{Suh2015, LianZhang2014}.
Our Theorem \ref{theo:main} allows \cgram{the use of} new constraint sets reflecting non-injective mappings, 
which is an interesting topic of further research. \crevision{Another possible generalization is 
to go beyond third-order (this paper) or fourth-order \cite{Quynh2015} hypergraphs which would 
require a generalization of Theorem \ref{theo:main}. At the moment it is unclear
to us if this is possible.}

\ifCLASSOPTIONcompsoc
  \section*{Acknowledgments}
\else
  \section*{Acknowledgment}
\fi
This work has been supported by the ERC grant NOLEPRO.
We sincerely thank the reviewers for their helpful comments 
and Cristian Caloian for helping us to correct grammatical mistakes.
\ifCLASSOPTIONcaptionsoff
  \newpage
\fi



%

{\small
\bibliographystyle{ieee}
\bibliography{regul}

\begin{thebibliography}{10}\itemsep=-1pt

\bibitem{willowObjectDataset}
http://www.di.ens.fr/willow/research/graphlearning/.

\bibitem{AroGlo2013}
C.~Arora and A.~Globerson.
\newblock Higher order matching for consistent multiple target tracking.
\newblock In {\em ICCV}, 2013.

\bibitem{AryEtal}
S.~Arya, D.~M. Mount, N.~S. Netanyahu, R.~Silverman, and A.~Wu.
\newblock An optimal algorithm for approximate nearest neighbor searching.
\newblock {\em Journal of the ACM}, 45:891--923, 1998.

\bibitem{Boros2002}
E.~Boros and P.~L. Hammer.
\newblock Pseudo-boolean optimization.
\newblock {\em Discrete Applied Mathematics}, 123:155--225, 2002.

\bibitem{BurAmiMar2012}
R.~E. Burkhard, M.~Dell'Amico, and S.~Martello.
\newblock {\em Assignment problems}.
\newblock SIAM, Philadelphia, 2012.

\bibitem{CarHancork2003}
M.~Carcassoni and E.~R. Hancock.
\newblock Spectral correspondence for point pattern matching.
\newblock {\em Pattern Recognition}, 36:193--204, 2003.

\bibitem{ChengEtal2012}
Z.-Q. Cheng, Y.~Chen, M.~R.R., Y.-K. Lai, and A.~Wang.
\newblock Supermatching: Feature matching using supersymmetric geometric
  constraints.
\newblock {\em IEEE Transactions on Visualization and Computer Graphics},
  10:1885--1894, 2012.

\bibitem{CheKel2010}
M.~Chertok and Y.~Keller.
\newblock Efficient high order matching.
\newblock {\em PAMI}, 32:2205--2215, 2010.

\bibitem{ChoEtal2013}
M.~Cho, K.~Alahari, and J.~Ponce.
\newblock Learning graphs to match.
\newblock In {\em ICCV}, 2013.

\bibitem{ChoLeeLee2010}
M.~Cho, J.~Lee, and K.~M. Lee.
\newblock Reweighted random walks for graph matching.
\newblock In {\em ECCV}, 2010.

\bibitem{ChoLee2012}
M.~Cho and K.~M. Lee.
\newblock Progressive graph matching: Making a move of graphs via probabilistic
  voting.
\newblock In {\em CVPR}, 2012.

\bibitem{ChoEtAl2014}
M.~Cho, J.~Sun, O.~Duchenne, and J.~Ponce.
\newblock Finding matches in a haystack: A max-pooling strategy for graph
  matching in the presence of outliers.
\newblock In {\em CVPR}, 2014.

\bibitem{ConteEtal2004}
D.~Conte, P.~Foggia, C.~Sansone, and M.~Vento.
\newblock Thirty years of graph matching.
\newblock {\em PAMI}, 18:265--298, 2004.

\bibitem{ShiEtal2007}
T.~Cour, P.~Srinivasan, and J.~Shi.
\newblock Balanced graph matching.
\newblock In {\em NIPS}, 2007.

\bibitem{DucEtAl2011}
O.~Duchenne, F.~Bach, I.~Kweon, and J.~Ponce.
\newblock A tensor-based algorithm for high-order graph matching.
\newblock {\em PAMI}, 33:2383--2395, 2011.

\bibitem{DucJouPon2011}
O.~Duchenne, A.~Joulin, and J.~Ponce.
\newblock A graph-matching kernel for object categorization.
\newblock In {\em ICCV}, 2011.

\bibitem{FilatovEtal1995}
A.~Filatov, A.~Gitis, and I.~Kil.
\newblock Graph-based handwritten digit string recognition.
\newblock In {\em Proc. Third Int'l Conf. Document Analysis and Recognition},
  1995.

\bibitem{FoggiaEtal2014}
P.~Foggia, G.~Percannella, and M.~Vento.
\newblock Graph matching and learning in pattern recognition in the last 10
  years.
\newblock {\em Pattern Recognition and Artificial Intelligence}, 28, 2014.

\bibitem{HartleyZiss1994}
R.~Hartley and A.~Zisserman.
\newblock Tutorial i: The algebraic approach to invariance.
\newblock In {\em ESPRIT Workshop}, 1994.

\bibitem{HaysEtal2006}
J.~Hays, M.~Leordeanu, A.~A. Efros, and Y.~Liu.
\newblock Discovering texture regularity as a higher-order correspondence
  problem.
\newblock In {\em ECCV}, 2006.

\bibitem{HopWong1974}
J.~E. Hopcroft and J.~K. Wong.
\newblock Linear time algorithm for isomorphism of planar graphs (preliminary
  report).
\newblock {\em Proc. Sixth Ann. ACM Symp. Theory of Computing}, pages 172--184,
  1974.

\bibitem{Kolmogorov2007}
V.~Kolmogorov and C.~Rother.
\newblock Minimizing nonsubmodular functions with graph cuts-a review.
\newblock {\em PAMI}, 29:1274--1279, 2007.

\bibitem{Kuh1955}
H.~W. Kuhn.
\newblock The hungarian method for the assignment problem.
\newblock {\em Naval Research Logistics Quarterly}, 2:83--97, 1955.

\bibitem{LeeChoLee2010}
J.~Lee, M.~Cho, and K.~M. Lee.
\newblock A graph matching algorithm using data-driven markov chain monte carlo
  sampling.
\newblock In {\em ICPR}, 2010.

\bibitem{LeeChoLee2011}
J.~Lee, M.~Cho, and K.~M. Lee.
\newblock Hyper-graph matching via reweighted random walks.
\newblock In {\em CVPR}, 2011.

\bibitem{LeeLiu1999}
R.~Lee and J.~Liu.
\newblock An oscillatory elastic graph matching model for recognition of
  offline handwritten chinese characters.
\newblock In {\em Knowledge-Based Intelligent Information Engineering Systems},
  1999.

\bibitem{LeoHeb2005}
M.~Leordeanu and M.~Hebert.
\newblock A spectral technique for correspondence problems using pairwise
  constraints.
\newblock In {\em ICCV}, 2005.

\bibitem{LeoHeb2009}
M.~Leordeanu and M.~Hebert.
\newblock Unsupervised learning for graph matching.
\newblock In {\em CVPR}, 2009.

\bibitem{LeoHebSuk2009}
M.~Leordeanu, M.~Hebert, and R.~Sukthankar.
\newblock An integer projected fixed point method for graph matching and map
  inference.
\newblock In {\em NIPS}, 2009.

\bibitem{LianZhang2014}
W.~Lian and L.~Zhang.
\newblock Point matching in the presence of outliers in both point sets: A
  concave optimization approach.
\newblock In {\em CVPR}, 2014.

\bibitem{LiuEtal2012}
Z.-Y. Liu, H.~Qiao, and L.~Xu.
\newblock An extended path following algorithm for graph-matching problem.
\newblock {\em PAMI}, 34:1451--1456, 2012.

\bibitem{Low1999}
D.~G. Lowe.
\newblock Object recognition from local scale-invariant features.
\newblock In {\em ICCV}, 1999.

\bibitem{LuoHancock2000}
B.~Luo and E.~Hancock.
\newblock Alignment and correspondence using singular value decomposition.
\newblock {\em Advances in Pattern Recognition}, 1876:226--235, 2000.

\bibitem{Joa2003}
J.~Maciel and J.~P. Costeira.
\newblock A global solution to sparse correspondence problems.
\newblock {\em PAMI}, 25:187--199, 2003.

\bibitem{MundyEtal1993}
J.~L. Mundy, A.~Zisserman, and D.~Forsyth.
\newblock {\em Applications of Invariance in Computer Vision}.
\newblock Springer, 1993.

\bibitem{Quynh2015}
Q.~Nguyen, A.~Gautier, and M.~Hein.
\newblock A flexible tensor block coordinate ascent scheme for hypergraph
  matching.
\newblock In {\em CVPR}, 2015.

\bibitem{SchSchnor2005}
C.~Schellewald and C.~Schn\"orr.
\newblock Probabilistic subgraph matching based on convex relaxation.
\newblock In {\em EMMCVPR}, 2005.

\bibitem{SharEtal2011}
A.~Sharma, R.~Horaud, J.~Cech, and E.~Boyer.
\newblock Topologically- robust 3d shape matching based on diffusion geometry
  and seed growing.
\newblock In {\em CVPR}, 2011.

\bibitem{SinghEtal2007}
R.~Singh, J.~Xu, and B.~Berger.
\newblock Pairwise global alignment of protein interaction networks by matching
  neighborhood topology.
\newblock {\em Proc. 11th Int’l Conf. Research in Computational Molecular
  Biology}, pages 16--31, 2007.

\bibitem{Suh2015}
Y.~Suh and K.~M. Lee.
\newblock Subgraph matching using compactness prior for robust feature
  correspondence.
\newblock In {\em CVPR}, 2015.

\bibitem{Taylor2002}
W.~Taylor.
\newblock Protein structure comparison using bipartite graph matching and its
  application to protein structure classification.
\newblock {\em Molecular and Cellular Proteomics}, 1:334--339, 2002.

\bibitem{TorEtal2008}
L.~Torresani, V.~Kolmogorov, and C.~Rother.
\newblock Feature correspondence via graph matching: Models and global
  optimization.
\newblock In {\em ECCV}, 2008.

\bibitem{WangHancock2006}
H.~F. Wang and E.~R. Hancock.
\newblock Correspondence matching using kernel principal components analysis
  and label consistency constraints.
\newblock {\em Pattern Recognition}, 39:1012--1025, 2006.

\bibitem{WangEtal2004}
Y.~Wang, F.~Makedon, and J.~Ford.
\newblock A bipartite graph matching framework for finding correspondences
  between structural elements in two proteins.
\newblock In {\em Proc. IEEE 26th Ann. Int’l Conf. Eng. in Medicine and
  Biology Society}, 2004.

\bibitem{Yan2015}
J.~Yan, C.~Zhang, H.~Zha, W.~Liu, X.~Yang, and S.~M. Chu.
\newblock Discrete hyper-graph matching.
\newblock In {\em CVPR}, 2015.

\bibitem{ZasEtl2009}
M.~Zaslavskiy, F.~Bach, and J.~Vert.
\newblock A path following algorithm for the graph matching problem.
\newblock {\em PAMI}, 31:2227--2242, 2009.

\bibitem{ZasSha2008}
R.~Zass and A.~Shashua.
\newblock Probabilistic graph and hypergraph matching.
\newblock In {\em CVPR}, 2008.

\bibitem{Zeng2010}
Y.~Zeng, C.~Wang, Y.~Wang, X.~Gu, D.~Samaras, and N.~Paragios.
\newblock Dense non-rigid surface registration using high-order graph matching.
\newblock In {\em CVPR}, 2010.

\bibitem{ZhoTor2012}
F.~Zhou and F.~{De la Torre}.
\newblock Factorized graph matching.
\newblock In {\em CVPR}, 2012.

\bibitem{ZhoTor2013}
F.~Zhou and F.~{De la Torre}.
\newblock Deformable graph matching.
\newblock In {\em CVPR}, 2013.

\end{thebibliography}
}

%
\vspace*{-3\baselineskip}
\begin{IEEEbiography}[{\includegraphics[width=1in,height=1.25in,clip,keepaspectratio]{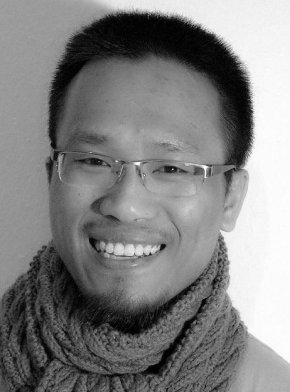}}]
{Quynh Nguyen}
is currently a PhD student in the Machine Learning Group of Saarland University.
He received his BSc degree in Information Technology from Vietnam National University in Hanoi in 2011.
In 2014, he got his Msc degree in Computer Science from Saarland University 
through the support of the IMPRS for Computer Science in Saarbruecken, Germany.
His research interests are in machine learning, optimization and computer vision.
\end{IEEEbiography}

\vspace*{-3\baselineskip}
\begin{IEEEbiography}[{\includegraphics[width=1in,height=1.25in,clip,keepaspectratio]{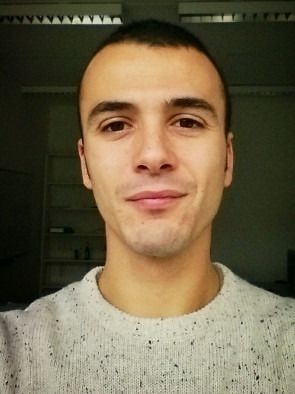}}]
{Francesco Tudisco} is a Postdoctoral fellow in Mathematics at Saarland University, within the Machine Learning Group.  
He received his diploma and his Ph.D. in Mathematics from the University of Rome "Tor Vergata". 
His main scientific interests include Operator Theory, Numerical Linear Algebra, Network Analysis and Numerical Optimization. 
\end{IEEEbiography}

\vspace*{-3\baselineskip}
\begin{IEEEbiography}[{\includegraphics[width=1in,height=1.25in,clip,keepaspectratio]{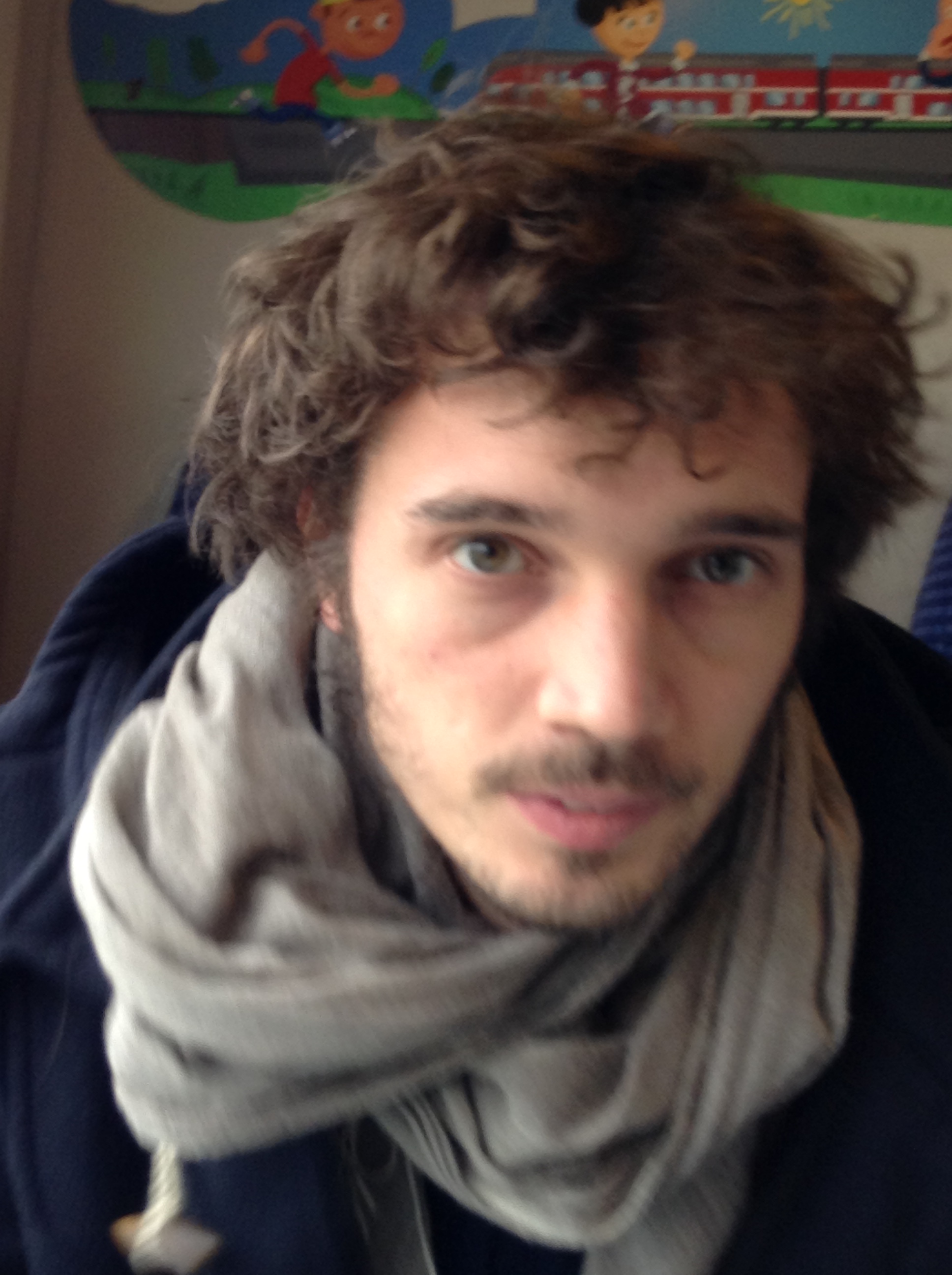}}]
{Antoine Gautier} is a PhD student in the machine learning group of Saarland university since 2014. 
He received a B.Sc. (2011) and a M.Sc. (2013) degree in applied mathematics from EPFL, Lausanne, Switzerland. 
In 2010, he spent a year at the Humboldt university of Berlin for an Erasmus exchange. 
His main research interests are higher-order structures, optimization, nonlinear spectral theory 
and their applications in computational science.
\end{IEEEbiography}

\vspace*{-3\baselineskip}
\begin{IEEEbiography}[{\includegraphics[width=1in,height=1.25in,clip,keepaspectratio]{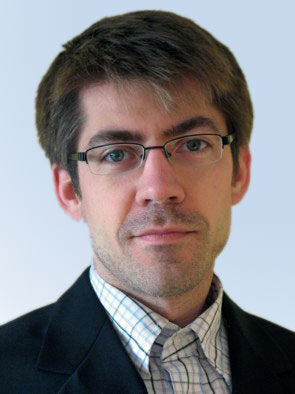}}]
{Matthias Hein} is professor of mathematics and computer science at Saarland University.
He received his diploma in physics from the University of T\"ubingen in 2002 and
his PhD in computer science from the University of Darmstadt in 2005.
His research interests are in machine learning, statistics and optimization and
the application of machine learning to problems in bioinformatics, computer vision
and other fields in computer science and the natural sciences. 
\end{IEEEbiography}

\appendices
\section{}
\begin{customTheorem}{3.3}
    Let $D\subset\RR^n$ be a compact set,
    $\MF: \RR^n\times\RR^n\times\RR^n \to \RR$ a symmetric third-order multilinear form and 
    \crevision{
    $S: \RR^n \to \RR$
    }
    the associated third-order score function
    such that $S(\bx)=F(\bx,\bx,\bx),$
    then we have the following chain of implications:  \crevision{1) $\Rightarrow$ 2) $\Rightarrow$ 3) $\Rightarrow$ 4).}
    \begin{enumerate}
	\item $\nabla^2 S(\bx)$ is positive semidefinite for all $\bx\in D.$
	\item $\MF(\bx,\by-\bz,\by-\bz) \geq 0$ for all $\bx,\by,\bz \in D.$
	\item 
	It holds for all $\bx,\by,\bz\in D$
	\begin{equation*}
	    \MF(\bx,\by,\bz) \leq \max\limits_{\bu\in\set{\bx,\by,\bz}} \MF(\bu,\bu,\bu) .
	\end{equation*}
	\item The optimization of the multilinear form is equivalent to the optimization of its associated score function
	$$
	    \!\!\!\!\max\limits_{\bx,\by,\bz\in D} \MF(\bx,\by,\bz) = \max\limits_{\bx,\by\in D} \MF(\bx,\bx,\by) 
	    = \max\limits_{\bx\in D} \MF(\bx,\bx,\bx) .
	$$
    \end{enumerate}
\end{customTheorem}
\begin{proof}
    \setitemize[0]{leftmargin=10pt,itemindent=5pt}
    \begin{itemize}
     \item \crevision{$1) \implies 2):$} 
    We first recall the definition of the score function as
    $$
	S(\bx) = \MF(\bx,\bx,\bx) = \sum_{i,j,k=1}^n \FC_{ijk} \crevision{\bx_i \bx_j \bx_k}.
    $$
    One can compute the Hessian of $S$ at $\bx$ as follows
    $$
	\left(\nabla^2 S(\bx)\right)_{ij} = 6 \sum_{k=1}^n \FC_{ijk} \bx_k \quad \forall \, i,j=1,\ldots,n.
    $$
    Using our notation, the Hessian can be rewritten as
    $$
	\nabla^2 S(\bx) = 6\,\MF(\bx,\,\cdot\,,\,\cdot\,).
    $$
    The Hessian $\nabla^2 S(\bx)$ is is positive semidefinite for all $\bx\in D$
    if and only if $\inner{\bt, \nabla^2 S(\bx) \, \bt} \geq 0$ for all $\bx\in D, \bt\in \RR^n$, 
    which is equivalent to $\MF(\bx,\bt,\bt)\geq 0$ for all $\bx\in D, \bt\in \RR^n$ and thus,
    $$
	\MF(\bx,\by-\bz,\by-\bz) \geq 0 \quad \forall\, \bx,\by,\bz\in D .
    $$
    
    \item \crevision{ $2) \implies 3):$ }
    Given $\bx, \by, \bz \in D$, the statement 2) implies that 
    $\MF(\bx,\bx-\by,\bx-\by) \geq 0$ and $\MF(\by,\bx-\by,\bx-\by) \geq 0$, thus,
    \begin{eqnarray*}
	0 &\leq& \MF(\bx,\bx-\by,\bx-\by) + \MF(\by,\bx-\by,\bx-\by) \nonumber \\
	&=& \MF(\bx,\bx,\bx) + \MF(\by,\by,\by) - \MF(\bx,\bx,\by) - \MF(\bx,\by,\by) \\ 
    \end{eqnarray*}
    By switching the roles of $\bx,\by,\bz$, we can derive the following inequalities
    \begin{eqnarray*}
	\MF(\bx,\bx,\by) + \MF(\bx,\by,\by) &\leq& \MF(\bx,\bx,\bx) + \MF(\by,\by,\by) \\
	\MF(\bx,\bx,\bz) + \MF(\bx,\bz,\bz) &\leq& \MF(\bx,\bx,\bx) + \MF(\bz,\bz,\bz) \\
	\MF(\by,\by,\bz) + \MF(\by,\bz,\bz) &\leq& \MF(\by,\by,\by) + \MF(\bz,\bz,\bz) \\
    \end{eqnarray*}
    Summing up these equations gives us
    \begin{eqnarray}
	&&\MF(\bx,\bx,\by) + \MF(\bx,\by,\by) + \MF(\bx,\bx,\bz) + \nonumber\\
	&&\MF(\bx,\bz,\bz) + \MF(\by,\by,\bz) + \MF(\by,\bz,\bz) \nonumber\\
	&\leq& 2\MF(\bx,\bx,\bx) + 2\MF(\by,\by,\by) + 2\MF(\bz,\bz,\bz) . \label{eq_proof:1}
    \end{eqnarray}
    The statement 2) also implies the following inequalities
    \begin{eqnarray*}
	0 \leq \MF(\bx,\by-\bz,\by-\bz) \\
	0 \leq \MF(\by,\bz-\bx,\bz-\bx) \\
	0 \leq \MF(\bz,\bx-\by,\bx-\by)
    \end{eqnarray*}
    which can be expanded as
    \begin{eqnarray*}
	2 \MF(\bx,\by,\bz) \leq \MF(\bx,\by,\by) + \MF(\bx,\bz,\bz) \\
	2 \MF(\by,\bz,\bx) \leq \MF(\by,\bz,\bz) + \MF(\by,\bx,\bx) \\
	2 \MF(\bz,\bx,\by) \leq \MF(\bz,\bx,\bx) + \MF(\bz,\by,\by) .
    \end{eqnarray*}
    By summing up these inequalities and taking into account the symmetry of $\MF$, one obtains
    \begin{eqnarray}
	6\,\MF(\bx,\by,\bz) &\leq& \MF(\bx,\by,\by) + \MF(\bx,\bz,\bz) + \MF(\by,\bx,\bx) +  \nonumber \\
	&&\MF(\by,\bz,\bz) + \MF(\bz,\bx,\bx) + \MF(\bz,\by,\by). \label{eq_proof:2}
    \end{eqnarray}
    Combining \eqref{eq_proof:1} and \eqref{eq_proof:2} to achieve
    \begin{eqnarray*}
	6\,\MF(\bx,\by,\bz) \leq  2\MF(\bx,\bx,\bx) + 2\MF(\by,\by,\by) + 2\MF(\bz,\bz,\bz) ,
    \end{eqnarray*} which implies
    \begin{eqnarray*}
	\MF(\bx,\by,\bz) \leq \max \set{\MF(\bx,\bx,\bx),\MF(\by,\by,\by),\MF(\bz,\bz,\bz)} .
    \end{eqnarray*}
    
    \item \crevision{ $3) \implies 4):$ }
    From the statement 3)
    $$
	\MF(\bx,\by,\bz) \leq \max\limits_{\bu\in\set{\bx,\by,\bz}} \MF(\bu,\bu,\bu),
    $$ 
    one takes the maximum of both sides over $D$ to achieve
    \begin{eqnarray*}
      \max\limits_{\bx,\by,\bz \in D} \MF(\bx,\by,\bz) &\leq& 
      \max\limits_{\bx,\by,\bz\in D} \left\lbrace \max\limits_{\bu\in\set{\bx,\by,\bz}} \MF(\bu,\bu,\bu) \right\rbrace \\
      &=& \max\limits_{\bx \in D} \MF(\bx,\bx,\bx) .
    \end{eqnarray*} 
    However, it holds for any compact set $D \subset \RR^n$ that
    \begin{eqnarray*}
	\max\limits_{\bx,\by,\bz \in D} \MF(\bx,\by,\bz)
	\geq \max\limits_{\bx,\by \in D} \MF(\bx,\bx,\by) 
	\geq \max\limits_{\bx \in D} \MF(\bx,\bx,\bx) 
    \end{eqnarray*}
    thus, the equality must hold for the whole chain, that is,
    \begin{equation*}
	\max\limits_{\bx,\by,\bz \in D} \MF(\bx,\by,\bz) 
	= \max\limits_{\bx,\by \in D} \MF(\bx,\bx,\by) 
	= \max\limits_{\bx \in D} \MF(\bx,\bx,\bx) .
    \end{equation*} 
    \end{itemize}
\end{proof}
\begin{customProposition}{3.4}
    Let $\MF: \RR^n\times \RR^n \times \RR^n \to \RR$ be a symmetric third-order multilinear form
    and $\MG: \RR^n\times \RR^n \times \RR^n \to \RR$ as defined in Eq. \eqref{eq:mod_mf}.
    We consider the new multilinear form $\MF_\alpha$ defined by 
    $$
	\MF_\alpha(\bx,\by,\bz) = \MF(\bx,\by,\bz) + \alpha \, \MG(\bx,\by,\bz).
    $$
    For all $\alpha \geq \frac{27}{4} \max\limits_{i=1,2,\ldots,n} \sqrt{\sum_{j,k=1}^n (\FC_{ijk})^2 }$ 
    it holds
    \begin{enumerate}
	\item $\MF_\alpha$ is a symmetric third-order multilinear form.
	\item $\nabla^2 S_\alpha(\bx)$ is positive \cred{semidefinite} for all $\bx\in M.$
	\item The new problem is equivalent to the original one
	$$
	    \argmax_{\bx\in M} S_\alpha(\bx) = \argmax_{\bx\in M} S(\bx) .
	$$
    \end{enumerate}
\end{customProposition}
\begin{proof}
    \begin{enumerate}
	\item 
	Since $\MF$ and $\MG$ are symmetric multilinear forms, $\MF_\alpha$ must be symmetric as well.
	
	\item In the following, we define a new tensor $\GC_\eps$ as
	$$ \GC_\eps = \sum_{i=1}^n \be_i^\eps \otimes \be_i^\eps \otimes \be_i^\eps $$    
	where $\be_i^\eps = \eps\ones + (1-\eps)\be_i$ for $i=1,2,\ldots,n$ and $0 < \eps < 1.$ 
	Let $\MG_\eps$ be the associated multilinear form
	\begin{eqnarray}\label{eq:proof_Geps}
	    \MG_\eps(\bx,\by,\bz) &=& \sum_{p,q,r=1}^n {(\GC_\eps)}_{pqr} \bx_p \by_q \bz_r  \nonumber \\
	    &=& \sum_{i=1}^n \inner{\be_i^\eps, \bx} \inner{\be_i^\eps, \by} \inner{\be_i^\eps, \bz} ,
	\end{eqnarray}
	and the new score function is defined as
	$$
	    S_{\alpha,\eps}(\bx) = \MF(\bx,\bx,\bx) + \alpha \, \MG_\eps(\bx,\bx,\bx) .
	$$
	One first observes that the new score function is related with the original one via $S_\alpha=S_{\alpha,\frac{1}{3}}.$
	For any given $0<\eps<1$, we will prove that $\alpha \geq \frac{1}{\eps(1-\eps)^2} \max\limits_{i=1,2,\ldots,n} \sqrt{\sum_{j,k=1}^n (\FC_{ijk})^2 }$ 
	is a sufficient condition for $\nabla^2 S_{\alpha,\eps}(\bx)$ to be positive semidefinite at every $\bx\in M.$
	Indeed, $\nabla^2 S_{\alpha,\eps} (\bx) = 6 \left(\MF(\bx,\,\cdot\,,\,\cdot\,) + \alpha \MG_\eps(\bx,\,\cdot\,,\,\cdot\,) \right) \succeq 0$ for all $\bx\in M$ if and only if
	$\MF(\bx,\by,\by) + \alpha \MG_\eps(\bx,\by,\by) \geq 0$ for all $\bx\in M$ and $\by \in \RR^n$. 
	One has
	\begin{eqnarray*}
	    \abs{\MF(\bx,\by,\by)} &=& \abs{\sum_{i,j,k=1}^n \FC_{ijk} \bx_i \by_j \by_k} \\
	    &=& \abs{\sum_{i=1}^n \bx_i \left(\sum_{j,k=1}^n \FC_{ijk} \by_j \by_k\right) }  \\
	    &\leq& \sum_{i=1}^n \bx_i \abs{ \sum_{j,k=1}^n \FC_{ijk} \by_j \by_k} \\
	    &\leq& \sum_{i=1}^n \bx_i \sqrt{\sum_{j,k=1}^n \crevision{(\FC_{ijk})^2}} \sqrt{\sum_{j,k=1}^n \by_j^2 \by_k^2} \\
	    &\leq&  \max\limits_{i=1,2,\ldots,n} \sqrt{\sum_{j,k=1}^n \crevision{(\FC_{ijk})^2}} \norm{\bx}_1 \norm{\by}_2^2 \\
	\end{eqnarray*}
	Therefore, a sufficient condition is to have
	\begin{equation}\label{eq_proof:suff_cond_1}
	    \alpha \, \MG_\eps(\bx,\by,\by) \geq \max\limits_{i=1,2,\ldots,n} \sqrt{\sum_{j,k=1}^n (\FC_{ijk})^2 } \norm{\bx}_1 \norm{\by}^2_2 
	\end{equation} for all $\bx\in M, \by\in \RR^n$.
	Moreover, it holds for all $\bx\in M$ and $\by\in \RR^n$ that
	\begin{eqnarray*}
	    \MG_\eps(\bx,\by,\by) 
	    &=& \sum_{i=1}^n \inner{\be_i^\eps, \bx} \inner{\be_i^\eps, \by}^2 \\
	    &=& \sum_{i=1}^n \inner{(1-\eps)\be_i + \eps \ones, \, \bx} \inner{\be_i^\eps, \by}^2 \\
	    &=& \sum_{i=1}^n \big((1-\eps)\bx_i + \eps\inner{\ones, \bx} \big) \inner{\be_i^\eps, \by}^2 \\
	    &\geq& \sum_{i=1}^n \eps \inner{\ones, \bx} \inner{\be_i^\eps, \by}^2 \\
	    &=& \sum_{i=1}^n \eps \norm{\bx}_1 \big((1-\eps)\by_i + \eps\inner{\ones, \by} \big)^2 \\
	    &\geq& \eps (1-\eps\big)^2 \norm{\bx}_1 \norm{\by}_2^2 .
	\end{eqnarray*}
	Combining this with \eqref{eq_proof:suff_cond_1} yields the following sufficient condition
	\begin{eqnarray*}
	    \alpha \,\geq\, \frac{1}{\eps (1-\eps)^2} \, \max\limits_{i=1,2,\ldots,n} \sqrt{\sum_{j,k=1}^n (\FC_{ijk})^2 } .
	\end{eqnarray*}
	In order to minimize this bound, one solves the following one-dimensional optimization problem
	$$
	    \min\limits_{0<\eps<1} \frac{1}{\eps(1-\eps)^2}
	$$
	which leads to $\eps=1/3.$
	Finally, applying the above result for $\eps=1/3$, one obtains that 
	$\alpha \geq \frac{27}{4} \max\limits_{i=1,2,\ldots,n} \sqrt{\sum_{j,k=1}^n (\FC_{ijk})^2 }$ 
	is a sufficient condition for $\nabla^2 S_\alpha(\bx)$ to be positive semidefinite at every $\bx$ in $M.$
	
	\item
	From Eq. \eqref{eq:proof_Geps}, one has the relation $\MG(\bx,\bx,\bx)=\MG_{\eps=1/3}(\bx,\bx,\bx).$
	It holds for all $\bx \in M$
	\begin{eqnarray}\label{eq:proof_Gxxx}
	    \MG_\eps(\bx,\bx,\bx) &=& \sum_{i=1}^n \inner{\be_i^\eps, \bx} \inner{\be_i^\eps, \bx} \inner{\be_i^\eps, \bx} \nonumber\\    
	    &=& \sum_{i=1}^n \big((1-\eps)\bx_i + \eps\inner{\ones, \bx} \big)^3 \nonumber\\    
	    &=& \sum_{i=1}^n \big((1-\eps)\bx_i + \eps\norm{\bx}_1 \big)^3 \nonumber\\    
	    &=& (1-\eps)^3 \norm{\bx}_3^3 + n \eps^3 \norm{\bx}_1^3 + \nonumber\\
	    && 3(1-\eps)^2 \eps \norm{\bx}_2^2 \norm{\bx}_1 + \nonumber\\
	    && 3(1-\eps) \eps^2 \norm{\bx}_1^3 . 
	\end{eqnarray}
	However, $\norm{\bx}_1=\norm{\bx}_2^2=\norm{\bx}_3^3=n_1=\textrm{const}$
	for all $\bx \in M$, thus $\MG_\eps(\bx,\bx,\bx)$ is constant on $M$ for any given $0<\eps<1.$
	This results in
	\begin{eqnarray*}
	    S_\alpha(\bx)
	    &=& \MF(\bx,\bx,\bx) + \alpha \, \MG(\bx,\bx,\bx) \\
	    &=& S(\bx) + \textrm{const} ,
	\end{eqnarray*}
	which implies the modified problem is still equivalent to the original problem, that is,
	\begin{eqnarray*}
	    \argmax_{\bx\in M} S_\alpha(\bx) = \argmax_{\bx\in M} S(\bx) .
	\end{eqnarray*}
    \end{enumerate}
\end{proof}
\begin{customTheorem}{3.5}
    Let $\FC$ be a symmetric third-order tensor 
    \crevision{and $\alpha\geq\frac{27}{4}\max\limits_{i=1,2,\ldots,n} \sqrt{\sum_{j,k=1}^n (\FC_{ijk})^2 }$ .}
    Then the following holds for Algorithm \ref{algo:bcagm3}:
    \begin{enumerate}
	\item The sequence $\MF_\alpha (\bx^{k},\by^{k},\bz^{k})$ for $k=1,2,\ldots$ 
	is strictly monotonically increasing or terminates.
	\item The sequence of scores $S(\bu^m)$ for $m=1,2,\ldots$ is strictly monotonically increasing or terminates. 
	For every $m$, $\bu^m \in M$ is a valid assignment matrix.
	\item The algorithm terminates after a finite number of iterations.
    \end{enumerate}
\end{customTheorem}
\begin{proof}
    \begin{enumerate}
	\item 
	It follows from the definition of steps $4)-6)$ in Algorithm \ref{algo:bcagm3}, 
	\begin{eqnarray}
		\MF_{\alpha,k}  &:= &\!\!\! \MF_\alpha(\bx^k,\by^{k},\bz^{k}) \nonumber \\
		&\leq & \MF_\alpha(\tbx^{k+1},\by^{k},\bz^{k}) \nonumber \\
		&\leq&  \MF_\alpha(\tbx^{k+1},\tby^{k+1},\bz^{k}) \\
		&\leq&  \MF_\alpha(\tbx^{k+1},\tby^{k+1},\tbz^{k+1}) =:\ \tilde{\MF}_{\alpha,k+1}.\nonumber
	\end{eqnarray}
	Either $\tilde{\MF}_{\alpha,k+1} > \MF_{\alpha,k}$ in which case
	$$\bx^{k+1} = \tbx^{k+1},\by^{k+1} = \tby^{k+1}, \bz^{k+1}=\tbz^{k+1}$$
	and
      \begin{equation*} 
	    \MF_\alpha(\bx^{k+1},\by^{k+1},\bz^{k+1}) > \MF_\alpha(\bx^{k},\by^{k},\bz^{k}),
      \end{equation*}
	  or $\tilde{\MF}_{\alpha,k+1}=\MF_{\alpha,k}$ and the algorithm enters step 7). 
	  Since the Hessian of $S_\alpha$ is positive semidefinite at every point on $M$ for the chosen value of $\alpha$, 
	  we get by the inequality \eqref{eq:main_inequality} in Theorem \ref{theo:main}
      \begin{eqnarray*}
	  \tilde{\MF}_{\alpha,k+1} &=& \MF_\alpha(\tbx^{k+1},\tby^{k+1},\tbz^{k+1}) \\
	  &\leq & \max\limits_{\bv\in\{\tilde\bx^{k+1},\tilde\by^{k+1},\tilde\bz^{k+1}\}}\MF_\alpha(\bv,\bv,\bv) \\
	  &=& \MF_\alpha(\bu^{m+1},\bu^{m+1},\bu^{m+1}) \\ 
	  &=& S_\alpha(\bu^{m+1}) .
      \end{eqnarray*}
	If the strict inequality does not hold then the termination condition at step 12) of the algorithm is met. 
	Otherwise, we get strict ascent
      \begin{eqnarray*}
	  \tilde{\MF}_{\alpha,k+1} &<& \MF_\alpha(\bu^{m+1},\bu^{m+1},\bu^{m+1}) \\
	  &=& S_\alpha(\bu^{m+1}) = \MF_\alpha(\bx^{k+1},\by^{k+1},\bz^{k+1}) .
      \end{eqnarray*}
      
      \item 
      From
      \begin{eqnarray*}
	  S_\alpha(\bu^{m+1}) &=& \MF_\alpha(\bu^{m+1},\bu^{m+1},\bu^{m+1}) \\ 
	  &=& \MF_\alpha(\bx^{k+1},\by^{k+1},\bz^{k+1}) ,
      \end{eqnarray*}
      it follows that $S_\alpha(\bu^{m})$ for $m=1,2,\ldots$ is a subsequence of 
      $\MF_\alpha (\bx^{k},\by^{k},\bz^{k})$ for $k=1,2,\ldots$ 
      and thus it holds either $S_\alpha(\bu^{m}) = S_\alpha(\bu^{m+1})$ in which case the algorithm terminates 
      or $S_\alpha(\bu^{m}) < S_\alpha(\bu^{m+1})$.
      However, by Proposition \ref{prop:convexification}, the additional term which has been added to $S$ 
      is constant on $M$, that is $S_\alpha(\bx) = S(\bx) + \textrm{const}$ for all $\bx\in M.$
      It follows that either $S(\bu^{m}) = S(\bu^{m+1})$ and the algorithm terminates or 
      $S(\bu^{m}) < S(\bu^{m+1}).$  
      
      \item
      The algorithm yields a strictly monotonically increasing sequence 
      $S(\bu^m),m=1,2,\ldots$ or it terminates. 
      Since there is only a finite number of possible assignment matrices, 
      the algorithm has to terminate after a finite number of iterations.
    \end{enumerate}
\end{proof}
\begin{customTheorem}{3.7}
    Let $\FC$ be a symmetric \cblue{third-order} tensor
    \crevision{and $\alpha\geq\frac{27}{4}\max\limits_{i=1,2,\ldots,n} \sqrt{\sum_{j,k=1}^n (\FC_{ijk})^2 }$ .}
    Let $\Psi$ be an algorithm for the QAP which yields monotonic ascent w.r.t. the current iterate.
    Then the following holds for Algorithm \ref{algo:bcagm3_psi}:
    \begin{enumerate}
	  \itemsep0em
	  \item The sequence $\MF_\alpha (\bx^{k},\by^{k},\by^{k})$ for $k=1,2,\ldots$ 
	  is strictly monotonically increasing or terminates.
	  \item The sequence of scores $S(\bu^m)$ for $m=1,2,\ldots$ is strictly monotonically increasing or terminates. 
		For every $m$, $\bu^m \in M$ is a valid assignment matrix.
    	  \item The algorithm terminates after a finite number of iterations.
        \end{enumerate}
    \label{theo:quadratic}
\end{customTheorem}
\begin{proof}
      The proof is similar to that of \crevision{Theorem \ref{theo:bcagm3}.}
\end{proof}
\begin{customProposition}{3.8}
    Let $\MF: \RR^n\times \RR^n \times \RR^n \to \RR$ be a symmetric third-order multilinear form
    and $\MG: \RR^n\times \RR^n \times \RR^n \to \RR$ as defined in Eq. \eqref{eq:mod_mf}.
    For each $\alpha \in \RR$, let $\MF_\alpha$ be defined as
    \begin{equation*}
	\MF_\alpha(\bx,\by,\bz) = \MF(\bx,\by,\bz) + \alpha\,\MG(\bx,\by,\bz) . 
    \end{equation*}
    Then the following holds 
    \begin{enumerate}
	\item $\MF_\alpha$ is a symmetric multilinear form.
	\item 
	For all non-homogeneous tuple $(\bx,\by,\bz) \in M\, (\ie \bx\neq\by \textrm{ or } \by\neq\bz \textrm{ or } \bz\neq\bx),$ 
	the following inequality holds 
	\begin{equation*}
	    \MF_\alpha(\bx,\by,\bz) \leq \max\limits_{\bu\in\set{\bx,\by,\bz}} \MF_\alpha(\bu,\bu,\bu)
	\end{equation*}
	if and only if
	$$
	    \alpha \geq \Lambda(\bx,\by,\bz) \bydef \frac{ \MF(\bx,\by,\bz) - \max\limits_{\bu\in\set{\bx,\by,\bz}} \MF(\bu,\bu,\bu) } 
	    { \MG(\bx,\bx,\bx) - \MG(\bx,\by,\bz) } .
	$$
	\item For every $\alpha \in \RR$, it holds
	$$
	    \argmax\limits_{\bx\in M} S_\alpha(\bx) = \argmax\limits_{\bx\in M} S(\bx) .
	$$
    \end{enumerate}
\end{customProposition}
\begin{proof}
    \begin{enumerate}
	\item 
	Since $\MG$ is a symmetric multilinear form, $\MF_\alpha$ must be symmetric as well.
	
	\item 
	It holds for all $\bx,\by,\bz\in M$
	\begin{eqnarray*}
	    && \MF_\alpha(\bx,\by,\bz) \leq \max\limits_{\bu\in\set{\bx,\by,\bz}} \MF_\alpha(\bu,\bu,\bu) \\
	    &\Leftrightarrow& 
	    \MF_\alpha(\bx,\by,\bz) \leq \max\limits_{\bu\in\set{\bx,\by,\bz}} \left\lbrace \MF(\bu,\bu,\bu) 
	    + \alpha \, \MG(\bu,\bu,\bu) \right\rbrace \\
	    &\Leftrightarrow& 
	    \MF_\alpha(\bx,\by,\bz) \leq \max\limits_{\bu\in\set{\bx,\by,\bz}} \left\lbrace \MF(\bu,\bu,\bu) \right\rbrace
	    + \alpha \, \MG(\bx,\bx,\bx) .
	\end{eqnarray*}
	Note that we have used in the last step the fact that $\MG(\bu,\bu,\bu)=\MG(\bx,\bx,\bx)=\MG(\by,\by,\by)=\MG(\bz,\bz,\bz)$
	for all $\bx,\by,\bz\in M$ (see Eq. \eqref{eq:proof_Gxxx}).
	By substituting $\MF_\alpha(\bx,\by,\bz) = \MF(\bx,\by,\bz) + \alpha \, \MG(\bx,\by,\bz)$, the above inequality chain 
	is equivalent to
	\begin{eqnarray*}
	     \alpha (\MG(\bx,\bx,\bx) - \MG(\bx,\by,\bz)) \geq \MF(\bx,\by,\bz) - 
	     \!\!\!\!\! \max\limits_{\bu\in\set{\bx,\by,\bz}} \!\!\!\!\!\! \MF(\bu,\bu,\bu) .
	\end{eqnarray*}
	Notice that $\MG(\bx,\bx,\bx) > \MG(\bx,\by,\bz)$ (see proof below) for every non-homogeneous tuple $\set{\bx,\by,\bz}\in M,$
	thus, one can divide both sides by $(\MG(\bx,\bx,\bx) - \MG(\bx,\by,\bz))$ to get
	\begin{eqnarray*}
	     \alpha \geq \frac{ \MF(\bx,\by,\bz) - \max\limits_{\bu\in\set{\bx,\by,\bz}} \MF(\bu,\bu,\bu) }
	     { \MG(\bx,\bx,\bx) - \MG(\bx,\by,\bz) } .
	\end{eqnarray*} 
	**) Assume $\set{\bx,\by,\bz}\in M$ is a non-homogeneous tuple,
	we will prove for any $0<\eps<1$ that 
	\begin{equation}\label{eq_proof:nonhomo_ineq}
	    \MG_\eps(\bx,\bx,\bx) > \MG_\eps(\bx,\by,\bz) .
	\end{equation}    
	From Eq. \eqref{eq:proof_Geps}, one has
	\begin{eqnarray*}
	    &&\MG_\eps(\bx,\by,\bz) \\
	    &\bydef& \sum_{i=1}^n \inner{\be_i^\eps, \bx} \inner{\be_i^\eps, \by} \inner{\be_i^\eps, \bz} \\    
	    &=& \sum_{i=1}^n \big((1-\eps)\bx_i + \eps\inner{\ones, \bx} \big) 
	    \cdot \big((1-\eps)\by_i + \eps\inner{\ones, \by} \big) \cdot \\
	    && \big((1-\eps)\bz_i + \eps\inner{\ones, \bz} \big) \\
	    &=& \sum_{i=1}^n \big((1-\eps)\bx_i + \eps n_1 \big) 
	    \cdot \big((1-\eps)\by_i + \eps n_1 \big) \cdot \\
	    && \big((1-\eps)\bz_i + \eps n_1 \big) \\
	    &=& 3 (1-\eps) \eps^2 n_1^3 + (1-\eps)^3 \sum_{i=1}^n \bx_i\by_i\bz_i + \\
	    && (1-\eps)^2\eps n_1 \sum_{i=1}^n  (\bx_i\by_i + \by_i\bz_i + \bx_i\bz_i) + n \eps^3 n_1^3 .
	\end{eqnarray*}
	Since $\bx,\by,\bz\in M$, one has $\sum_{i=1}^n \bx_i\by_i\bz_i \leq \sum_{i=1}^n \bx_i\bx_i\bx_i,$
	$\sum_{i=1}^n \bx_i\by_i \leq \sum_{i=1}^n \bx_i\bx_i,$
	$\sum_{i=1}^n \by_i\bz_i \leq \sum_{i=1}^n \by_i\by_i = n_1 = \sum_{i=1}^n \bx_i\bx_i,$ and
	$\sum_{i=1}^n \bx_i\bz_i \leq \sum_{i=1}^n \bz_i\bz_i = n_1 = \sum_{i=1}^n \bx_i\bx_i.$
	Therefore, $\MG_\eps(\bx,\by,\bz) \leq \MG_\eps(\bx,\bx,\bx).$
	The inequality becomes equality if and only if $\bx=\by=\bz.$
	Applying this result for $\eps=1/3$ finishes the proof.
	
	\item 
	It follows from Eq. \eqref{eq:proof_Gxxx} that $\MG(\bx,\bx,\bx)$ is constant for all $\bx\in M.$
	Thus, it holds for every $\alpha\in \RR$ and $\bx\in M$ that
	\begin{eqnarray*}
	      S_\alpha(\bx) &\bydef& \MF_\alpha(\bx,\bx,\bx) \\
	      &=& \MF(\bx,\bx,\bx) + \alpha \, \MG(\bx,\bx,\bx) \\
	      &=& S(\bx) + \textrm{const} ,
	\end{eqnarray*}
	which implies
	\begin{eqnarray*}
	      \argmax\limits_{\bx\in M} S_\alpha(\bx) = \argmax\limits_{\bx\in M} S(\bx) .
	\end{eqnarray*}
    \end{enumerate}
\end{proof}
\begin{customLemma}{3.6}
    Suppose Algorithm \ref{algo:bcagm3} runs with $\alpha=\alpha_0$ for some $\alpha_0\in\RR_+$.
	If $\tbx^{k+1}=\tby^{k+1}=\tbz^{k+1}=:\bu$ holds at some iteration $k$ then 
        for all $\alpha\geq\alpha_0$, it holds
	$$
	    \bu = \argmax\limits_{\bz\in M} \MF_\alpha(\bz,\bu,\bu) .
	$$ 
\end{customLemma}
\begin{proof}
	Recall from Algorithm \ref{algo:bcagm3} the multilinear form $\MG$ defined as
	$\MG(\bx,\by,\bz) = \sum_{i=1}^n \inner{\bar{\be}_i, \bx} \inner{\bar{\be}_i, \by} \inner{\bar{\be}_i, \bz},$ 
	and one has the relation
	$$
	    \MF_{\alpha}(\bx,\by,\bz) = \MF(\bx,\by,\bz) + \alpha \, \MG(\bx,\by,\bz) .
	$$
	By substituting $\tbx^{k+1}=\tby^{k+1}=\tbz^{k+1}=\bu$ and $\alpha=\alpha_0$
	into step (6) of Algorithm \ref{algo:bcagm3}, we have
	\begin{eqnarray} \label{eq_proof:claim1} 
	    \bu &=& \argmax_{\bz\in M} \MF_{\alpha_0}(\bu,\bu,\bz) .
	\end{eqnarray}
	Besides, one can show that 
	\begin{eqnarray} \label{eq_proof:claim2} 
	    \bu=\argmax_{\bz\in M}\MG(\bu,\bu,\bz) .
	\end{eqnarray}
	\cblue{
	Indeed, it follows from Eq. \eqref{eq_proof:nonhomo_ineq} that
	$\MG(\bu,\bu,\bu) > \MG(\bu,\bu,\bz)$ for all $\bz\in M$ and $\bz\neq\bu,$ 
	which implies the maximum must be attained at $\bz=\bu.$
	}
	Combining \eqref{eq_proof:claim1} and \eqref{eq_proof:claim2}, one obtains for all $\alpha\geq\alpha_0$
	\begin{eqnarray*}
	    \bu &=& \argmax_{\bz\in M} \left\lbrace \MF_{\alpha_0}(\bu,\bu,\bz) + (\alpha-\alpha_0) \, \MG(\bu,\bu,\bz) \right\rbrace \\
	     &=& \argmax_{\bz\in M} \left\lbrace \MF(\bu,\bu,\bz) + \alpha \, \MG(\bu,\bu,\bz) \right\rbrace \\
	     &=& \argmax_{\bz\in M} \MF_{\alpha}(\bu,\bu,\bz) .
	\end{eqnarray*}	
\end{proof}

\section{}

\subsection{Adaptive methods vs Non-adaptive methods}
Table \ref{tab:adapt3_vs_nonadapt3}
compares non-adaptive methods from Algorithm \ref{algo:bcagm3} and Algorithm \ref{algo:bcagm3_psi}
with adaptive methods from Algorithm \ref{algo:adapt_bcagm3} and Algorithm \ref{algo:adapt_bcagm3_psi}.
Table \ref{tab:adapt4_vs_nonadapt4} shows the same comparison for fourth-order algorithms.

\begin{table*}
    \centering
    \renewcommand{\arraystretch}{1.3}
    \begin{tabular}{|c|c|c|c|c|c|c|c|c|}
	\hline
	& \multicolumn{2}{c|}{Synthetic} & \multicolumn{2}{c|}{House} & \multicolumn{2}{c|}{Face+Duck+Winebottle} \\
	\hline \hline
	\textbf{Comparison} & No. & Avg(\%) 		& No. & Avg(\%) 	& No. & Avg(\%) \\ \hline 
	\hline
	\rowcolor{Gray}
	Adapt-BCAGM3 $>$ BCAGM3 & 			185&321.9	&  		7&10.4 		& 		4141&16.0  \\ \hline
	Adapt-BCAGM3 $<$ BCAGM3 & 			0&0.0 		&  		0&0.0 		& 		6&3.0	\\ \hline
	Adapt-BCAGM3 $=$ BCAGM3 & 			4915&0.0	&  		1673&0.0 	& 		58853&0.0  \\ \hline
	\hline
	\rowcolor{Gray}
	Adapt-BCAGM3+IPFP $>$ BCAGM3+IPFP & 		167&20.1	&  		3&9.9 		& 		7566&32.0  \\ \hline
	Adapt-BCAGM3+IPFP $<$ BCAGM3+IPFP & 		0&0.0 		&  		0&0.0 		& 		0&0.0  \\ \hline
	Adapt-BCAGM3+IPFP $=$ BCAGM3+IPFP & 		4933&0.0	&  		1677&0.0	& 		55434&0.0  \\ \hline
	\hline
	\rowcolor{Gray}
	Adapt-BCAGM3+MP $>$ BCAGM3+MP & 		168&146.7	&  		2&61.9 		& 		4234&51.0  \\ \hline
	Adapt-BCAGM3+MP $<$ BCAGM3+MP & 		0&0.0 		&  		0&0.0 		& 		0&0.0  \\ \hline
	Adapt-BCAGM3+MP $=$ BCAGM3+MP & 		4932&0.0	&  		1678&0.0	& 		58766&0.0  \\ \hline
    \end{tabular}
     \caption{Adaptive third-order methods (Adapt-BCAGM3, Adapt-BCAGM3+$\Psi$) 
    vs. non-adaptive third-order methods (BCAGM3, BCAGM3+$\Psi$).
    The first column (No.) shows the number of test cases where one algorithm achieves better matching score than the other.
    The second column (Avg(\%)) shows the average performance gain achieved by the better method in each case,
    which is computed using Eq. \eqref{eq:avg_gain}.}
    \label{tab:adapt3_vs_nonadapt3}
\end{table*}

\begin{table*}
    \centering
    \renewcommand{\arraystretch}{1.3}
    \begin{tabular}{|c|c|c|c|c|c|c|c|c|}
	\hline
	& \multicolumn{2}{c|}{Synthetic} & \multicolumn{2}{c|}{House} & \multicolumn{2}{c|}{Face+Duck+Winebottle} \\
	\hline \hline
	\textbf{Comparison} & No. & Avg(\%) 		& No. & Avg(\%) 	& No. & Avg(\%) \\ \hline 
	\hline
	\rowcolor{Gray}
	Adapt-BCAGM $>$ BCAGM & 			52&4.7		&  		2&0.8 		& 		1448&8.0  \\ \hline
	Adapt-BCAGM $<$ BCAGM & 			0&0.0 		&  		0&0.0 		& 		4&13.0  \\ \hline
	Adapt-BCAGM $=$ BCAGM & 			5048&0.0	&  		1678&0.0 	& 		61548&0.0  \\ \hline
	\hline
	\rowcolor{Gray}
	Adapt-BCAGM+IPFP $>$ BCAGM+IPFP & 		53&8.1		&  		1&5.1 		& 		2130&14.0  \\ \hline
	Adapt-BCAGM+IPFP $<$ BCAGM+IPFP & 		0&0.0 		&  		0&0.0 		& 		0&0.0  \\ \hline
	Adapt-BCAGM+IPFP $=$ BCAGM+IPFP & 		5047&0.0	&  		1679&0.0	& 		60870&0.0  \\ \hline
	\hline
	\rowcolor{Gray}
	Adapt-BCAGM+MP $>$ BCAGM+MP & 			50&4.4		&  		0&0.0 		& 		876&8.0  \\ \hline
	Adapt-BCAGM+MP $<$ BCAGM+MP & 			0&0.0 		&  		0&0.0 		& 		0&0.0  \\ \hline
	Adapt-BCAGM+MP $=$ BCAGM+MP & 			5050&0.0	&  		1680&0.0 	& 		62124&0.0  \\ \hline
    \end{tabular}
    \caption{Adaptive fourth-order methods (Adapt-BCAGM, Adapt-BCAGM+$\Psi$) 
    vs. non-adaptive fourth-order methods (BCAGM, BCAGM+$\Psi$ \cite{Quynh2015}).}
    \label{tab:adapt4_vs_nonadapt4}
\end{table*}

\subsection{Adaptive third-order methods vs. Standard fourth-order methods}
Table \ref{tab:adapt3_vs_fourth} shows the comparison between the proposed adaptive third-order methods with 
the state-of-the-art fourth-order approaches \cite{Quynh2015}.
\begin{table*}
    \centering
    \renewcommand{\arraystretch}{1.3}
    \begin{tabular}{|c|c|c|c|c|c|c|c|c|}
	\hline
	& \multicolumn{2}{c|}{Synthetic} & \multicolumn{2}{c|}{House} & \multicolumn{2}{c|}{Face+Duck+Winebottle} \\
	\hline \hline
	\textbf{Comparison} 	& No. & Avg(\%) 	& No. & Avg(\%) 	& No. & Avg(\%) \\ \hline 
	\hline
	\rowcolor{Gray}
	Adapt-BCAGM3 $>$ BCAGM & 		295&44.2	& 	5&1.3 		& 		7467&26.0  \\ \hline
	Adapt-BCAGM3 $<$ BCAGM & 		233&52.8  	& 	3&20.3 		& 		5394&24.0  \\ \hline
	Adapt-BCAGM3 $=$ BCAGM & 		4572&0.0 	& 	1672&0.0 	& 		50139&0.0  \\ \hline
	\hline
	\rowcolor{Gray}
	Adapt-BCAGM3+IPFP $>$ BCAGM+IPFP & 	239&3.7 	&   	20&7.1 		& 		11699&21.0  \\ \hline
	Adapt-BCAGM3+IPFP $<$ BCAGM+IPFP & 	99&61.1  	&   	3&1.9 		& 		5854&28.0  \\ \hline
	Adapt-BCAGM3+IPFP $=$ BCAGM+IPFP & 	4762&0.0 	&   	1657&0.0 	& 		45447&0.0  \\ \hline
	\hline
	\rowcolor{Gray}
	Adapt-BCAGM3+MP $>$ BCAGM+MP & 	246&35.2 	&   	4&12.2 		& 		6722&28.0  \\ \hline
	Adapt-BCAGM3+MP $<$ BCAGM+MP & 	169&61.4  	&   	4&10.4 		& 		5491&26.0  \\ \hline
	Adapt-BCAGM3+MP $=$ BCAGM+MP & 	4685&0.0 	&   	1672&0.0 	& 		50787&0.0  \\ \hline
    \end{tabular}
    \caption{Adaptive third-order methods (Adapt-BCAGM3, Adapt-BCAGM3+$\Psi$) 
    vs. non-adaptive fourth-order methods (BCAGM, BCAGM+$\Psi$ \cite{Quynh2015}).}
    \label{tab:adapt3_vs_fourth}
\end{table*}

\subsection{Running time}
Tab. \ref{tab:runtime_synthetic}, Tab. \ref{tab:runtime_house} and Tab. \ref{tab:runtime_willow}
show the running time of all higher-order algorithms in seconds on the previous experiments.
\begin{table*}
    \centering
    \renewcommand{\arraystretch}{1.3}
    \begin{tabular}{|c|c|c|c|c|c|c|c|c|c|c|c|}
	\hline
	& $n_{out} = 0$ & $40$ & $80$ & $120$ & $160$ & $200$ \\ \hline 
	\hline
	\rowcolor{Gray}
	TM& 			0.0582&    0.4969&    0.9056&    1.3159&    1.7609&    2.2812 \\ \hline
	RRWHM& 			0.0066&    0.0537&    0.1345&    0.2255&    0.3912&    0.4648 \\ \hline
	\rowcolor{Gray}
	BCAGM& 			0.0039&    0.0552&    0.1465&    0.2674&    0.4280&    0.6756 \\ \hline
	BCAGM+MP& 		0.0057&    0.0747&    0.2009&    0.4515&    0.8428&    1.5472 \\ \hline
	\rowcolor{Gray}
	BCAGM3& 		0.0020&    0.0247&    0.0636&    0.1178&    0.2000&    0.3327 \\ \hline
	BCAGM3+MP&		0.0025&    0.0344&    0.0972&    0.2238&    0.4309&    0.8144 \\ \hline
	\rowcolor{Gray}
	Adapt-BCAGM3&		0.0019&    0.0247&    0.0636&    0.1182&    0.2019&    0.3421 \\ \hline
	Adapt-BCAGM3+MP&	0.0025&    0.0346&    0.0981&    0.2256&    0.4412&    0.8355 \\ \hline
    \end{tabular}
    \caption{Average running time (in seconds) of higher-order algorithms on the synthetic outlier settings.
    (see Fig. \ref{fig:exp_out} for their matching score and matching accuracy.)
    }
    \label{tab:runtime_synthetic}
\end{table*}
\begin{table*}
    \centering
    \renewcommand{\arraystretch}{1.3}
    \begin{tabular}{|c|c|c|c|c|c|c|c|c|c|c|c|}
	\hline
	& $baseline=10$ & $20$ & $30$ & $40$ & $50$ & $60$ & $70$ & $80$ & $90$ & $100$ \\ \hline 
	\hline
	\rowcolor{Gray}
	TM& 			0.0401&    0.0404&    0.0406&    0.0406&    0.0415&    0.0419&    0.0421&    0.0412&    0.0405&    0.0412 \\ \hline
	RRWHM& 			0.0021&    0.0022&    0.0022&    0.0023&    0.0033&    0.0054&    0.0085&    0.0091&    0.0107&    0.0083 \\ \hline
	\rowcolor{Gray}
	BCAGM& 			0.0024&    0.0024&    0.0024&    0.0025&    0.0026&    0.0027&    0.0026&    0.0025&    0.0024&    0.0024 \\ \hline
	BCAGM+MP& 		0.0038&    0.0039&    0.0038&    0.0039&    0.0043&    0.0044&    0.0044&    0.0043&    0.0042&    0.0042 \\ \hline
	\rowcolor{Gray}
	BCAGM3& 		0.0013&    0.0013&    0.0013&    0.0013&    0.0014&    0.0013&    0.0013&    0.0013&    0.0012&    0.0013 \\ \hline
	BCAGM3+MP& 		0.0017&    0.0017&    0.0017&    0.0017&    0.0018&    0.0019&    0.0020&    0.0019&    0.0018&    0.0019 \\ \hline
	\rowcolor{Gray}
	Adapt-BCAGM3& 		0.0013&    0.0013&    0.0012&    0.0013&    0.0013&    0.0013&    0.0013&    0.0013&    0.0012&    0.0012 \\ \hline
	Adapt-BCAGM3+MP& 	0.0016&    0.0017&    0.0017&    0.0017&    0.0019&    0.0019&    0.0020&    0.0019&    0.0018&    0.0019 \\ \hline
    \end{tabular}
    \caption{Average running time (in seconds) of higher-order algorithms on the CMU House dataset.
    (see Fig. \ref{fig:exp_house} for their matching score and matching accuracy.)
    }
    \label{tab:runtime_house}
\end{table*}
\begin{table*}
    \centering
    \renewcommand{\arraystretch}{1.3}
    \begin{tabular}{|c|c|c|c|c|c|c|c|c|c|c|c|}
	\hline
	& $n_{out} = 0$ & $5$ & $10$ & $15$ \\ \hline 
	\hline
	\rowcolor{Gray}
	TM& 			0.0914&    0.1885&    0.3219&    0.3567 \\ \hline
	RRWHM& 			0.0220&    0.0499&    0.2326&    0.1682 \\ \hline
	\rowcolor{Gray}
	BCAGM& 			0.0062&    0.0123&    0.0212&    0.0285 \\ \hline
	BCAGM+MP& 		0.0098&    0.0247&    0.0490&    0.0654 \\ \hline
	\rowcolor{Gray}
	BCAGM3& 		0.0032&    0.0061&    0.0103&    0.0142 \\ \hline
	BCAGM3+MP& 		0.0047&    0.0117&    0.0232&    0.0343 \\ \hline
	\rowcolor{Gray}
	Adapt-BCAGM3& 		0.0031&    0.0059&    0.0101&    0.0140 \\ \hline
	Adapt-BCAGM3+MP& 	0.0046&    0.0119&    0.0246&    0.0370 \\ \hline
    \end{tabular}
    \caption{Average running time (in seconds) of higher-order algorithms on Willow dataset.
    (see Fig. \ref{fig:exp_willow} for their matching score and matching accuracy.)
    }
    \label{tab:runtime_willow}
\end{table*}

\subsection{Demo of matching results}
\begin{figure*}
    \subfloat[Input: 10 pts vs 30 pts, baseline = 50]{ \includegraphics[width=0.31\linewidth]{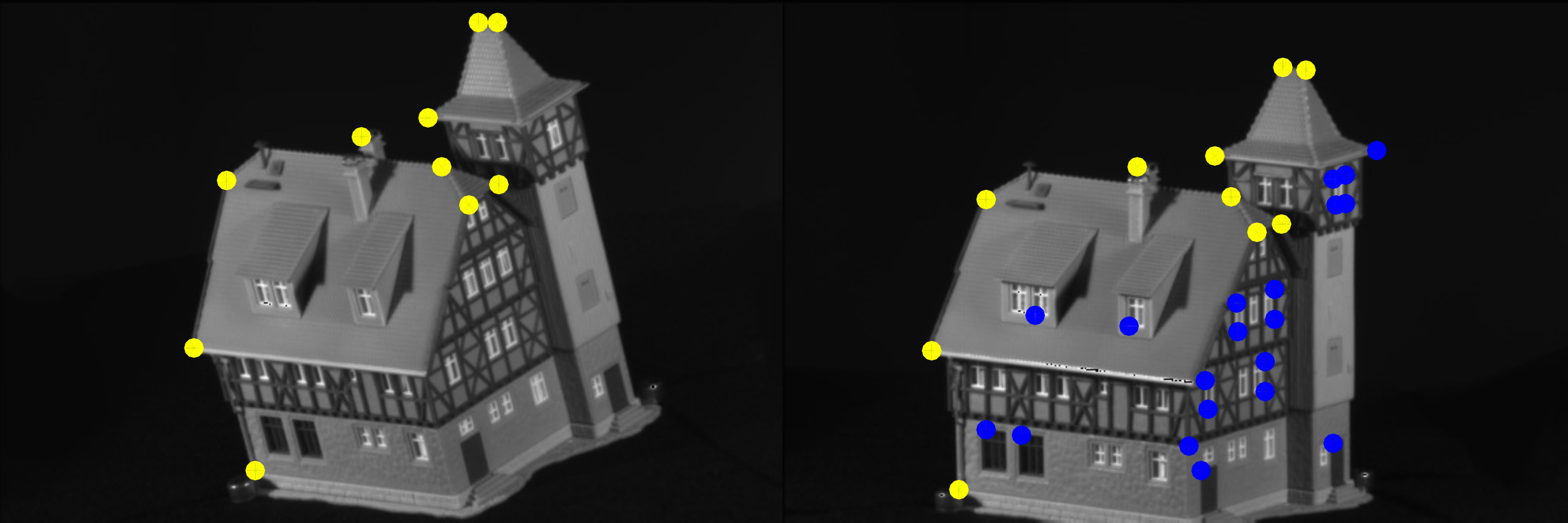} }  
    \subfloat[IPFP\cite{LeoHebSuk2009} (44.8)]{ \includegraphics[width=0.31\linewidth]{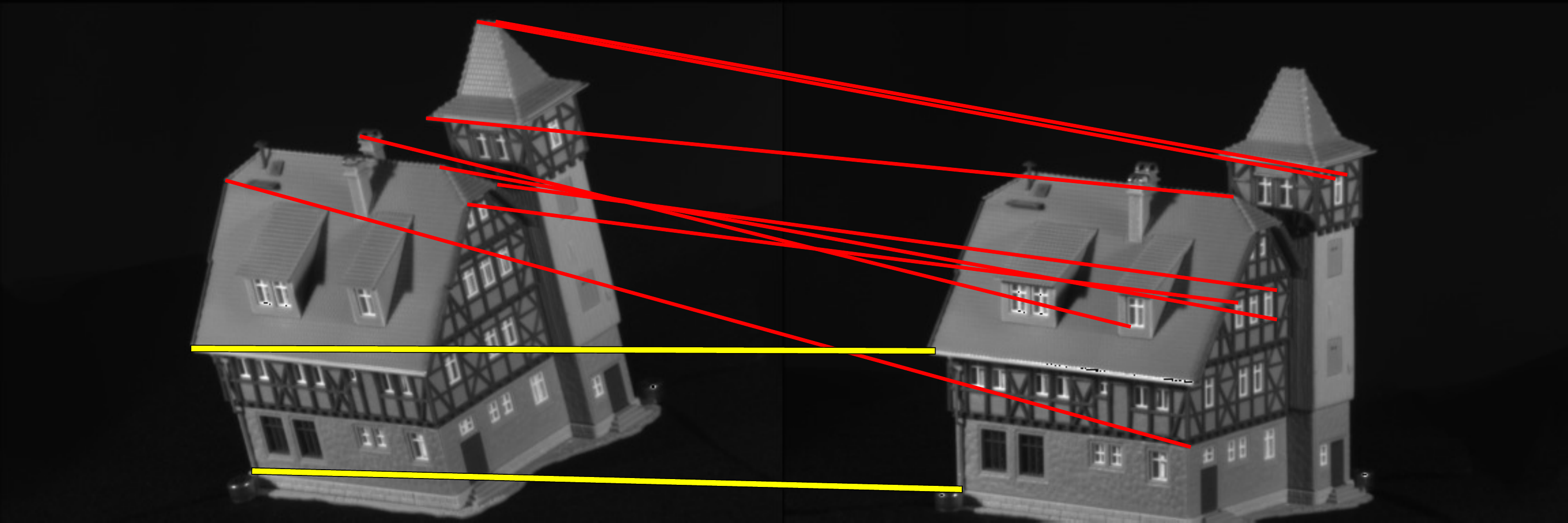} }  
    \subfloat[MPM\cite{ChoEtAl2014} (112.2)]{ \includegraphics[width=0.31\linewidth]{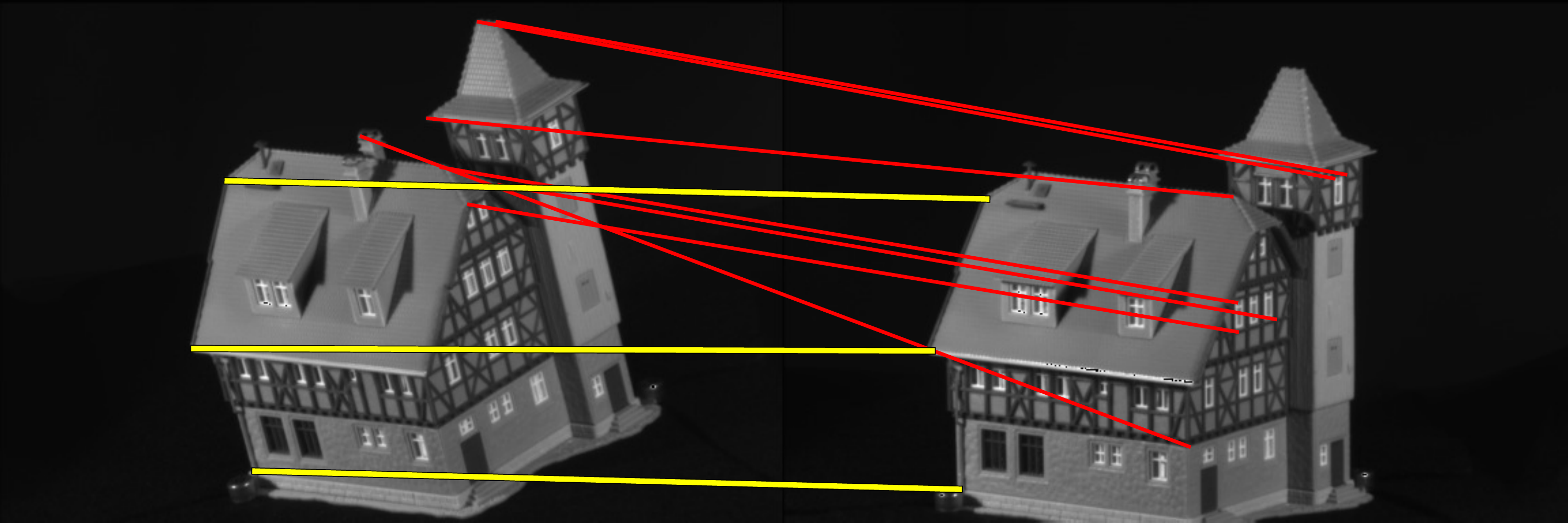} } \\  
    \subfloat[TM\cite{DucEtAl2011} (217.6)]{ \includegraphics[width=0.31\linewidth]{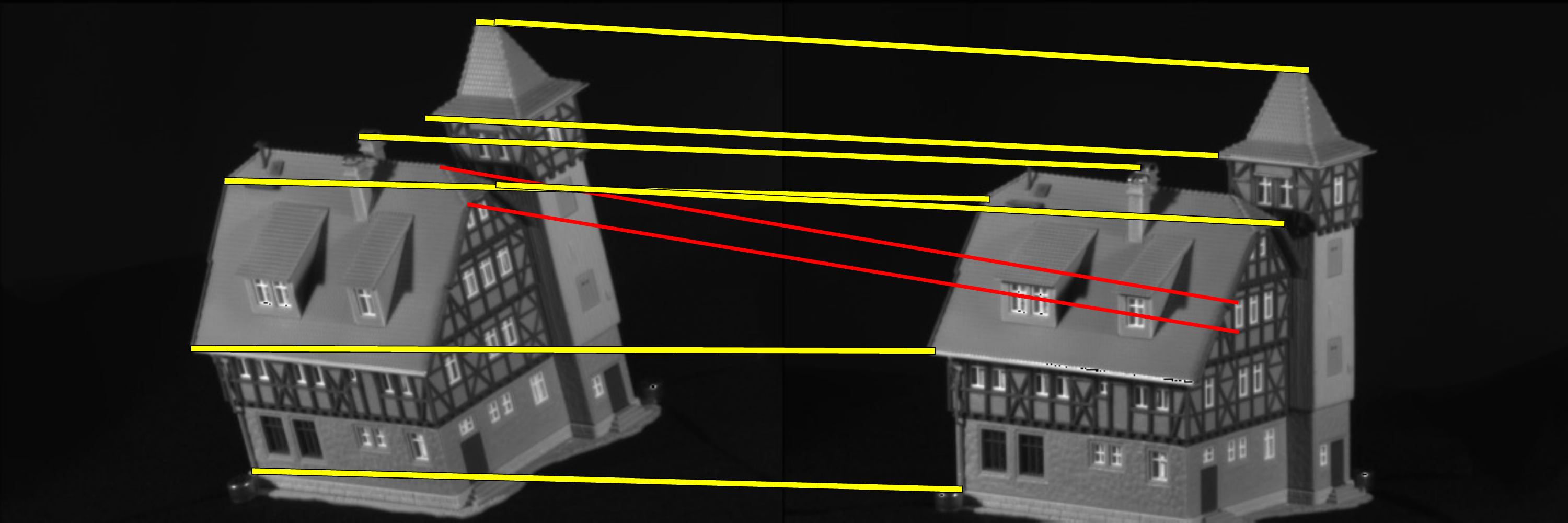} }  
    \subfloat[RRWHM\cite{LeeChoLee2011} (285.9)]{ \includegraphics[width=0.31\linewidth]{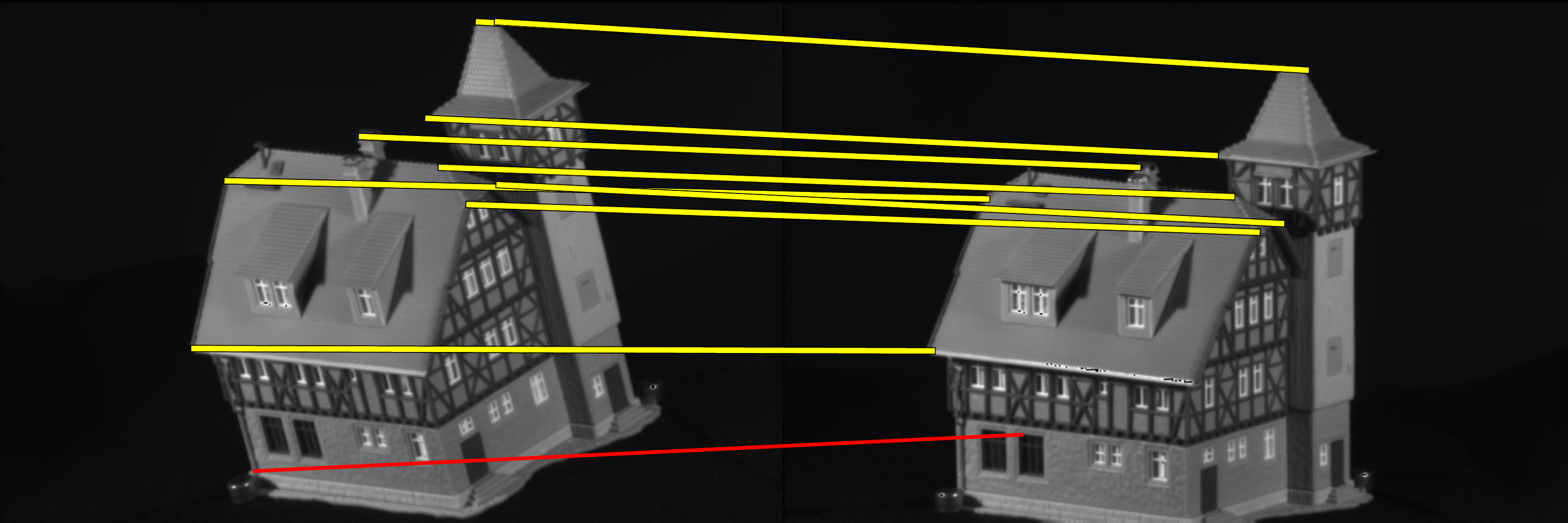} }  
    \subfloat[BCAGM\cite{Quynh2015} (329.6)]{ \includegraphics[width=0.31\linewidth]{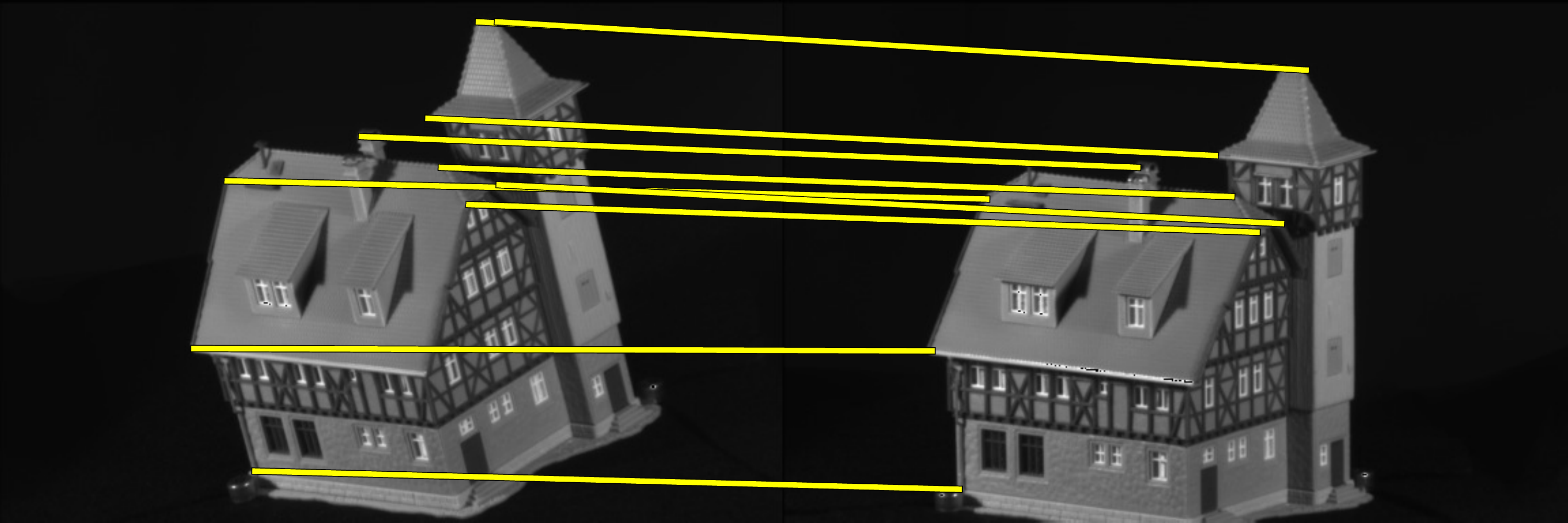} }  \\
    \subfloat[BCAGM+MP\cite{Quynh2015} (329.6)]{ \includegraphics[width=0.31\linewidth]{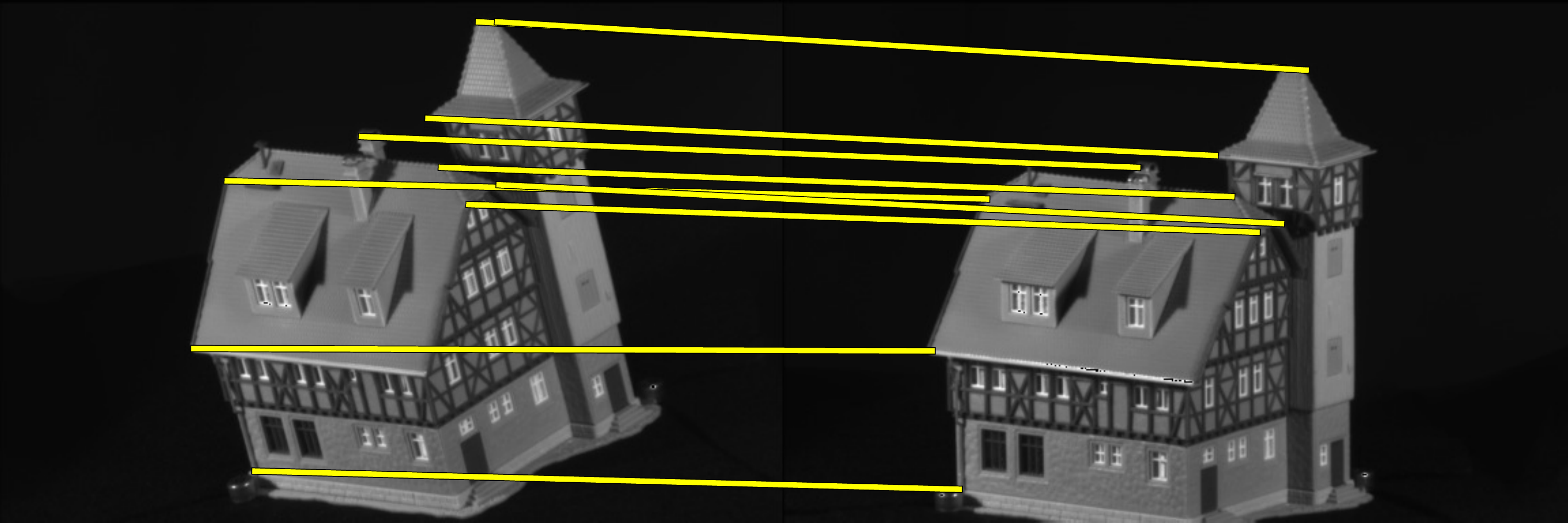} } 
    \subfloat[Adapt-BCAGM3 (329.6)]{ \includegraphics[width=0.31\linewidth]{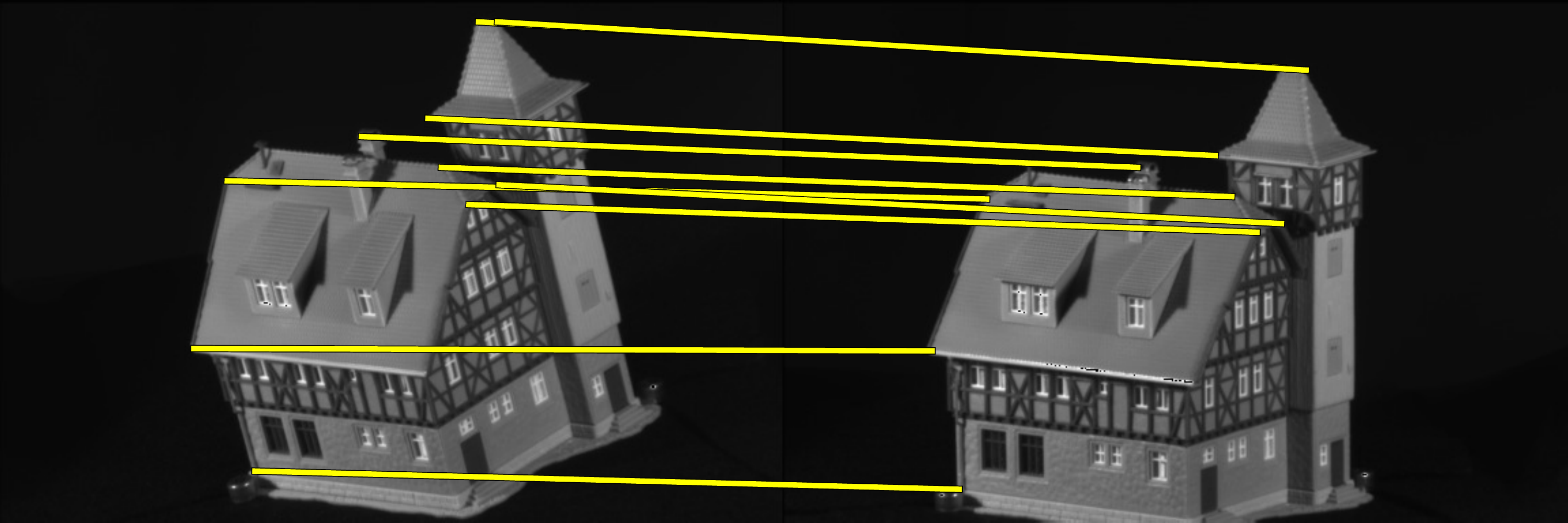} }  
    \subfloat[Adapt-BCAGM3+MP (329.6)]{ \includegraphics[width=0.31\linewidth]{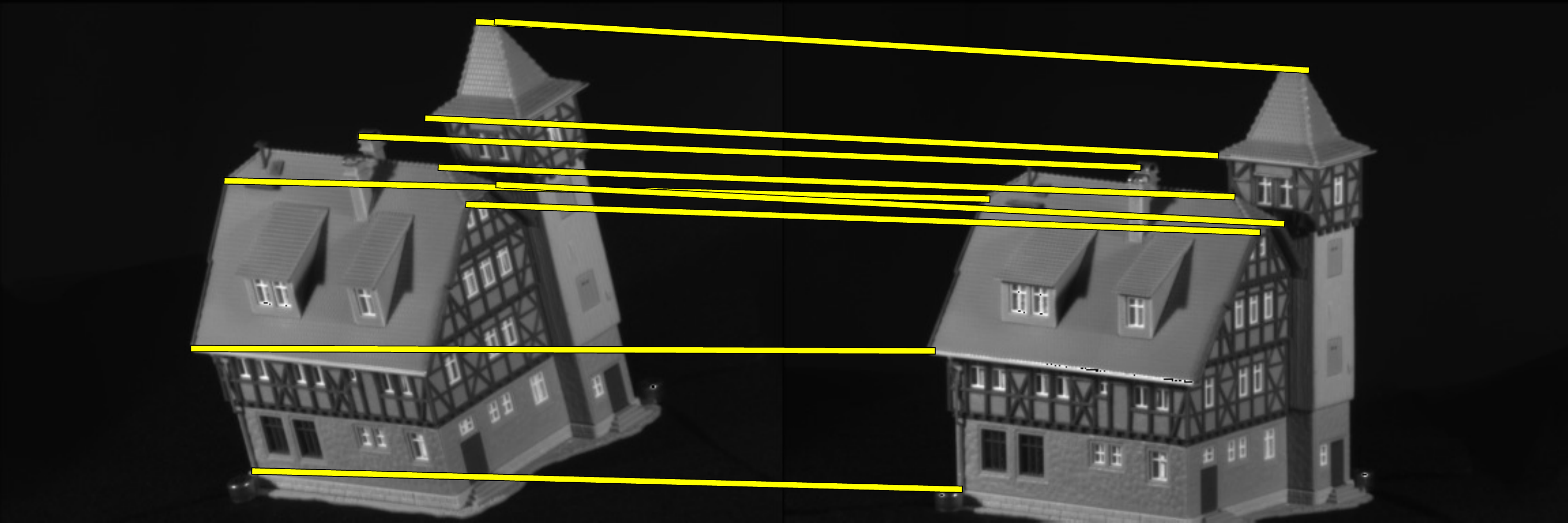} }
\caption{Demo of matching results on CMU House dataset with \textit{small baseline} ($\textrm{baseline}=50$).
a) Input images. Yellow dots denote inlier points, blue dots denote outlier points.
b) c) Matching results of previous second-order methods. 
d) e) Matching results of previous higher-order methods. 
f) g) h) i) Matching results of our higher-order approaches.
The yellow/red lines indicate correct/incorrect matches. 
Matching score is reported for each method. (Best viewed in color.)}
\label{fig1:house_matches}
\end{figure*}

\begin{figure*}
    \subfloat[Input: 10 pts vs 30 pts, baseline = 80]{ \includegraphics[width=0.31\linewidth]{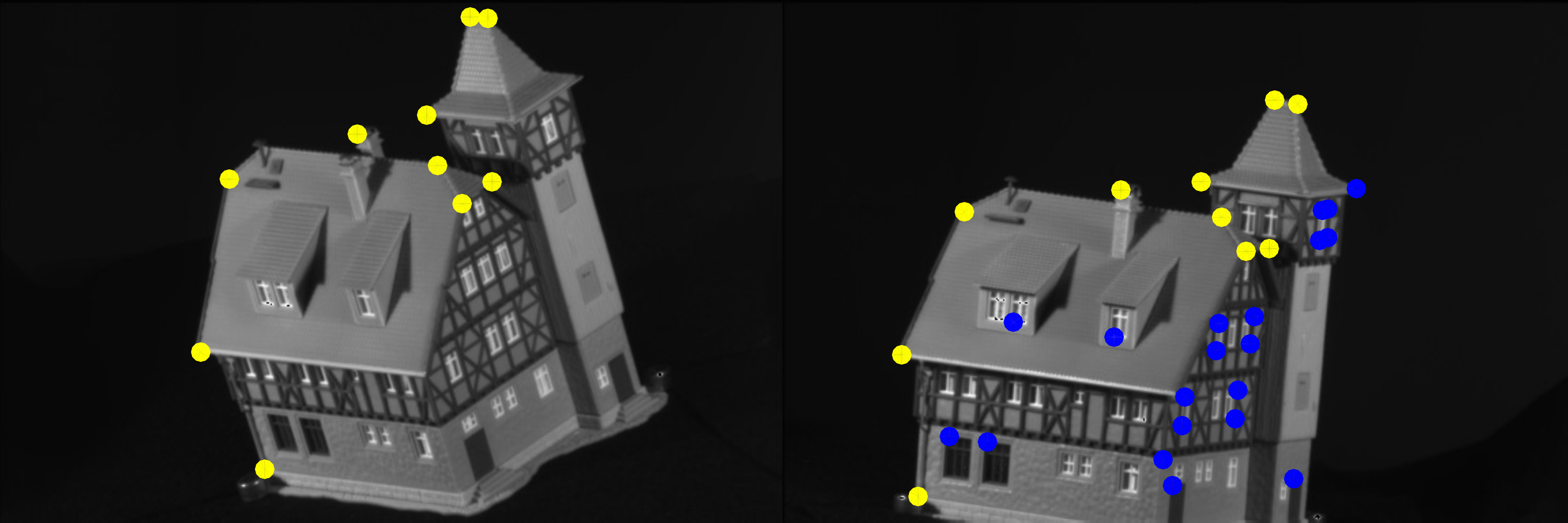} }  
    \subfloat[IPFP\cite{LeoHebSuk2009} (86.6)]{ \includegraphics[width=0.31\linewidth]{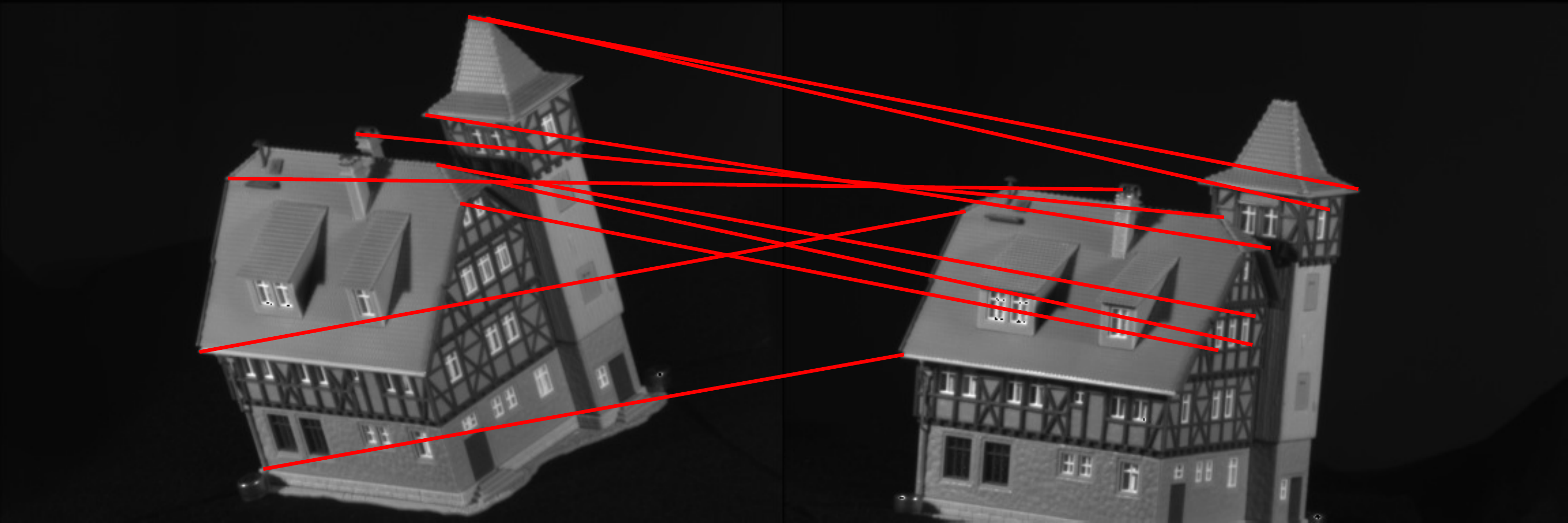} }  
    \subfloat[MPM\cite{ChoEtAl2014} (34)]{ \includegraphics[width=0.31\linewidth]{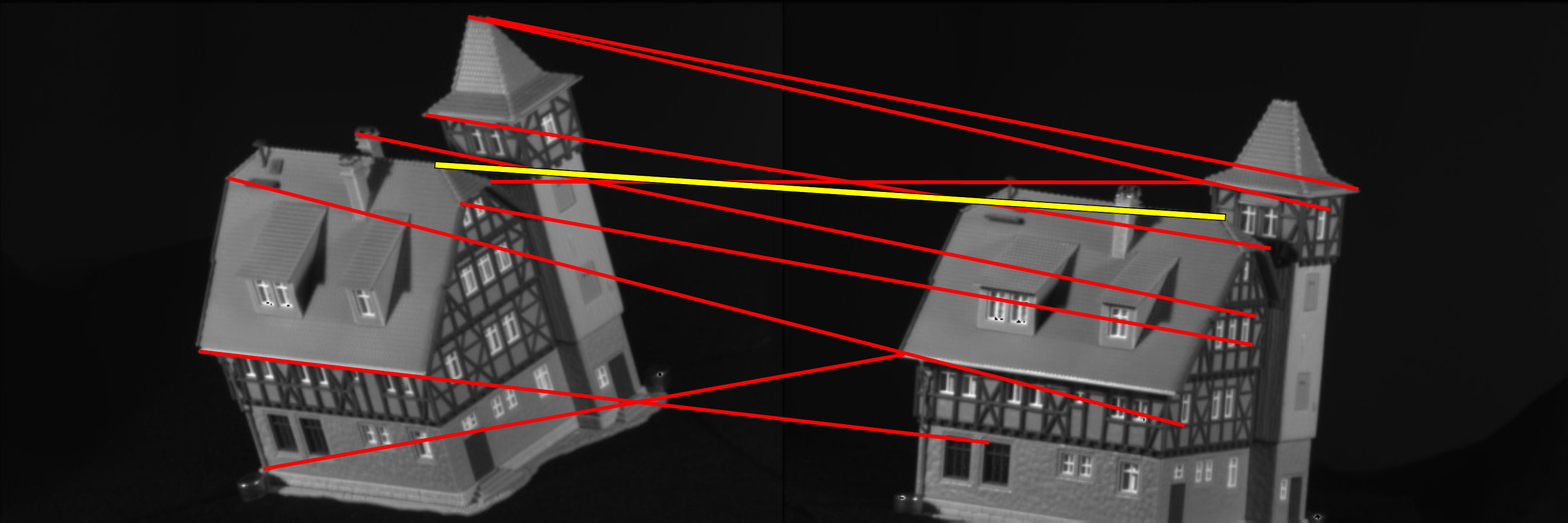} } \\  
    \subfloat[TM\cite{DucEtAl2011} (80.1)]{ \includegraphics[width=0.31\linewidth]{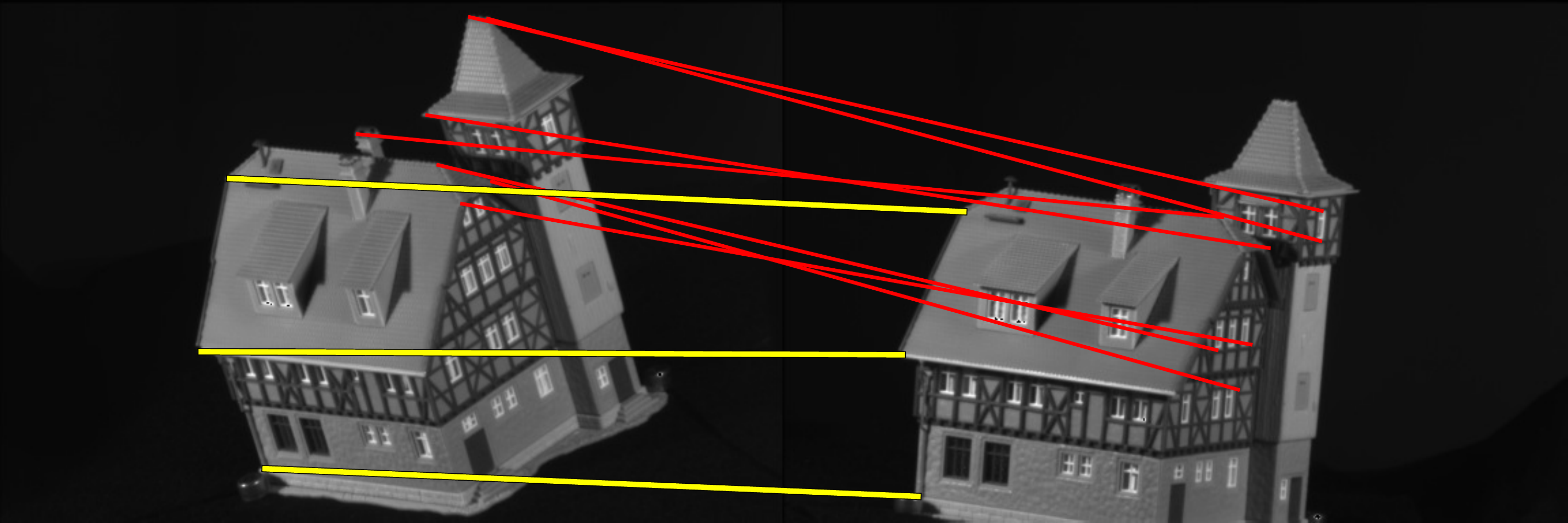} }  
    \subfloat[RRWHM\cite{LeeChoLee2011} (108.8)]{ \includegraphics[width=0.31\linewidth]{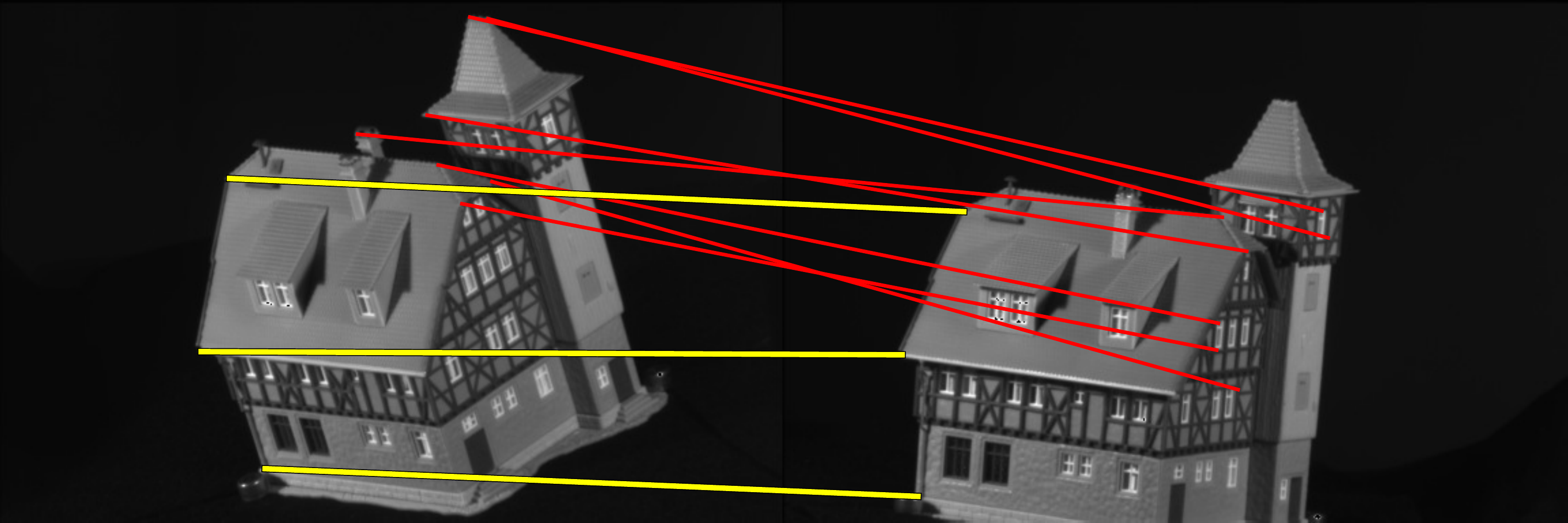} }  
    \subfloat[BCAGM\cite{Quynh2015} (194.4)]{ \includegraphics[width=0.31\linewidth]{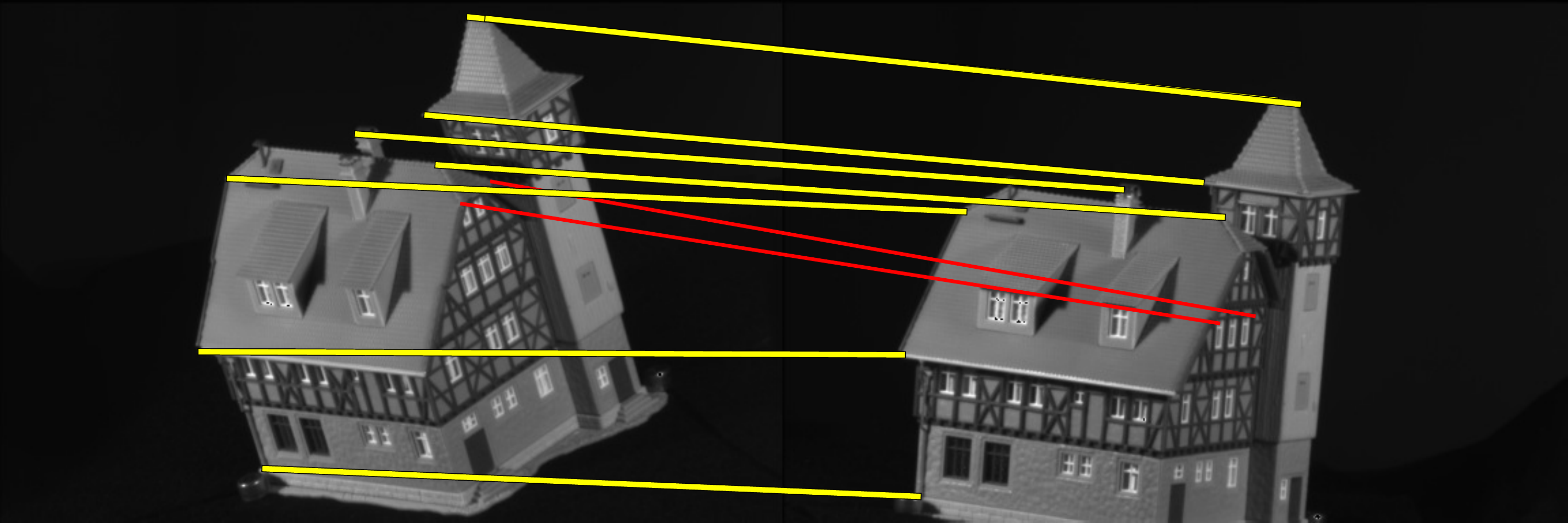} }  \\
    \subfloat[BCAGM+MP\cite{Quynh2015} (194.4)]{ \includegraphics[width=0.31\linewidth]{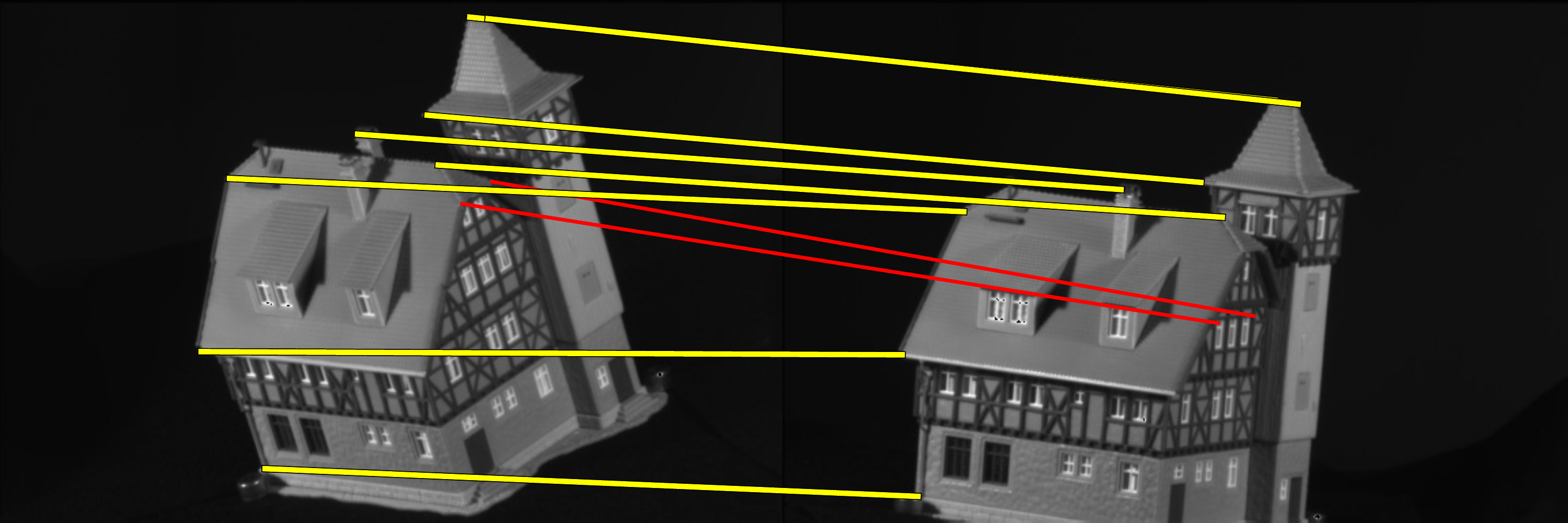} } 
    \subfloat[Adapt-BCAGM3 (194.4)]{ \includegraphics[width=0.31\linewidth]{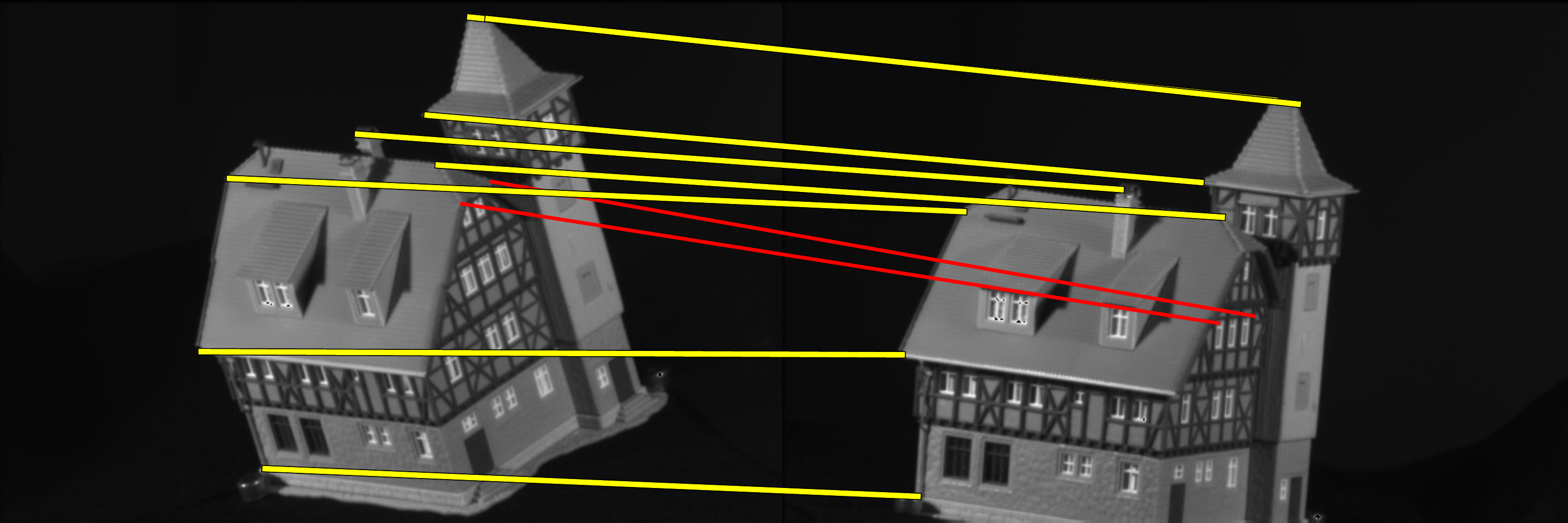} }  
    \subfloat[Adapt-BCAGM3+MP (194.4)]{ \includegraphics[width=0.31\linewidth]{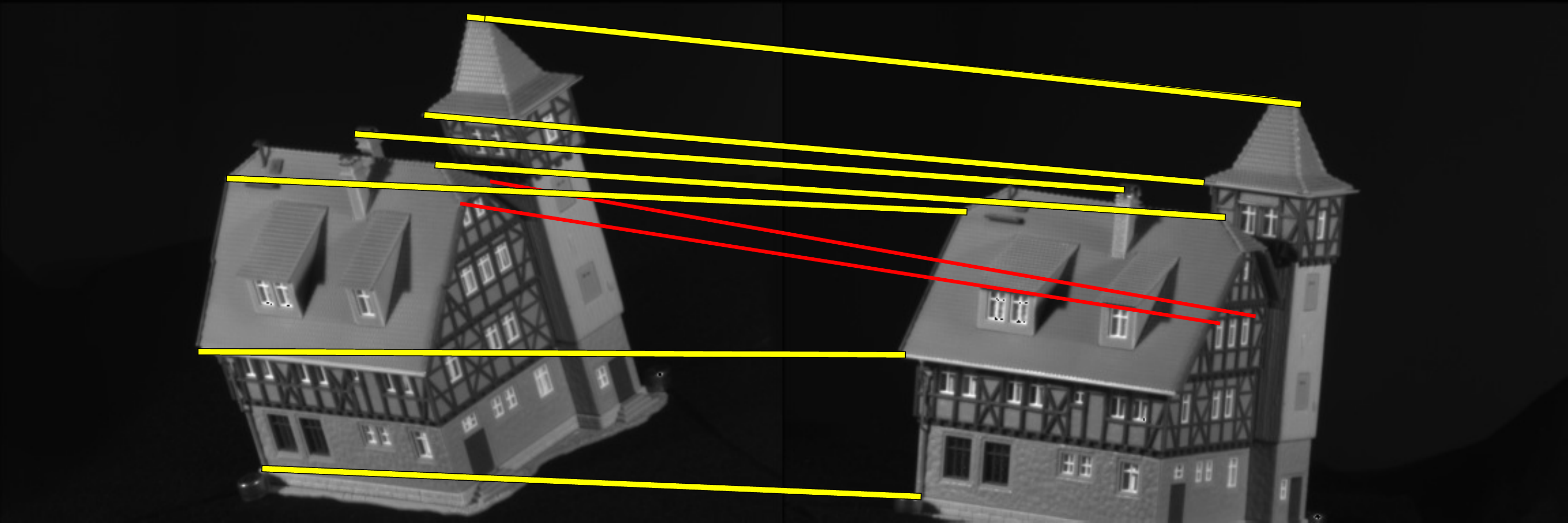} }
\caption{Demo of matching results on CMU House dataset with \textit{large baseline} ($\textrm{baseline}=80$).
a) Input images. Yellow dots denote inlier points, blue dots denote outlier points.
b) c) Matching results of previous second-order methods. 
d) e) Matching results of previous higher-order methods. 
f) g) h) i) Matching results of our higher-order approaches.
The yellow/red lines indicate correct/incorrect matches. 
Matching score is reported for each method. (Best viewed in color.)}
\label{fig2:house_matches}
\end{figure*}

\begin{figure*}
    \subfloat[Image 1]{ \includegraphics[width=0.24\linewidth]{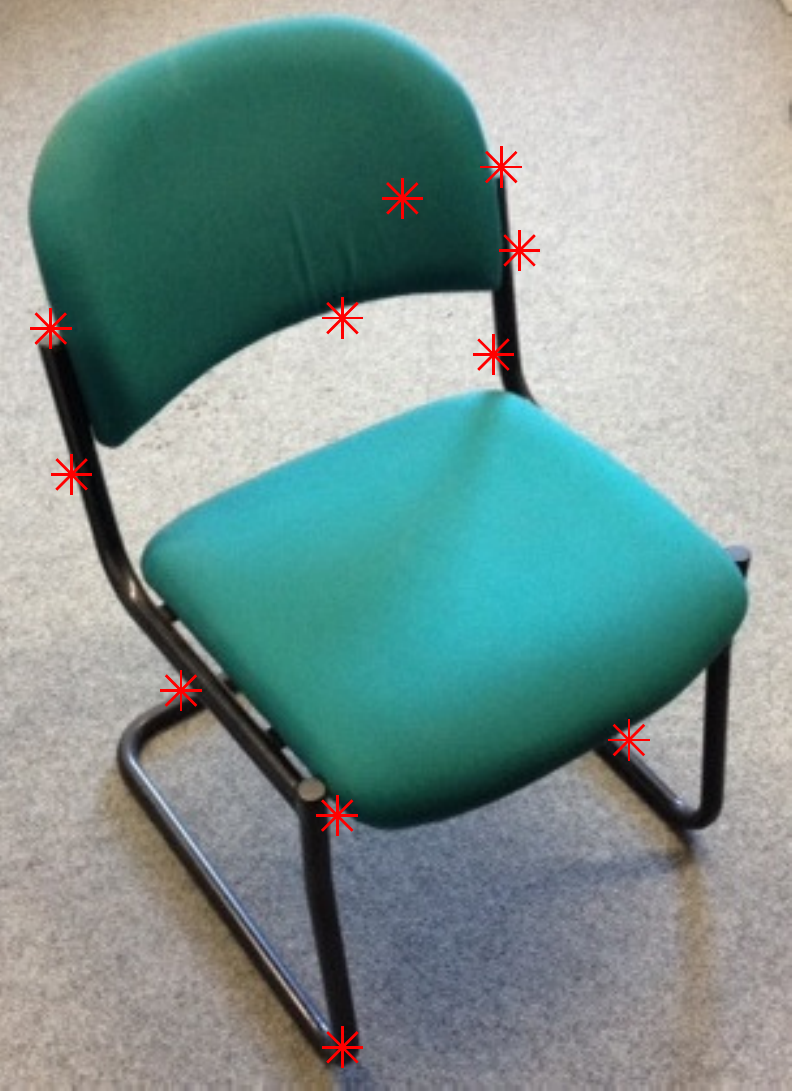} }  
    \subfloat[Image 2]{ \includegraphics[width=0.24\linewidth]{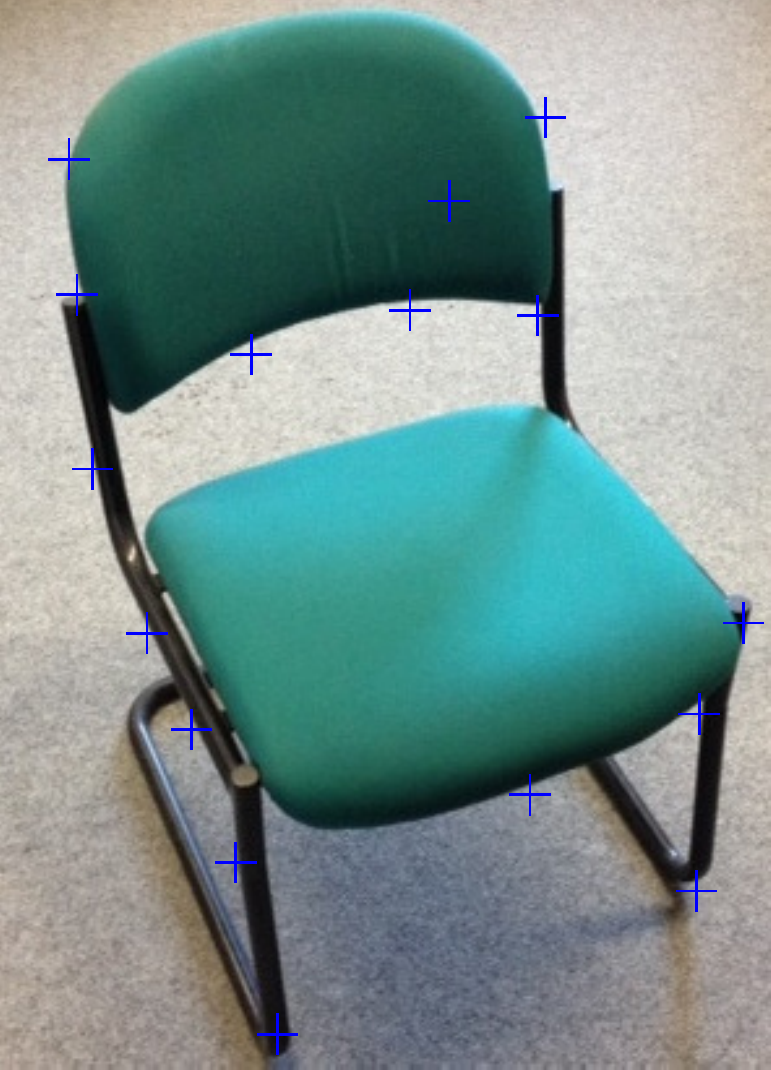} }  
    \subfloat[TM\cite{DucEtAl2011} (135)]{ \includegraphics[width=0.24\linewidth]{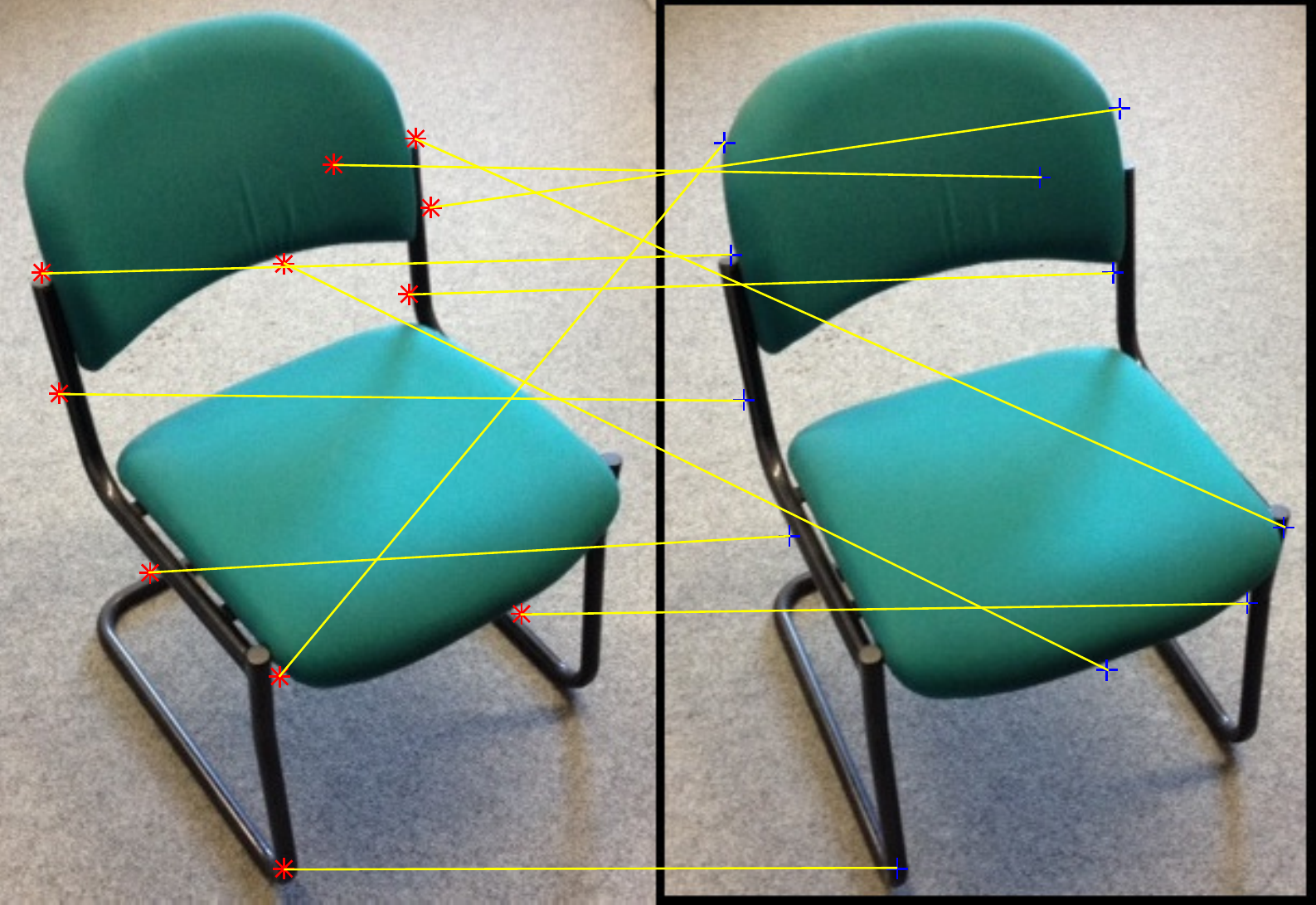} }  
    \subfloat[RRWHM\cite{LeeChoLee2011} (174)]{ \includegraphics[width=0.24\linewidth]{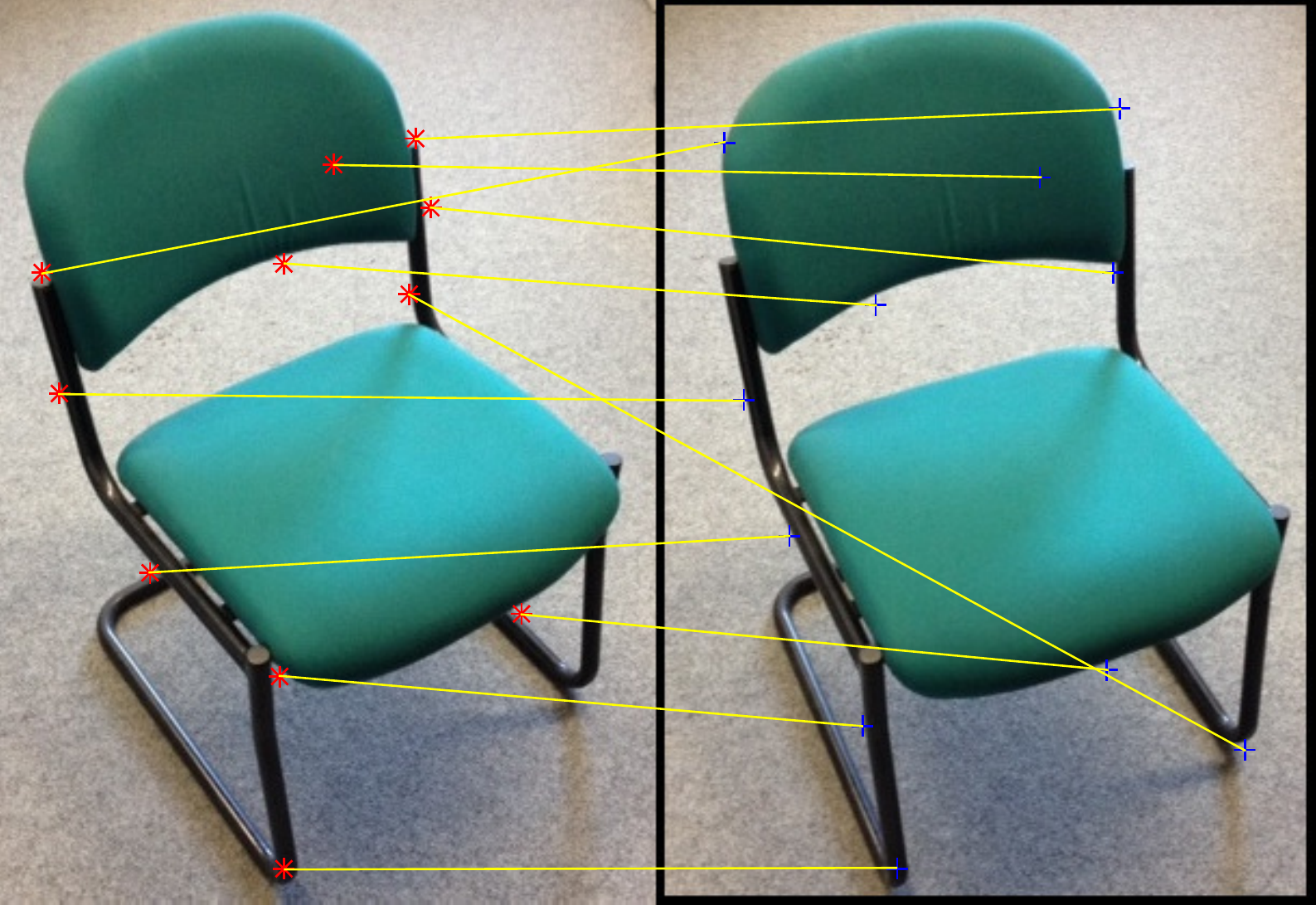} }  \\
    \subfloat[BCAGM\cite{Quynh2015} (195)]{ \includegraphics[width=0.24\linewidth]{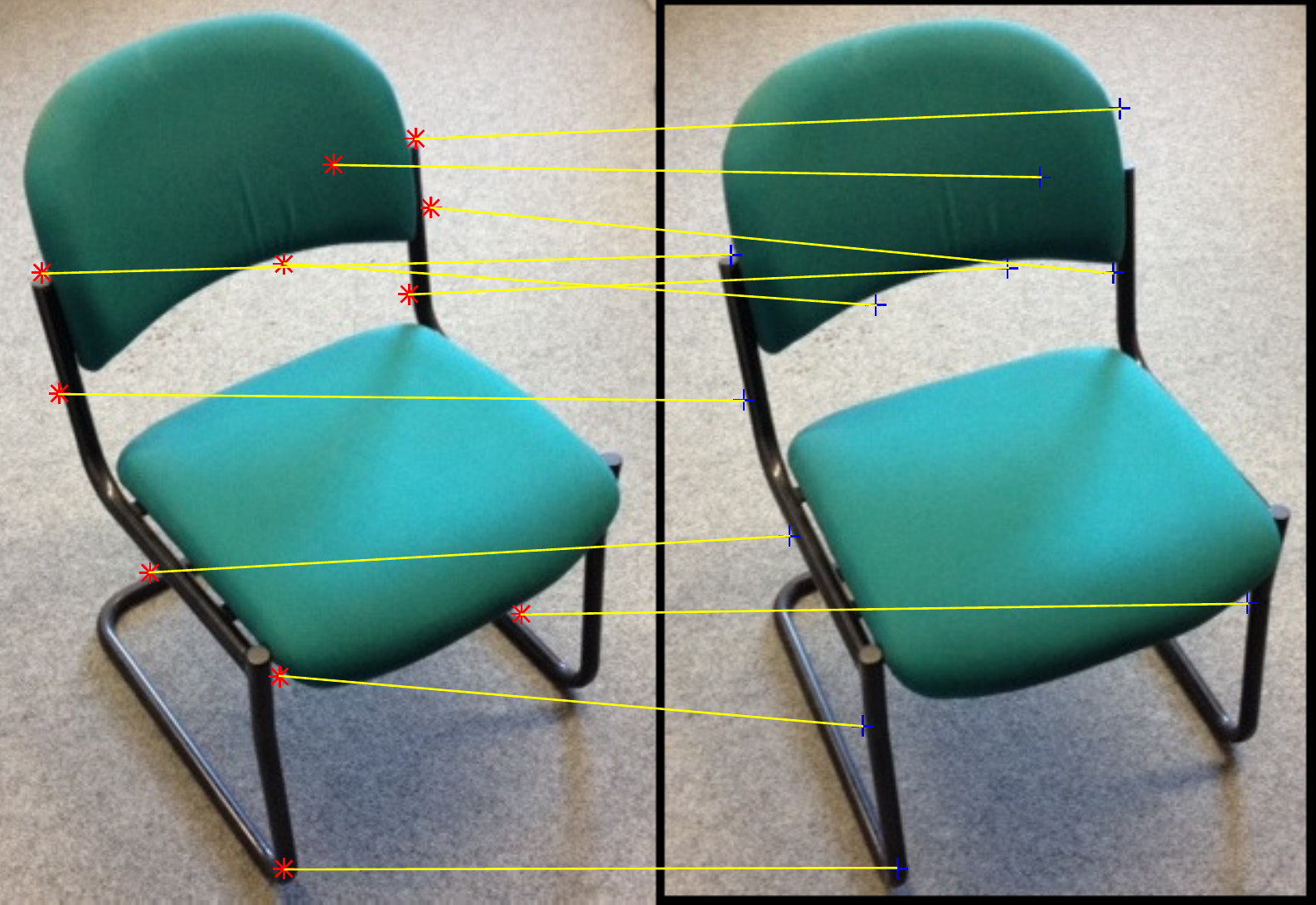} }  
    \subfloat[BCAGM+MP\cite{Quynh2015} (226)]{ \includegraphics[width=0.24\linewidth]{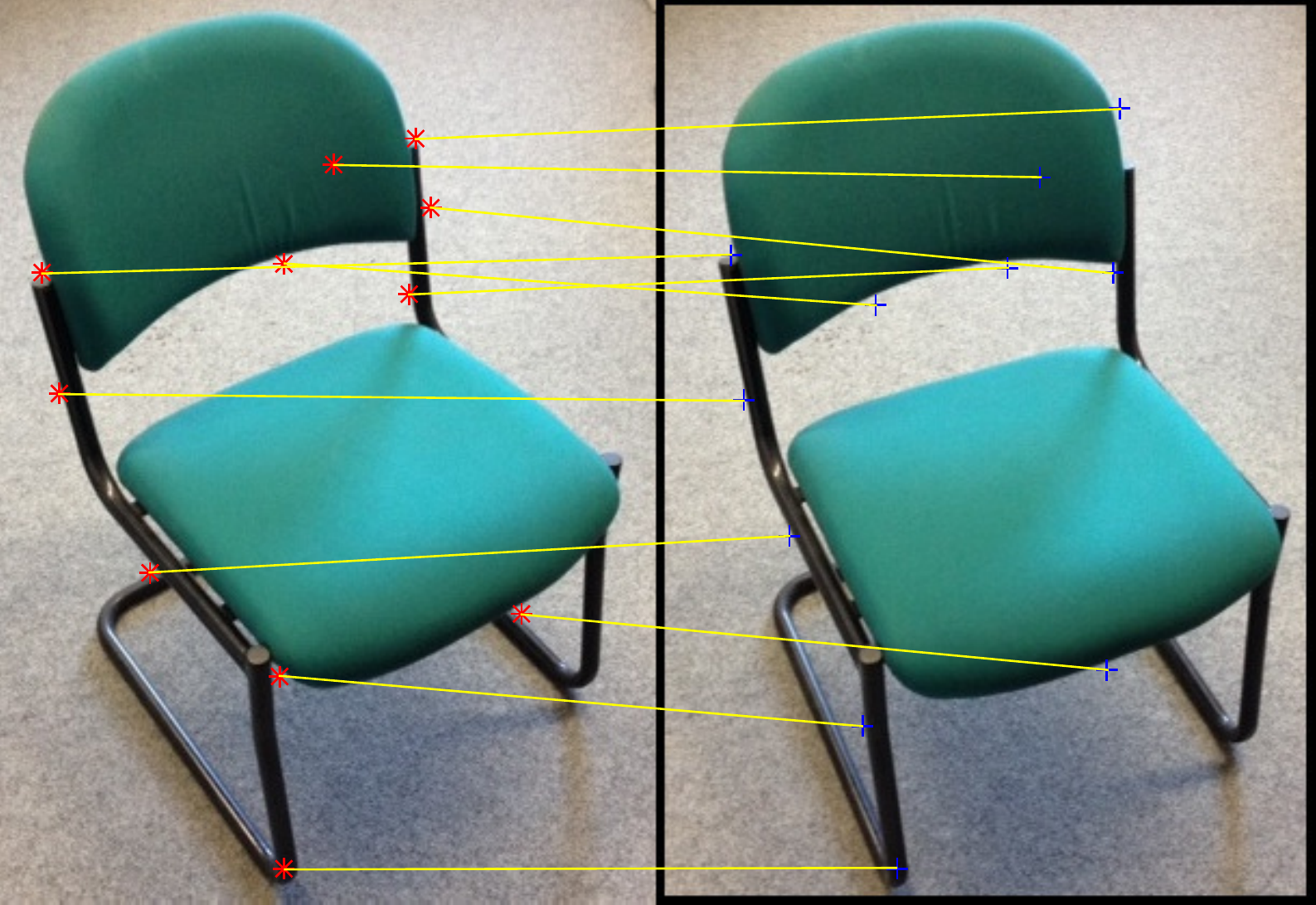} } 
    \subfloat[Adapt-BCAGM3 (218)]{ \includegraphics[width=0.24\linewidth]{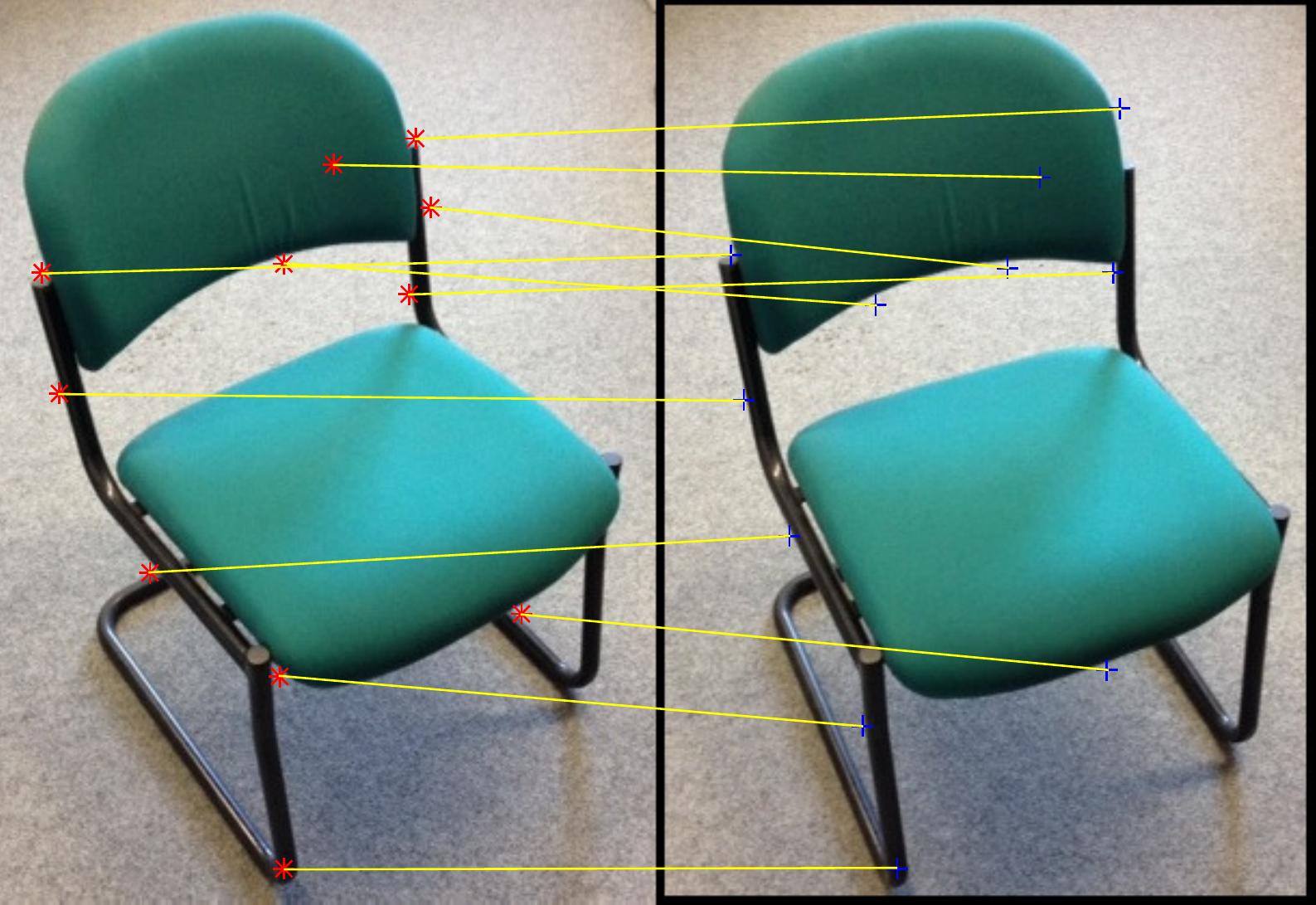} }  
    \subfloat[Adapt-BCAGM3+MP (226)]{ \includegraphics[width=0.24\linewidth]{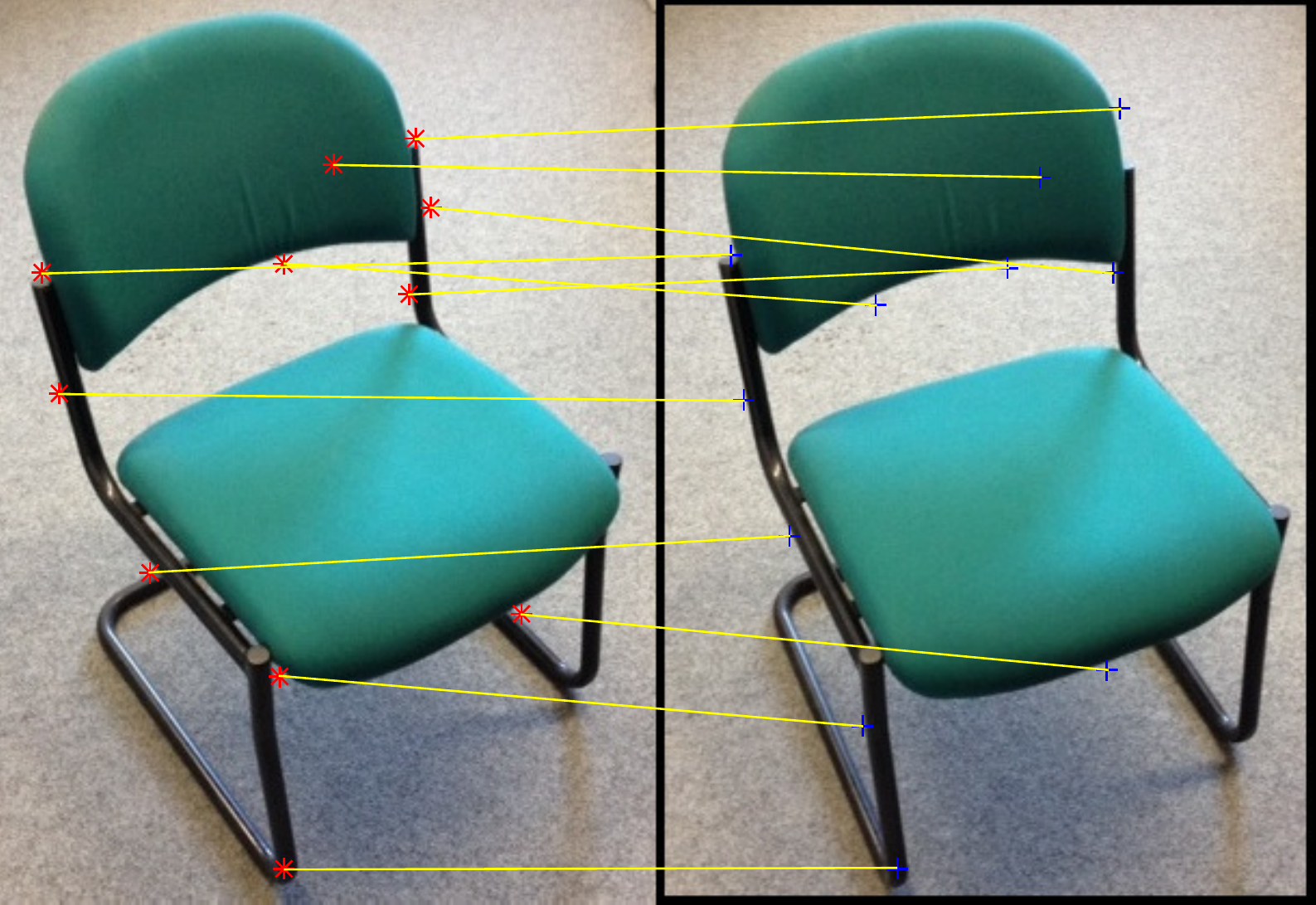} }
\caption{Demo of matching results. Matching score is reported for each method. (Best viewed in color.)}
\label{fig2:example_matches}
\end{figure*}

\begin{figure*}
    \subfloat[Image 1]{ \includegraphics[width=0.48\linewidth]{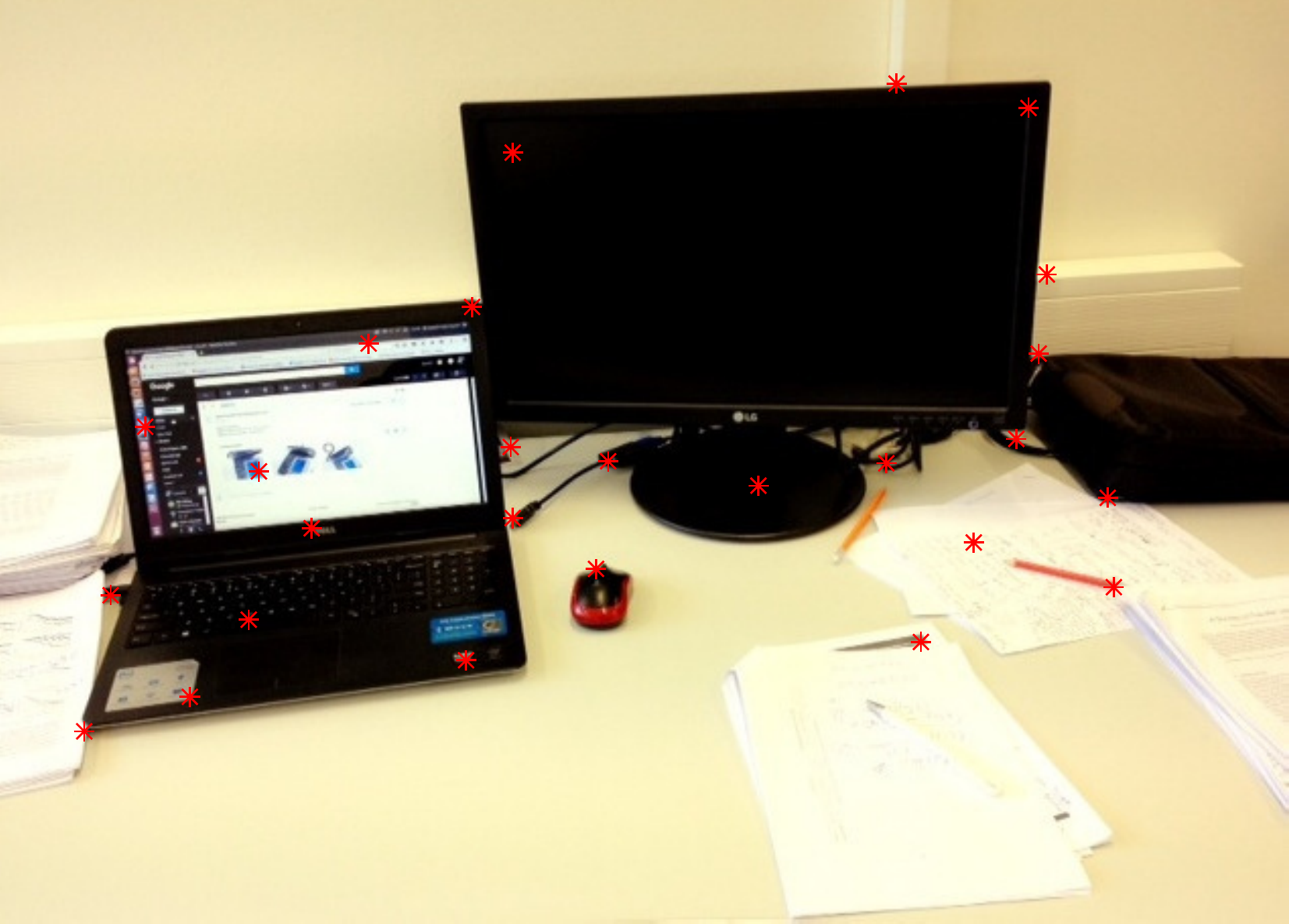} }  
    \subfloat[Image 2]{ \includegraphics[width=0.48\linewidth]{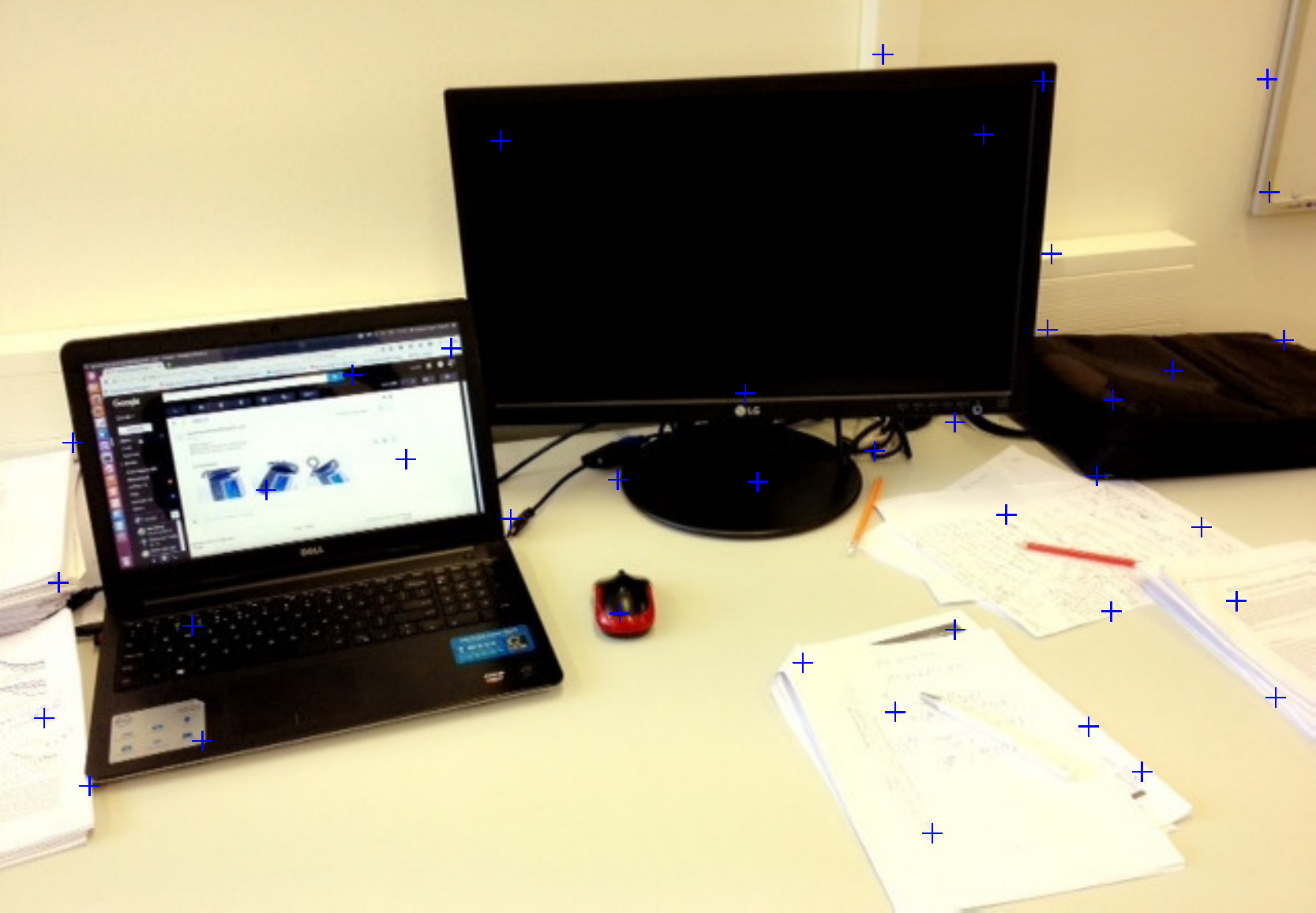} }  \\
    \subfloat[TM\cite{DucEtAl2011} (1071)]{ \includegraphics[width=0.48\linewidth]{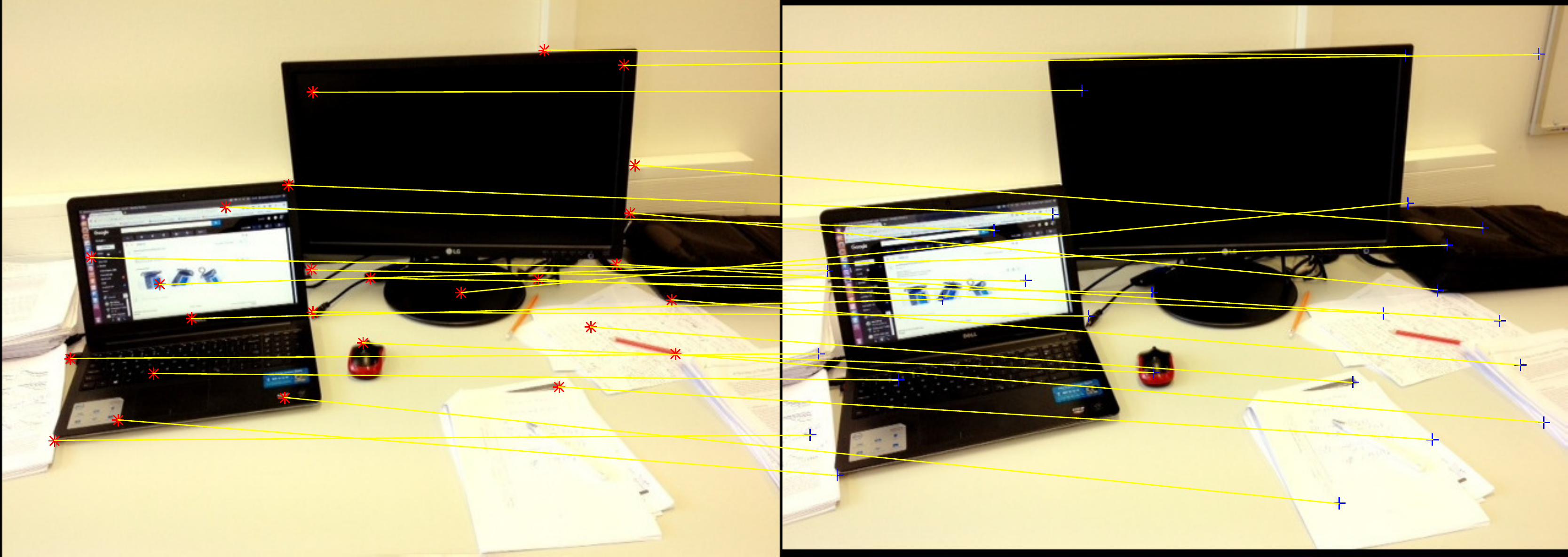} }  
    \subfloat[RRWHM\cite{LeeChoLee2011} (1639)]{ \includegraphics[width=0.48\linewidth]{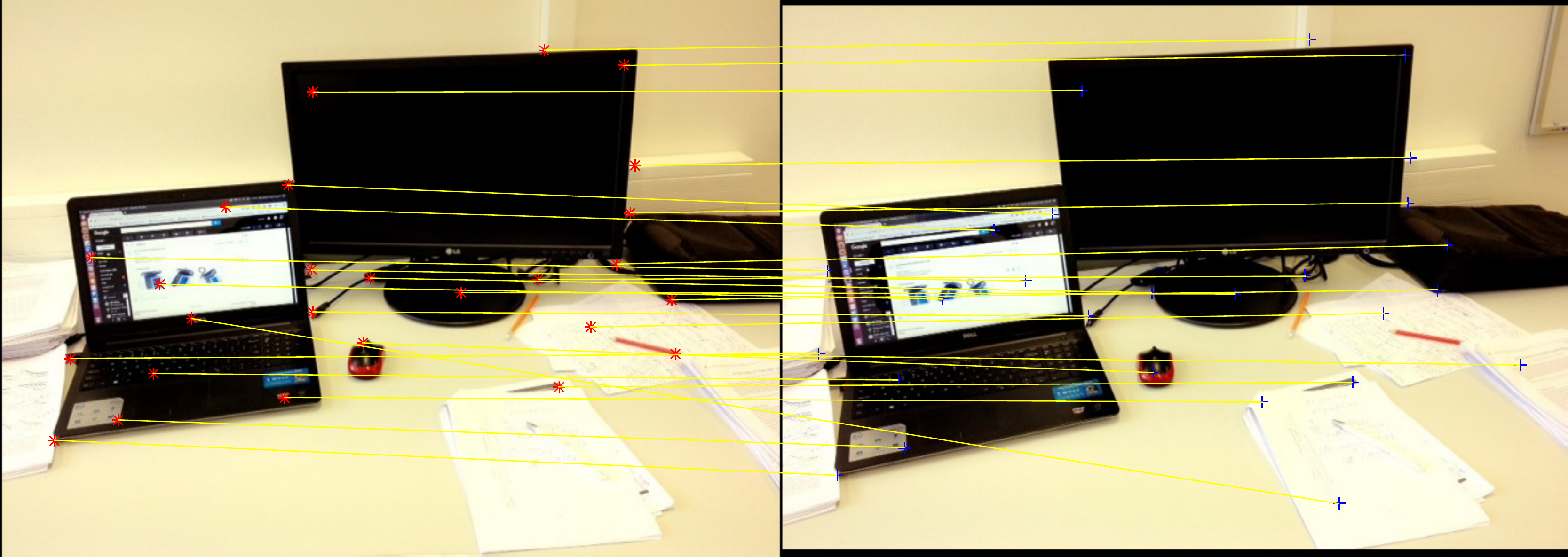} }  \\
    \subfloat[BCAGM\cite{Quynh2015} (1751)]{ \includegraphics[width=0.48\linewidth]{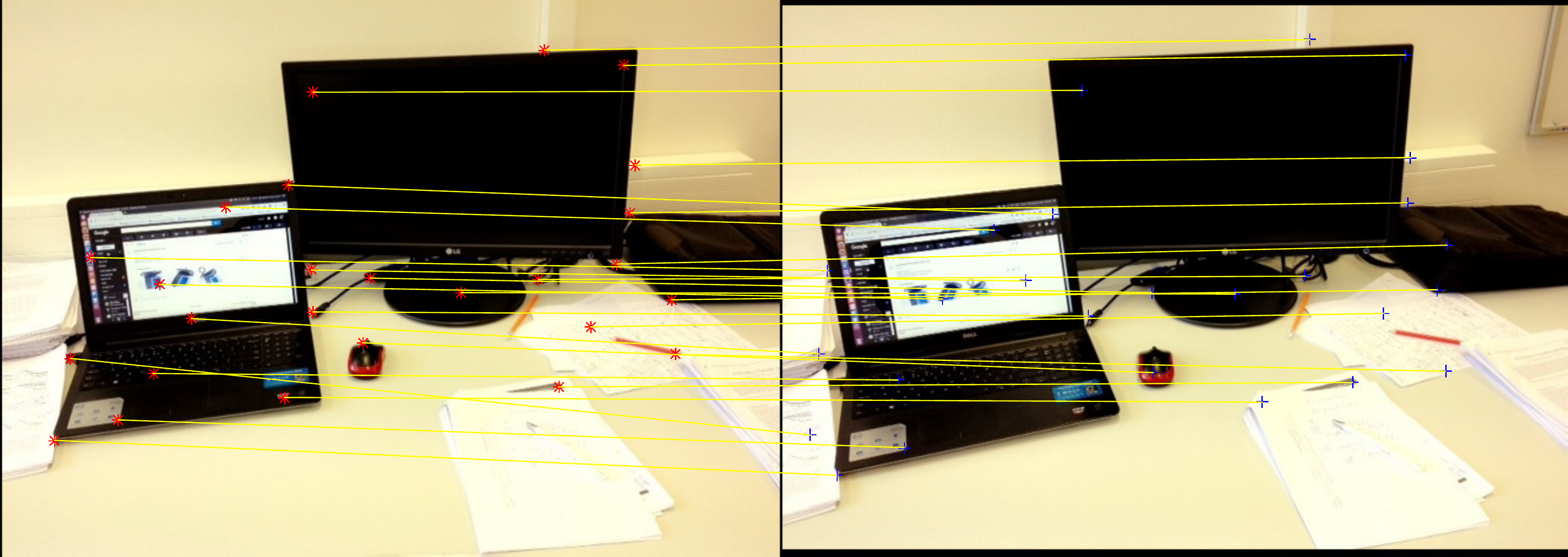} }  
    \subfloat[BCAGM+MP\cite{Quynh2015} (1751)]{ \includegraphics[width=0.48\linewidth]{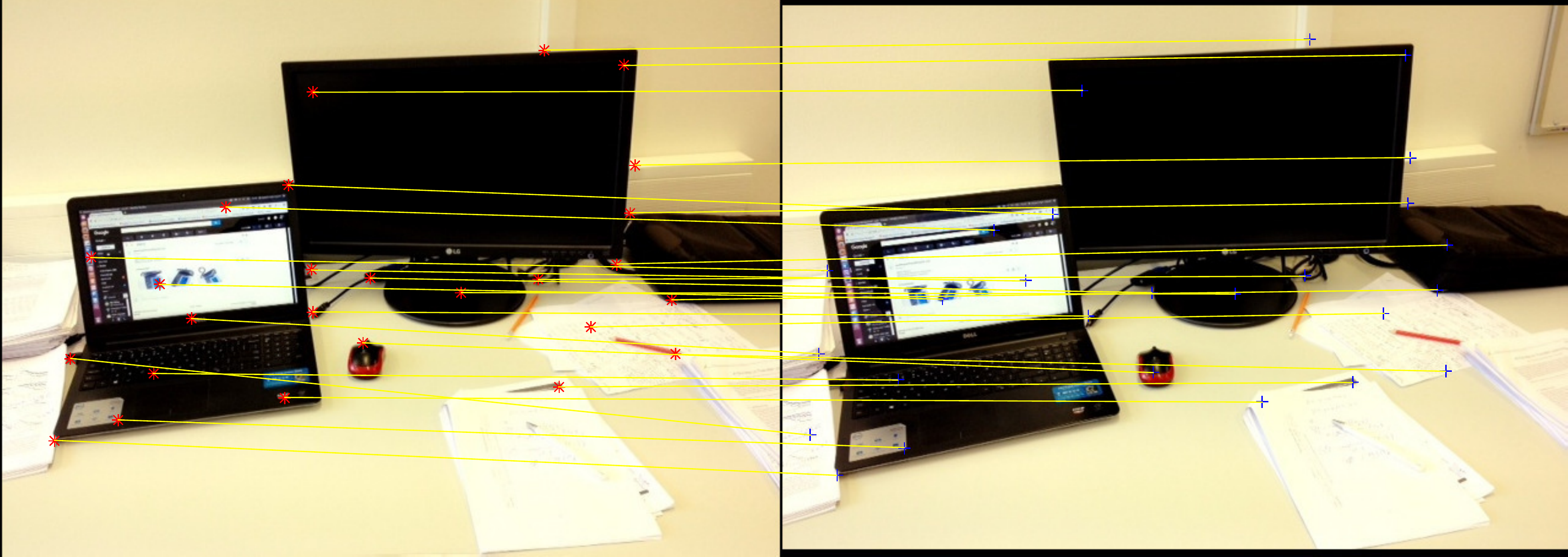} } \\  
    \subfloat[Adapt-BCAGM3 (1751)]{ \includegraphics[width=0.48\linewidth]{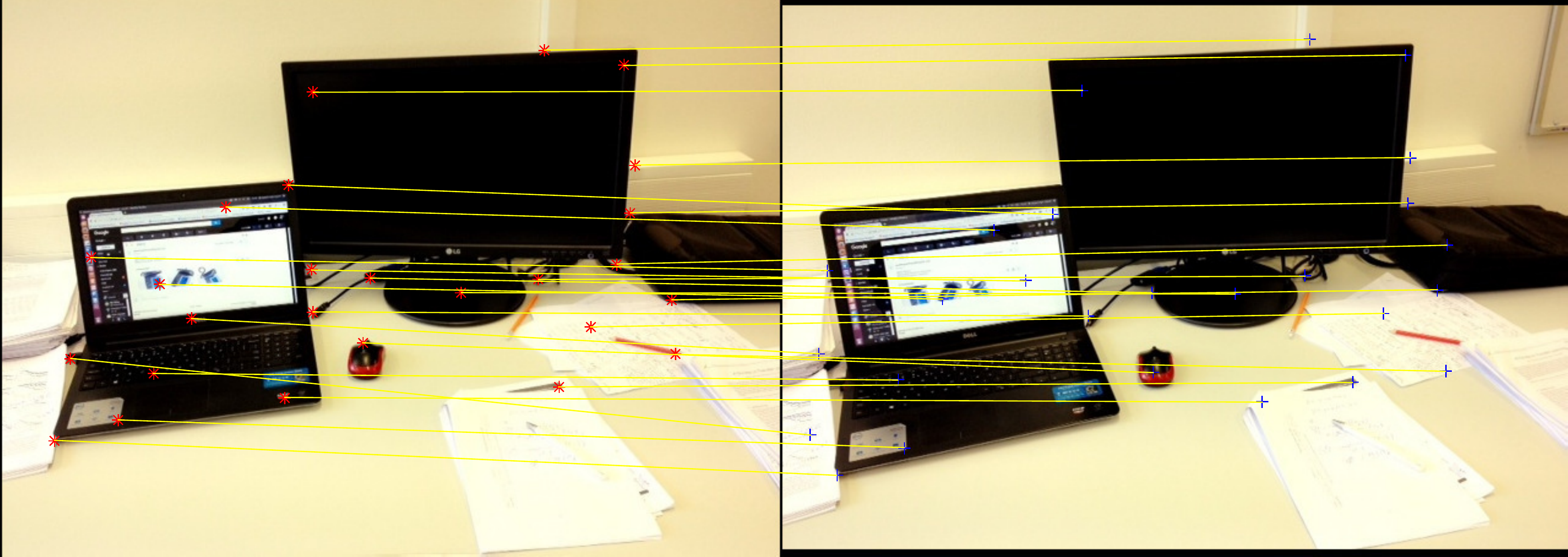} }  
    \subfloat[Adapt-BCAGM3+MP (1751)]{ \includegraphics[width=0.48\linewidth]{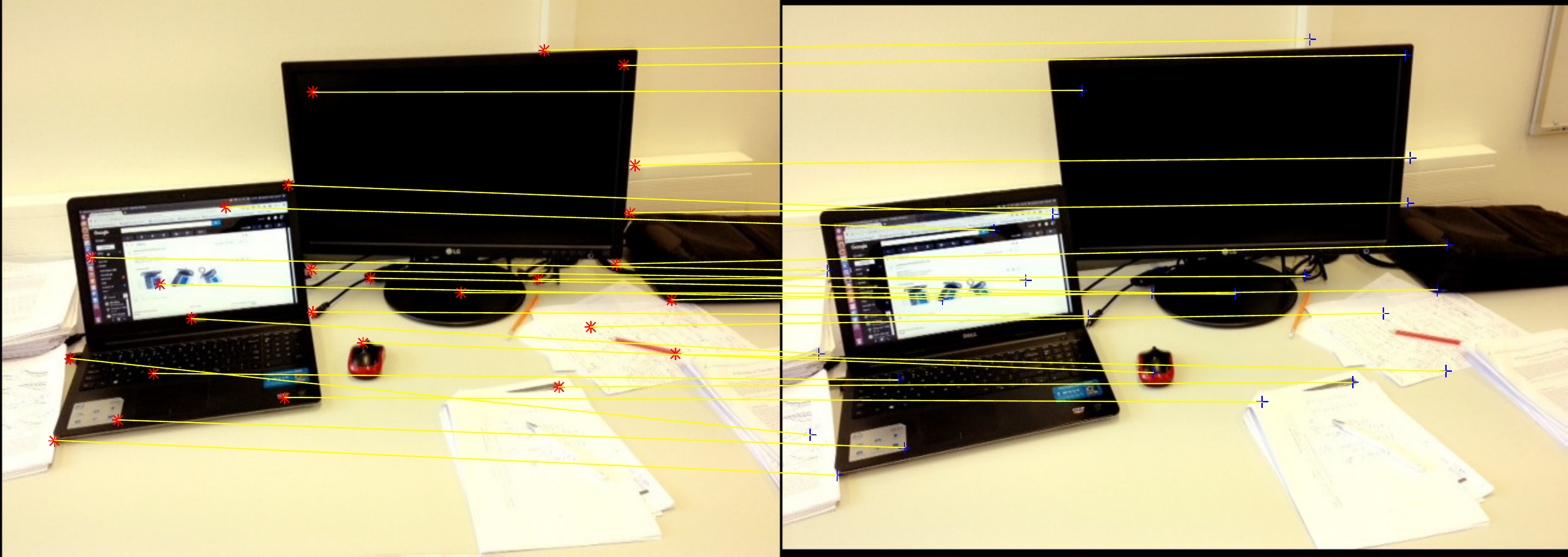} }
\caption{Demo of matching results. Matching score is reported for each method. (Best viewed in color.)}
\label{fig3:example_matches}
\end{figure*}

Fig. \ref{fig1:house_matches} and Fig. \ref{fig2:house_matches} show the examples of matching results on CMU House dataset.
Fig. \ref{fig2:example_matches} and Fig. \ref{fig3:example_matches}
shows the examples of matching results on realistic images 
where our algorithms return more reasonable matches and also achieve higher matching scores than 
the state-of-the-art higher-order approaches.








\end{document}